# Tactile MNIST: Benchmarking Active Tactile Perception

**Tim Schneider**[1,2], **Guillaume Duret**[2], **Cristiana de Farias**[1], **Roberto Calandra**[3], **Liming Chen**[2], **and Jan Peters**[4]

[1]Department of Computer Science, TU Darmstadt, Germany.
[2]LIRIS, CNRS UMR5205, Ecole Centrale de Lyon, France.
[3]LASR Lab & CeTI, TU Dresden, Germany.
[4]DFKI, Hessian.AI, and Centre for Cognitive Science, TU Darmstadt, Germany.

## Abstract

Tactile perception has the potential to significantly enhance dexterous robotic manipulation by providing rich local information that can complement or substitute for other sensory modalities such as vision. However, tactile sensing is an inherently local sensor modality, providing information only at the points of contact. This strict locality necessitates the use of active perception techniques to acquire useful, task-relevant information from manipulated objects or the environment in general. Hence, the agent must actively guide sensors toward regions with more informative or significant features and integrate this information over time to understand a scene or complete a task. While active perception and tactile sensing methods have received significant attention recently, both fields lack standardized benchmarks. To bridge this gap, we introduce the *Tactile MNIST Benchmark Suite*, an open-source, Gymnasium-compatible benchmark specifically designed for active tactile perception tasks, including classification, localization, pose estimation, and volume estimation. Our benchmark suite offers diverse simulation scenarios, from simple toy environments to complex tactile perception tasks using vision-based tactile sensors. Furthermore, we offer a comprehensive dataset comprising 13,580 synthetic 3D MNIST digit models and 153,600 real-world tactile samples collected from 600 3D-printed digits. Using this dataset, we train a CycleGAN for realistic tactile simulation rendering. By providing standardized protocols and reproducible evaluation frameworks, our benchmark suite facilitates systematic progress in tactile sensing and active perception more broadly. Code, datasets, and documentation are available on our project page: `https://timschneider42.github.io/tactile-mnist`.

## 1 Introduction

Tactile perception is fundamental for enabling agents to interact dexterously with their environments. Studies of humans with impaired touch reveal that they face significant challenges in grasping and performing routine manipulation tasks due to insufficient feedback about contact states between fingers and objects [1]. Moreover, touch often complements or even substitutes for other sensory modalities such as vision: we feel the shape of a hard-to-see object on a high shelf, count cookies in a jar without looking, or locate a key inside a bag purely by touch. However, unlike vision, which typically offers a broad field of view, touch provides highly localized yet information-rich feedback confined to the point of contact [2, 3]. This locality makes tactile perception fundamentally partial and sequential in nature, requiring agents to actively choose how to gather information through physical interaction. The tight coupling between control and sensory feedback is often referred to as active perception. More specifically, in active perception, an agent knows what it wishes to sense and must therefore decide *where*, *how*, and *when* to sample the environment [4, 5]. Progress in active tactile perception, however, requires standardized benchmarks for systematic evaluation.

Indeed, many areas of machine learning, perception, and control have benefited from standardized benchmarks. In computer vision, MNIST [6] established an early and widely used benchmark for handwritten digit recognition, while larger and more complex datasets such as ImageNet [7] and COCO [8] helped drive major advances in the field. Similarly, in control and reinforcement learning (RL), benchmark suites such as Gymnasium [9] and the DeepMind Control Suite [10] have provided standardized environments for evaluating sequential decision-making. Active tactile perception and active perception in general, however, lack such benchmarks. Tactile perception by itself has fewer broadly adopted benchmarks, and existing efforts often focus on specific sensors or tasks [11, 12].

In this work, we tackle the problem of benchmarking *active tactile perception*. We identify two central requirements for this setting: *memory* and *exploration efficiency*. Memory has only recently started to receive attention in benchmarks for long-context reasoning and partially observable decision-making [13, 14]. In our setting, memory is essential because tactile observations are local and often ambiguous in isolation. A single contact rarely suffices to determine an object's identity, pose, or shape; instead, an agent must integrate information across contacts and update its belief about the environment over time. Exploration efficiency is equally important because tactile sensing requires physical interaction. That is, each contact consumes time and can disturb the object or environment. Thus, an active tactile benchmark should evaluate not only whether an agent eventually reaches the correct prediction, but also whether it does so through informative and efficient interactions.

Thus, we introduce *Active Perception Gym* (`ap_gym`)[1], a framework compatible with Gymnasium [9], designed to benchmark active perception algorithms. `ap_gym` includes 11 toy scenarios where an agent must learn to efficiently extract information to solve perception tasks. And, building on `ap_gym`, we present the *Tactile MNIST Benchmark Suite* (TMBS)[2], which extends the framework to simulated active tactile perception problems. In TMBS, agents control a simulated GelSight Mini sensor [15] without access to visual inputs. Tactile perception tasks include MNIST-style classification of 3D digit models, pose estimation of tools on platforms, object volume estimation, and object counting. Across all tasks, the agent must actively determine *what* to sense and strategically select *where* and *when* to explore through touch. Thus, solving these tasks requires solving a dual problem: making accurate predictions from past observations while optimizing an exploration policy to maximize information gain. A visualization of our framework and tasks in `ap_gym` and TMBS is shown in Fig. 1.

With this benchmark, our goal is to provide a reproducible and accessible evaluation framework for the active perception and tactile sensing communities. To that end, TMBS is a fully *simulated* environment that is easy to set up and enables rapid, consistent comparisons, without the need to replicate a complex real-world setup. However, we acknowledge the inherent challenges and noise present in real-world tactile data that are often absent in simulation. To help bridge this sim-to-real gap, we complement our simulated environment with a large-scale dataset of 13,580 high-fidelity 3D object models. From this collection, we 3D-printed 600 objects and constructed a curated real-world dataset comprising 153,600 tactile contacts, each annotated with detailed temporal and spatial metadata. We further leverage this dataset to train a CycleGAN [16], enabling the rendering of realistic tactile signals within the simulation, and show in a proof-of-concept experiment that it helps policies trained in simulation transfer to real tactile data (Appendix H).

In summary, our main contributions are:

- To the best of our knowledge, we introduce the first benchmark suite for active *tactile* perception. Its tasks range from toy problems to challenging, high-dimensional scenarios.
- We further introduce `ap_gym`, an extensible framework for generic active perception algorithms. `ap_gym` is Gymnasium-compatible and, thus, integrates with existing RL pipelines.
- We provide an open dataset of 13,580 high-resolution 3D models of handwritten digits, designed for both simulated tactile-image generation and physical 3D printing.
- From our library of 3D models, we 3D-printed 600 objects and captured 153,600 tactile contacts using a GelSight Mini sensor, each annotated with spatial location and class labels.

[1] https://github.com/TimSchneider42/active-perception-gym

[2] https://github.com/TimSchneider42/tactile-mnist

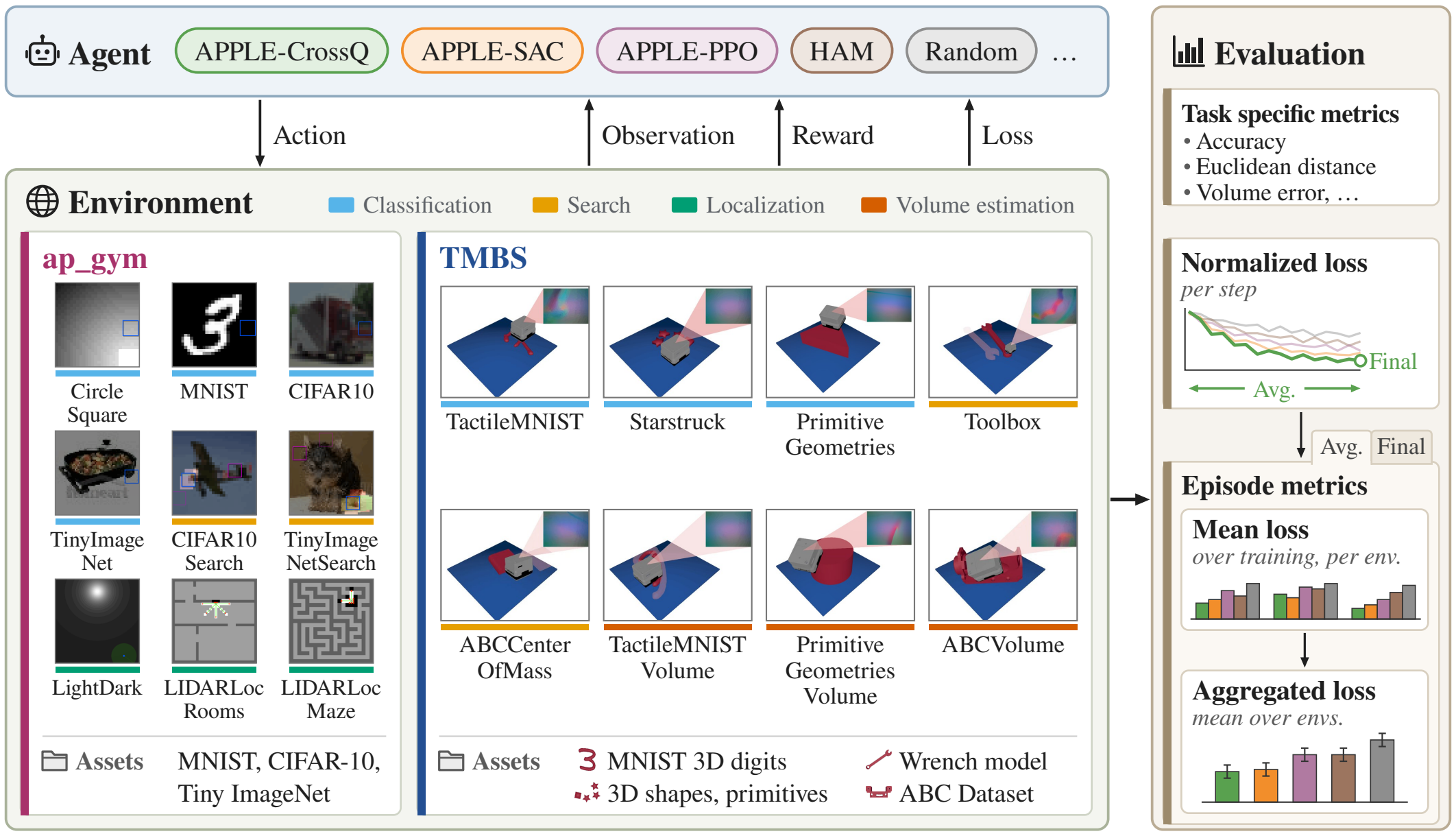

Figure 1: Overview of the *Active Perception Gym* (`ap_gym`), the *Tactile MNIST Benchmark Suite* (TMBS), and their associated assets. At the top, the agent interacts with the environment by sending actions and receiving observations, rewards, and losses. Below, we depict each environment from `ap_gym` and TMBS, colored by task category, along with the included asset sets. On the right, we show the evaluation: task-specific metrics and the normalized loss are available at every step. Per-step losses are summarized per episode (average and final), averaged over training into a mean loss per environment, and aggregated across environments.

## 2 Related Work

**(Active) Tactile Perception:** Tactile sensing enables robots to infer object geometry, texture, and material properties through physical contact, complementing or, in some cases, substituting vision. Vision-based tactile sensors such as GelSight [15] and DIGIT [17] have become widely available, producing high-resolution "tactile images" that capture local surface features and force distributions. These rich signals have been exploited for shape reconstruction and material recognition [18–22], force estimation [23, 24], and in-hand pose estimation [25], as well as for dexterous manipulation [12, 26–31], insertion and placing [32–34], imitation learning [35–37], and teleoperation [38]; see [39] for a recent survey.

Much of the work in tactile sensing, however, focuses on passive touch: the robot either registers a single contact or follows a predefined exploration policy. Inspired by the successes of active vision (as well as by early research on active touch in robotics [5, 40]), recent efforts have revisited the idea of tactile exploration as an active, decision-driven process. In this framing, the robot dynamically selects where and how to touch next, rather than relying on a fixed sequence of actions. Gaussian processes and Bayesian optimization have been used to drive active exploration, yielding significant improvements in tasks such as shape reconstruction, texture classification, and grasp planning [18, 41, 42]. RL-based approaches [43] tackle active exploration in high-dimensional tactile state spaces. Furthermore, [44] introduced HAM, a selective-attention mechanism, to optimize scene exploration, and in [45] task-agnostic strategies generalize active touch across different objectives. Additionally, [46] demonstrated how Kinect-based vision can guide active touch for material classification, and [47] proposed a self-supervised visuo-tactile pretraining scheme that benefits both passive and active perception tasks.

**Benchmarking Methods for Tactile Sensing & Active Perception:** Over the past decade, the active vision community has produced a number of datasets and challenges to evaluate a range of tasks. Early work such as the Active Vision Dataset (AVD) provided large-scale Kinect captures for navigation and class-incremental learning tasks [48, 49]. Subsequent efforts explored next-best-view planning for classification (NBV_Bench [50]), heuristic and data-driven view selection for grasp synthesis on YCB objects [51], and embodied 3D exploration in real indoor scenes (R3ED [52]).

Table 1: Overview of environments, including task types (classification/regression), descriptions, and assets.

| Suite | Environment | Type | Task Description | Assets |
|---|---|---|---|---|
| `ap_gym` | CircleSquare | Class. | Tell whether an image shows a circle or a square by moving a small glimpse. | |
| | MNIST | Class. | Standard MNIST digit recognition task. | MNIST [6] |
| | CIFAR10 | Class. | Classify images into 10 categories by moving a small glimpse. | CIFAR-10 [67] |
| | TinyImageNet | Class. | Classify images into 200 categories by moving a small glimpse. | Tiny ImageNet [68] |
| | CIFAR10Search | Reg. | Localize a glimpse within a natural image by moving a small glimpse. | CIFAR-10 [67] |
| | TinyImageNetSearch | Reg. | Localize a glimpse within a natural image by moving a small glimpse. | Tiny ImageNet [68] |
| | LightDark | Reg. | Position estimation from brightness-dependent noisy observations. | |
| | LIDARLocRoomsStatic | Reg. | Localization in a static map with LIDAR readings. | |
| | LIDARLocMazeStatic | Reg. | Localization in a static maze with LIDAR readings. | |
| | LIDARLocRooms | Reg. | Localization in procedurally generated maps with LIDAR readings. | |
| | LIDARLocMaze | Reg. | Localization in procedurally generated mazes with LIDAR readings. | |
| **TMBS** | TactileMNIST | Class. | Touch-based digit classification using a GelSight Mini. | MNIST 3D digits |
| | Starstruck | Class. | Count the number of stars among other objects using a GelSight Mini. | 3D shapes |
| | PrimitiveGeometries | Class. | Tell apart geometric primitives using a GelSight Mini. | 3D shapes |
| | Toolbox | Reg. | Estimate the 2D pose of a wrench using a GelSight Mini. | 3D wrench |
| | ABCCenterOfMass | Reg. | Estimate the center of mass of a CAD model using a GelSight Mini. | ABC Dataset [69] |
| | TactileMNISTVol. | Reg. | Estimate the volume of digits using a GelSight Mini. | MNIST 3D digits |
| | PrimitiveGeometriesVol. | Reg. | Estimate the volume of geometric primitives using a GelSight Mini. | 3D shapes |
| | ABCVolume | Reg. | Estimate the volume of a CAD model using a GelSight Mini. | ABC Dataset [69] |

Simulation-based approaches such as the Robotic Vision Scene Understanding Challenge [53] and Active Object Search [54] have further expanded evaluation protocols for semantic SLAM and object detection tasks. More recently, multimodal active perception has been addressed by ActiView, which tests an agent's ability to zoom and pan to answer vision-language queries [55].

In parallel, tactile sensing research has released a diverse set of characterization benchmarks (TIP Bench. [19]), texture and material recognition datasets (ActiveCloth [46], ViTac [56], SSVTP [47], GelFabric [57]), and cross-modal vision–touch benchmarks and datasets (VisGel [58], Touch and Go [59], PHYSICLEAR [21], FeelSight [12], TacBench [11]). Large-scale multimodal datasets such as Touch100k [60], TVL [20], and FoTa [61] now exceed tens of thousands of samples across vision, touch, and language modalities. Multisensory datasets like ObjectFolder [62, 63] further integrate tactile, visual, and audio data. In a similar manner, efforts to standardize vision-based tactile simulation have led to platforms like TACTO [64] and Taxim [65], which enable high-resolution visuo-tactile data generation, and physically based rendering narrows the sim-to-real gap of tactile images [66]. Despite these efforts, only the datasets provided by ActiveCloth [46] and SSVTP [47] support downstream active tactile perception tasks. However, these works do not provide standardized evaluation protocols or benchmarks to systematically evaluate active perception approaches. A comparison of existing datasets and benchmarks is summarized in Appendix J. To the best of our knowledge, our proposed Tactile MNIST Benchmark Suite is the first to introduce a dedicated and reproducible benchmark for *active tactile perception*, where tactile exploration is an integral component of the perceptual process.

## 3 Framework: Benchmarking Active Perception

In this section, we introduce both the *Active Perception Gym* (`ap_gym`), a framework for benchmarking active perception algorithms, and the *Tactile MNIST Benchmark Suite* (TMBS), which introduces environments for eight active tactile perception tasks. Fig. 1 depicts an overview of our framework.

### 3.1 Active Perception

In active perception tasks, an agent's main objective is to gather information and make predictions about a desired property of the environment, e.g., the class label or pose of an object. Examples of such properties could be the location of an object in the case of a search task or the class of an object for the agent in the case of a classification task. To gather information, the agent must interact with the environment, e.g., by moving a sensor around a platform in the case of TMBS.

We model active perception as an episodic process, where the agent can take a number of time-discrete sequential actions until the episode terminates and is reset. At every step, the agent obtains an observation (e.g., a tactile glimpse) from the environment that reveals some information but never the full state at once. Formally, that makes active perception problems a special case of Partially Observable Markov Decision Processes (POMDPs).

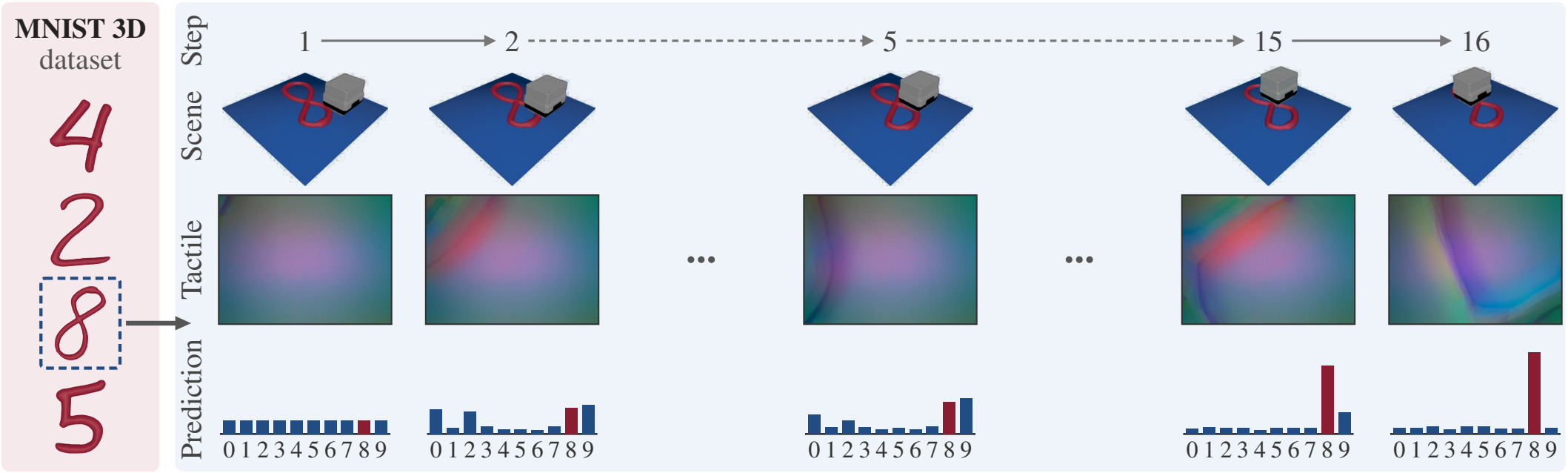

Figure 2: Illustration of the TactileMNIST classification task. Here, in each episode, one random digit from the MNIST 3D dataset is selected and presented to the agent. The agent can then move the sensor around and touch the object 16 times before the episode terminates. Notably, it does not receive any visual input and has to rely solely on tactile readings. After every touch, it has to make a prediction about the class label (the digit's numeric value), and its performance is measured by the average prediction accuracy throughout the episode.

POMDPs are defined by the tuple $(S, A, T, R, \Omega, O, \gamma)$, consisting of hidden states $S$, actions $A$, a transition function $T : S \times A \times S \to [0, 1]$, a reward function $R : S \times A \to \mathbb{R}$, a set of observations $\Omega$, an observation function $O : S \times A \times \Omega \to [0, 1]$, and a discount factor $\gamma \in [0, 1]$. The objective of the agent in a POMDP is to maximize the expected cumulative reward over time by selecting actions based on its belief about the underlying state. Since the agent does not have direct access to the true state, it maintains a belief distribution over states, updating it using observations and the observation function. The environment evolves according to the transition function, where taking an action leads to a probabilistic transition to a new state, which in turn generates an observation based on the observation function.

In the case of active perception problems, we assume that the hidden state $S$, the action $A$, the reward function $R$, and the transition function $T$ have specific structures. First, we assume that the target property the agent is tasked to predict is part of the hidden state. Hence, $S$ is defined as $S = S_{\text{base}} \times Y^*$, where $S_{\text{base}}$ is the set of base (hidden) states of the environment and $Y^*$ is the set of prediction targets. E.g., $Y^*$ could be the set of classes in a classification task or the set of possible locations in a localization task, while $S_{\text{base}}$ contains all the other hidden state information. To allow the agent to make predictions, the action space $A$ is defined as $A_{\text{base}} \times Y$, where $A_{\text{base}}$ is the base action space and $Y$ is the prediction space. The base action space $A_{\text{base}}$ contains all the actions the agent can take to interact with the environment, while $Y$ is the set of possible predictions the agent can make. Crucially, environments are defined in a way that the agent's prediction never influences the hidden state of the environment. Thus, the transition function $T$ is defined as $T(s, a, s') = T(s, (a_{\text{base}}, y), s') = T_{\text{base}}(s, a_{\text{base}}, s')$. An example of a base action could be a desired movement of the tactile sensor, while the prediction could be the logits of the agent's current class prediction. Finally, the reward function is defined as $R(s, a) = R((s_{\text{base}}, y^*), (a_{\text{base}}, y)) = R_{\text{base}}(s_{\text{base}}, a_{\text{base}}) - \ell(y^*, y)$, where $R_{\text{base}}$ is the base reward function and $\ell$ is a differentiable loss function. An example of a base reward could be an action regularization term, while the loss function $\ell$ could be a cross-entropy loss in a classification task. Hence, the agent has to make a prediction in every step, encouraging it to gather information quickly to maximize its prediction reward early on. A visualization of this process on the TactileMNIST digit classification task is shown in Fig. 2.

### 3.2 Active Perception Gym

`ap_gym` models active perception tasks as episodic processes in a way that is fully compatible with Gymnasium [9]. Each task is defined as a Gymnasium environment, bundled with the differentiable loss function $\ell(y^*, y)$ and the prediction target $y^*$. Since the loss functions need to convey gradient information to the learning algorithm, we currently provide them as either JAX [70] or PyTorch [71] functions, but more autograd frameworks might be supported in the future. During roll-outs, `ap_gym` automatically computes task-dependent metrics, such as accuracy for classification, or Euclidean distance for regression.

ap_gym provides a family of lightweight environments designed to isolate core exploration and decision-making behaviors in active perception. As summarized in Table 1, ap_gym includes 11 environments spanning both classification and regression tasks. Four progressively harder image-based classification benchmarks (CircleSquare, MNIST, CIFAR-10, TinyImageNet) evaluate an agent's ability to select informative glimpses from natural or synthetic visuals. Two image search tasks (CIFAR10Search, TinyImageNetSearch) require the agent to infer the position of a small glimpse within a larger image. Finally, five non-visual regression tasks (LightDark and four LIDAR-based 2D localization environments) challenge agents to reduce state uncertainty by navigating procedurally generated maps or lighting fields. Crucially, in the self-localization environments, the agent influences the property it is trying to infer — its own position — through its actions. Hence, the prediction target changes over time in these environments, which is an explicitly supported aspect of ap_gym environments.

For the image-based classification and localization tasks, ap_gym relies on a mix of third-party assets: Tiny ImageNet [68], CIFAR-10 [67], and MNIST [6] datasets. The LIDARLoc (rooms and maze) environments generate map layouts procedurally and require no external data. Whenever applicable, ap_gym defines two versions of each environment, one for training with the training split of the respective dataset, and one for evaluation with the test split.

By abstracting away complex contact models and dynamics, ap_gym environments enable rapid prototyping of active perception strategies and provide a controlled baseline for more complex scenarios in the TMBS suite. However, despite their simplistic appearance, all ap_gym environments impose significant challenges due to partial observability and non-immediate action payoffs. The tasks differ substantially from each other, testing the algorithm's capability to handle diverse scenarios.

### 3.3 The Tactile MNIST Benchmark Suite

Tactile sensing presents unique challenges for perception: as an interactive modality, touch can unintentionally shift objects during exploration, and modern vision-based tactile sensors, such as GelSight [15] and DIGIT [17], produce inherently high-dimensional observations. At the same time, tactile sensing provides highly localized information confined to points of contact, necessitating active perception.

The Tactile MNIST Benchmark Suite (TMBS) extends ap_gym to tactile tasks using vision-based tactile sensors. It contains eight environments: *TactileMNIST*, *Starstruck*, and *PrimitiveGeometries* for classification, where the agent must classify a 3D model of a handwritten digit (see Section 3.4), count the number of stars among other shapes, or tell apart geometric primitives of random size; *Toolbox* and *ABCCenterOfMass*, where the agent must estimate the pose (or center of mass) of an object; and *TactileMNISTVolume*, *PrimitiveGeometriesVolume*, and *ABCVolume*, which task the agent with inferring the volume of a given shape (see Table 1 for an overview and Appendix F.3 for details).

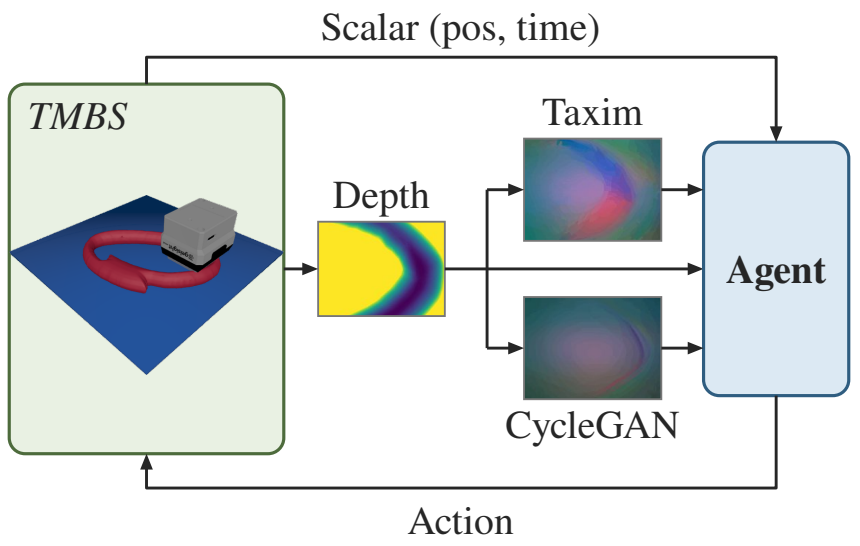

Figure 3: The agent receives the sensory information from the environment. This observation consists of scalar values and a tactile image, which can be rendered in one of three modes: depth, Taxim, or CycleGAN.

In each environment, the agent controls a simulated GelSight Mini tactile sensor [15] and is presented with one or more objects on a platform. By interacting with the object, the agent must infer task-dependent properties, such as the class of the object, its pose, or its volume. Particularly, aside from tactile data and proprioception, the agent does not receive any additional sensory data, so it has to infer the required property from touch alone. We do not simulate the physical interaction between the sensor and the object, as accurately simulating the deformation of the gel would require a soft-body simulation, which is either slow, inaccurate, or both, and a rigid-body simulation cannot model the

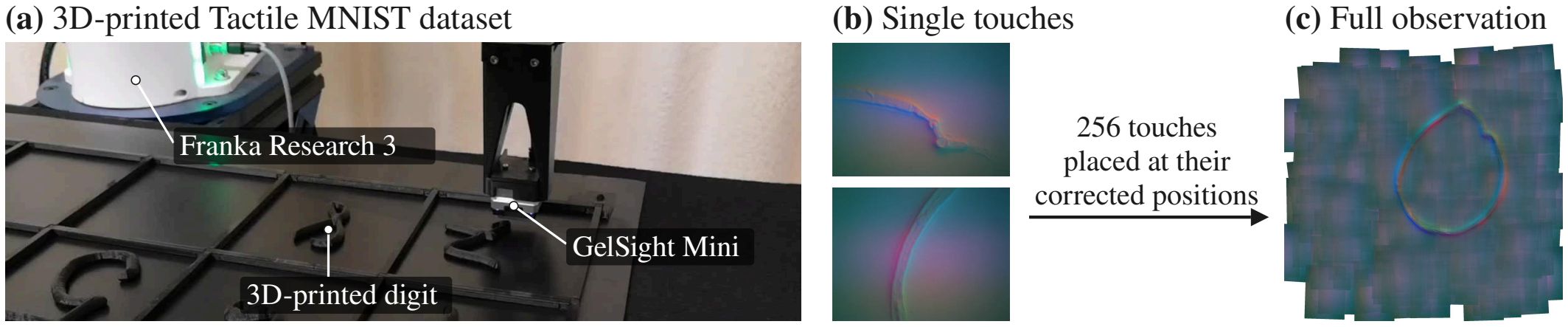

Figure 4: *Tactile MNIST Real Static* dataset collection. We mounted a GelSight Mini sensor on a Franka Research 3 (a) and collected 153,600 touches across 600 3D-printed MNIST digits. In (b), we show examples of individual touches, and in (c), all 256 touches of one digit overlaid based on their positions, which we corrected by image registration for the digit shifting and rotating in its cell during collection.

contact between the gel and the object well. Instead, we shift the objects around randomly to simulate unintended object movements. To simulate the tactile sensor, we support three rendering modes:

**Taxim:** Taxim [65] approximates the gel deformation and then runs a data-driven renderer.

**CycleGAN:** With data collected on 3D-printed MNIST 3D objects (see Section 3.5), we train a CycleGAN [16] for a style transfer between a depth image and a tactile image [72]. The resulting images are visually much more realistic than the Taxim renderings. In a proof-of-concept experiment, we show that both directions of the CycleGAN are key components of a recipe that transfers policies trained in simulation to real tactile data (see Appendix H). However, this mode is currently only available for the *TactileMNIST*-based environments. More details are in Appendix C.

**Depth:** Here, the agent receives a depth image clipped to the GelSight gel thickness (4.25mm).

A comparison of all rendering modes is shown in Fig. 3, a visualization of the TactileMNIST digit classification task is provided in Fig. 2, and an overview of all tasks in TMBS is given in Table 1.

### 3.4 The MNIST 3D Dataset

*MNIST 3D*[3] is a collection of 13,580 auto-generated 3D-printable meshes derived from a high-resolution MNIST variant [73] and scaled to fit in a 10x10cm square. The MNIST 3D dataset poses an exciting tactile classification challenge, as it has significant variability in shape and size within the classes, while also being large enough to facilitate learning from data. A single touch is rarely enough to classify objects from this dataset, as segments of handwritten digits are usually ambiguous. Hence, even after finding the digit, the agent has to apply some strategy (e.g., contour following) to gather enough information for a successful classification. In addition to tactile sensing, MNIST 3D could also be used as a benchmark for 3D mesh classification methods. More details are in Appendix D.

### 3.5 A Large Dataset of Real Tactile Interactions

To complement simulated renders, TMBS includes a real-world, static tactile dataset captured with a GelSight Mini sensor on 3D-printed MNIST 3D digits: the *Real Tactile MNIST Dataset*[4]. The dataset contains video sequences of 153,600 touches across 600 digits, which amounts to 256 touches per object collected in sequence. For data acquisition, we laid each 3D-printed MNIST digit in a 12x12cm grid on a rubber mat and used a Franka Research 3 robot arm [74] with a GelSight Mini tactile sensor to press the sensor down at random locations in the cell. Once we measured a normal force exceeding 5N, we stopped pressing and registered the timestamp. To prevent degradation of the elastomer gel, we replaced the GelSight sensor's gel pad after every 76,800 touches (i.e., halfway through the data collection). Finally, we partitioned the dataset into training (90 %) and test (10 %) splits, ensuring uniform class distributions across each split. Here, we also provide two processed versions of this dataset, where we replaced the videos with still images at the time of contact, one in full resolution at 320x240px[5] and one scaled to 64x64px[6] for faster loading and training.

[3] https://huggingface.co/datasets/TimSchneider42/tactile-mnist-mnist3d

[4] https://huggingface.co/datasets/TimSchneider42/tactile-mnist-touch-real-seq-t256-320x240

[5] https://huggingface.co/datasets/TimSchneider42/tactile-mnist-touch-real-single-t256-320x240

[6] https://huggingface.co/datasets/TimSchneider42/tactile-mnist-touch-real-single-t256-64x64

We note that an additional challenge introduced during data collection is the possibility of the digit shifting slightly due to contact with the sensor. This variability makes the dataset more representative of real-world scenarios and provides an opportunity for methods to learn robustness to object movement and misalignment. Thus, the Real Tactile MNIST Dataset serves several key roles in the TMBS benchmark. First, it enables training of a CycleGAN for a realistic simulation of tactile images and sim-to-real transfer, which we demonstrate in a proof-of-concept experiment in Appendix H. Second, it can be used for both pretraining and fine-tuning of learning-based perception models, enabling models to acquire basic tactile features before being deployed on a robot for active learning tasks and again facilitating sim-to-real transfer. Finally, via TactileMNISTRealSnap (Appendix F.3.3), the dataset provides a reproducible, offline benchmark for comparing active perception algorithms under realistic sensor noise and material artifacts. Appendix E details the data collection.

### 3.6 Evaluation Protocols

In this section, we describe the evaluation protocols used to assess agents in `ap_gym` and TMBS environments. Environment-specific metrics characterize performance within individual tasks, and the aggregated loss enables quantitative comparisons across environments.

**Environment-Specific Metrics** In `ap_gym` environments, there are two levels of exploration: (1) during an episode, the agent must explore to gather information, and (2) over the course of the training, the agent must explore the effects of its actions to optimize its model and policy. To disambiguate the measures of performance in these two levels, we will call the first one *exploration efficiency* and the second *sample efficiency*. Importantly, the former is a quality measure of the policy, while the latter is a quality measure of the learning algorithm. Here, we borrow the term *sample efficiency* from the RL literature, where it refers to the number of environment interactions the agent needs in order to learn to solve the given task. By *exploration efficiency*, on the other hand, we refer to the efficiency with which the agent collects information within an episode.

Regarding *exploration efficiency*, we consider two measures: the *average* prediction accuracy and the *final* prediction accuracy. Here, *average* prediction accuracy means the prediction accuracy the agent exhibited throughout an episode on average, while *final* prediction accuracy means the accuracy the agent exhibited at the final step of the episode. Normally, the agent starts each episode with little to no information and then keeps gathering information as the episode progresses. Hence, for a rational agent, we expect the prediction accuracy to increase over the course of the episode and to be highest at the end of the episode. Thus, the *average* prediction accuracy could be seen as a measure of how quickly the agent explored, while the *final* prediction accuracy could be seen as a measure of how thoroughly the agent explored throughout the episode. In `ap_gym` environments, both of these measures are tracked for a number of environment-specific prediction accuracy metrics, such as classification accuracy, mean squared error, pose error, and others. For a detailed list of metrics for each environment, refer to Appendix F.

Approaches evaluating on `ap_gym` or the Tactile MNIST Benchmark Suite should report both *average* and *final* metric values over the course of the training. If applicable, the metrics should be computed on the test variants of the environments, which use the test split instead of the training split. The objective is to maximize both *sample efficiency* and *exploration efficiency*. Section 4 serves as an example of an evaluation report on `ap_gym` and Tactile MNIST environments.

**Global Aggregated Loss** To enable quantitative comparisons across the diverse set of tasks in `ap_gym` and TMBS, we define an aggregated performance measure based on a normalized prediction loss. This measure allows us to compare classification and regression tasks on a unified scale.

In all environments, the prediction loss function $\ell(y^*, y)$ is normalized to lie between 0 and 1. For the losses used in this work (cross-entropy loss, mean squared error, and Huber loss), a lower bound of 0 is naturally given. To ensure an upper bound of 1, we scale the loss such that a "blind" baseline, representing an agent that consistently guesses the mean or uniform prior, yields a value of 1. Specifically, the normalized loss $\bar{\ell}$ is defined as $\bar{\ell}(y^*, y) = \frac{\ell(y^*, y)}{\ell_{\text{blind}}}$. For classification tasks using cross-entropy, $\ell_{\text{blind}} = \log(N)$, where $N$ is the number of classes. For regression tasks using MSE, $\ell_{\text{blind}}$ is the mean variance of the target data, calculated either from the dataset or from the platform's spatial bounds (e.g., for pose estimation). As the object volumes are heavy-tailed, volume estimation tasks use the Huber loss on the standardized volume instead, with $\ell_{\text{blind}}$ being the loss of the best constant prediction on the training objects (see Appendix F.3).

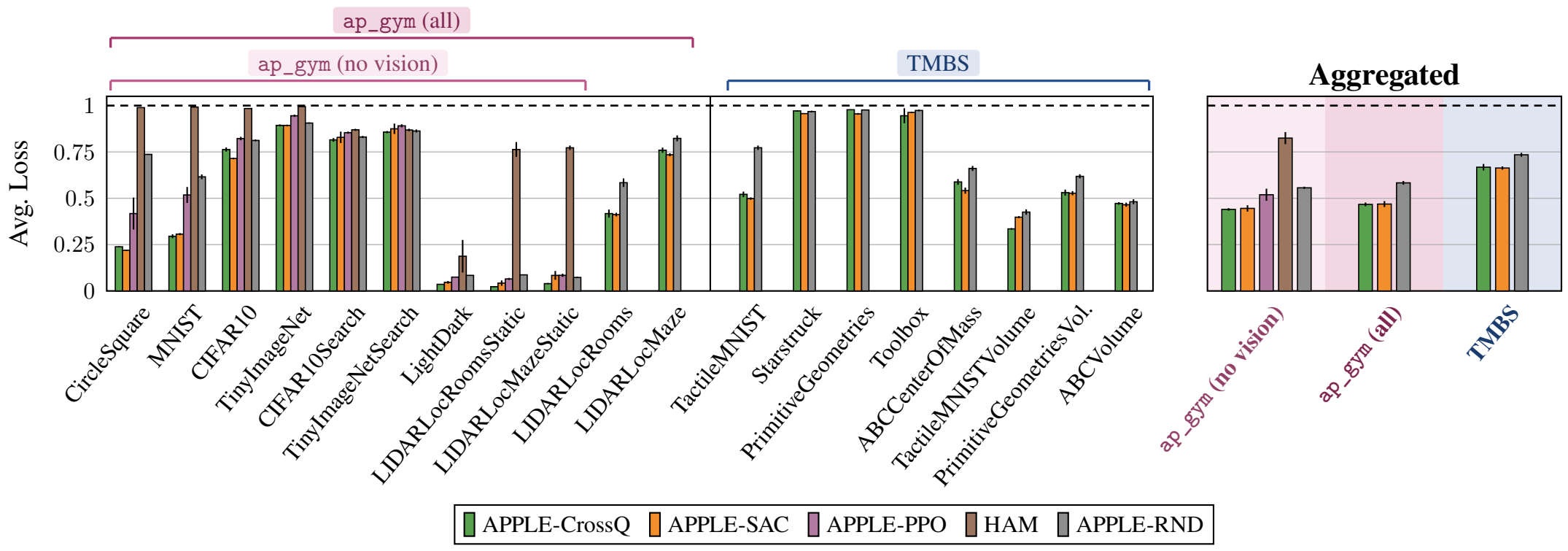

Figure 5: Average prediction loss of the baseline methods `APPLE-CrossQ`, `APPLE-SAC`, `APPLE-PPO`, `HAM` [44], and a random baseline `APPLE-RND` on every environment of `ap_gym` and TMBS (left) and aggregated per suite (right). Bars represent the mean of per-step prediction losses over the course of the training, normalized so that blind guessing yields 1, with the loss of each episode clipped at 1; lower values indicate better performance. Error bars denote the standard deviation across 5 seeds per environment. Prediction loss is computed on evaluation tasks with unseen objects where applicable. The brackets mark the environments behind each aggregate: all of `ap_gym`, `ap_gym` without the two vision-based tasks (LIDARLocRooms and LIDARLocMaze, which require a vision encoder to process the map), and TMBS. As `HAM` has no vision encoder and `APPLE-PPO` is unstable when used with vision, we evaluate these only on the `ap_gym` environments that do not require a vision encoder. Detailed learning curves are given in Appendix G.

Because environments in our suite have different step limits, we report the average normalized prediction loss per step over the duration of an episode. To obtain a single scalar value for a training run, we take the mean of these values across all steps in the training process. For a well-performing agent, this average loss resides in the interval $[0, 1]$, as a value higher than 1 would be worse than blind guessing. To ensure robustness against outliers and prevent a single poorly performing environment from dominating the aggregate, we clip the average normalized loss of each episode at a maximum value of 1 before aggregating. The clipping ensures that an agent performing worse than the blind baseline is penalized equally, regardless of the specific loss magnitude. To then obtain a per-task loss, we compute the mean and variance of the training run losses for multiple seeds. Finally, to aggregate across multiple tasks, we compute the mean and standard deviation (i.e., the square root of the mean variance across tasks) of the per-task losses, yielding a single global loss and confidence per method. An example of these aggregated losses can be viewed in Fig. 5.

# 4 Experiments

In this section, we evaluate various baseline methods on all environments of `ap_gym` and TMBS, including `APPLE` [45] and `HAM` [44]. Both `APPLE` and `HAM` are RL-based active perception methods and, thus, well suited for evaluation on TMBS. The main difference between them is that `APPLE` employs an actor-critic RL approach in combination with a transformer architecture, while `HAM` relies on REINFORCE and an LSTM model. `APPLE` provides two variants: `APPLE-SAC` and `APPLE-CrossQ`, based on `SAC` [75] and `CrossQ` [76]. We also evaluate `APPLE-PPO` which uses `APPLE`'s transformer model with PPO [77].

Fig. 5 shows the performance of all baselines on every environment and aggregated per suite; the corresponding learning curves are given in Appendix G. The environments range from simple entry points to largely unsolved problems. Among the classification tasks, in the CircleSquare environment, the agent can move a glimpse around an image and has to find and classify an object that can be either a circle or a square. The TactileMNIST environment tasks the agent with classifying MNIST 3D models by touch alone, and in the Starstruck environment, the agent must count the number of stars (1–3) among other objects on the platform. In TinyImageNet, the agent must classify natural images into 200 categories from small glimpses. Among the regression tasks, in the Toolbox environment, the agent must find a tool and determine its 2D pose and orientation on the platform. In LightDark and LIDARLocMaze, the agent has to localize itself: in LightDark, its position observations are only accurate close to a light source, while in LIDARLocMaze, it must match ambiguous LIDAR readings against a procedurally generated maze that changes in every episode. These environments represent a diverse set of challenges, and solving them requires the agent to adopt efficient exploration strategies, which is made evident by the consistently poor performance of `APPLE`'s random baseline `APPLE-RND` on most of them. Further experiments and details for the training can be found in Appendix G.

Together, the average and final performance in the learning curves of Appendix G show how efficiently the agents explore (see Section 3.6). The final performance measures what an agent ultimately infers, while the average performance additionally rewards agents that reach good predictions early in the episode. Hence, a small gap between the two indicates that the agent gathers the task-relevant information quickly. Fig. 28 shows how quickly the agents gather information in more detail for TactileMNIST, where we plot the prediction performance against the number of touches.

As visible in Fig. 5, and in more detail in Appendix G, with the exception of `APPLE-SAC` on LIDARLocMazeStatic, where it is on par with `APPLE-PPO`, `APPLE`'s off-policy methods perform consistently better than the on-policy baselines `HAM` and `APPLE-PPO`. This gap is likely due to off-policy methods being generally more sample-efficient than on-policy methods, since on-policy methods cannot reuse previously collected samples. Despite the better performance of `APPLE`, neither the `ap_gym` tasks nor the TMBS can be considered solved. In particular, on Starstruck and PrimitiveGeometries, no method clearly outperforms `APPLE-RND` (e.g., Starstruck: 35% vs. 37% final accuracy, with three classes); on TinyImageNet, CIFAR10Search, and TinyImageNetSearch, the best method lowers the average loss by at most 0.05 compared to `APPLE-RND` and Toolbox remains far from solved (3.6 cm and 74° final error of the best method, `APPLE-CrossQ`). The (near-)solved tasks, such as CircleSquare, MNIST, and LightDark, are deliberately simple entry points and sanity checks. In a proof of concept (Appendix H), `APPLE-SAC` policies trained in simulation transfer to real tactile data (67% vs. 80% final accuracy in simulation). All baselines require millions of environment interactions to learn viable exploration policies and on most environments fall short of perfect accuracy. These results show that more research in RL and active perception is needed before such methods can be deployed in the real world, and we hope that our benchmark helps drive this research.

# 5 Limitations and Conclusion

In this paper, we have introduced *Active Perception Gym* (`ap_gym`), a Gymnasium-compatible benchmark suite tailored for evaluating active perception tasks, and the *Tactile MNIST Benchmark Suite* (TMBS), which contains eight tactile-specific tasks designed for robust exploration. To support these benchmarks, we have released a dataset comprising 13,580 high-resolution 3D digit models and an extensive real-world dataset of 153,600 tactile samples collected from 600 3D-printed digits using a GelSight Mini sensor. Provided as an open-source framework, `ap_gym` and TMBS offer a structured, standardized, and reproducible benchmark intended to facilitate advancements in active perception research, including efficient exploration strategies, sim-to-real adaptation through CycleGAN training (see Appendix H for a proof of concept), and the pretraining and fine-tuning of tactile models.

However, our benchmark has limitations, most notably the absence of online, real-world evaluation scenarios beyond the static dataset. For now, our suite deliberately focuses on controlled, simulation-based exploration to ensure reproducibility and interoperability. However, in future work, we aim to address this limitation by designing carefully structured real-world tactile exploration experiments that extend the benchmark to real robotic systems. As an intermediate step, we will develop new tasks focused on robotic manipulation using multi-fingered hands in a physical simulator, opening the door to richer interaction scenarios and significantly broadening the benchmark's scope.

Ultimately, we view this benchmark as a foundational resource not only for advancing efficient robotic tactile manipulation but also for sparking further research on active perception in general. Active perception is not limited to tactile sensing; it is equally relevant to vision, agentic information retrieval, and long-horizon reasoning. Researching general active perception algorithms represents a necessary transition from passive pattern recognition to agency-driven intelligence, pushing the frontier of AI capabilities. Because purposeful information gathering remains a fundamental hurdle across machine learning, these domain-agnostic methodologies will advance AI across diverse fields. Addressing this crucial bottleneck provides the essential blueprint for agents that actively seek the information required to navigate and solve the unstructured reality we inhabit.

**Acknowledgements**

This work was supported by the German Federal Ministry of Research, Technology and Space (BMFTR) and the French Research Agency, l'Agence Nationale de Recherche (ANR), through the projects *Aristotle* (ANR-21-FAI1-0009-01), *Chiron* (ANR-20-IADJ-0001-01), and *Astérix* (ANR-23-EDIA-0002), by the EU's Horizon Europe project ARISE (Grant no.: 101135959), by the BMFTR's projects *Robotics Institute Germany* (RIG, 16ME1001) and *DEMETER* (01DR25003), and by the French national investment priority program's PSPC project *FAIR WASTE*. This work is also partly supported by the project *Genius Robot* (01IS24083) funded by the BMFTR, by the German Research Foundation (DFG, Deutsche Forschungsgemeinschaft) as part of Germany's Excellence Strategy – EXC 2050/1 – Project ID 390696704 – Cluster of Excellence *Centre for Tactile Internet with Human-in-the-Loop* (CeTI) of Technische Universität Dresden, and by the German Academic Exchange Service (DAAD) in project 57616814 (SECAI School of Embedded and Composite AI). The computations were conducted on the IAS Compute Cluster and on the hessian.AI cluster forty.THREE (Cluster 43). We gratefully acknowledge support from the hessian.AI Service Center (funded by the German Federal Ministry of Research, Technology and Space, BMFTR, grant no. 16IS22091) and the hessian.AI Innovation Lab (funded by the Hessian Ministry for Digital Strategy and Innovation, grant no. S-DIW04/0013/003). We acknowledge the EuroHPC Joint Undertaking for awarding this project access to the EuroHPC supercomputers LUMI, hosted by CSC – IT Center for Science (Finland) and the LUMI consortium, MareNostrum 5, hosted by the Barcelona Supercomputing Center (BSC-CNS, Spain), and Deucalion, hosted by the Minho Advanced Computing Centre (MACC, Portugal), through a EuroHPC Benchmark Access call (EHPC-BEN-2026B11-001) and a EuroHPC Development Access call (EHPC-DEV-2026D11-001).

## NeurIPS Paper Checklist

1. **Claims**

   Question: Do the main claims made in the abstract and introduction accurately reflect the paper's contributions and scope?

   Answer: [Yes]

   Justification: The main claim of the paper is the introduction of the first benchmarking protocol for active tactile perception. Section 2 supports this claim with a comprehensive literature review. Furthermore, our two benchmark suites and the datasets are introduced in Section 3, with additional details in the appendix, and in the code documentation.

   Guidelines:

   - The answer [N/A] means that the abstract and introduction do not include the claims made in the paper.
   - The abstract and/or introduction should clearly state the claims made, including the contributions made in the paper and important assumptions and limitations. A [No] or [N/A] answer to this question will not be perceived well by the reviewers.
   - The claims made should match theoretical and experimental results, and reflect how much the results can be expected to generalize to other settings.
   - It is fine to include aspirational goals as motivation as long as it is clear that these goals are not attained by the paper.

2. **Limitations**

   Question: Does the paper discuss the limitations of the work performed by the authors?

   Answer: [Yes]

   Justification: In Section 5, we discuss limitations and future directions of the work.

   Guidelines:

   - The answer [N/A] means that the paper has no limitation while the answer [No] means that the paper has limitations, but those are not discussed in the paper.
   - The authors are encouraged to create a separate "Limitations" section in their paper.
   - The paper should point out any strong assumptions and how robust the results are to violations of these assumptions (e.g., independence assumptions, noiseless settings, model well-specification, asymptotic approximations only holding locally). The authors should reflect on how these assumptions might be violated in practice and what the implications would be.
   - The authors should reflect on the scope of the claims made, e.g., if the approach was only tested on a few datasets or with a few runs. In general, empirical results often depend on implicit assumptions, which should be articulated.
   - The authors should reflect on the factors that influence the performance of the approach. For example, a facial recognition algorithm may perform poorly when image resolution is low or images are taken in low lighting. Or a speech-to-text system might not be used reliably to provide closed captions for online lectures because it fails to handle technical jargon.
   - The authors should discuss the computational efficiency of the proposed algorithms and how they scale with dataset size.
   - If applicable, the authors should discuss possible limitations of their approach to address problems of privacy and fairness.
   - While the authors might fear that complete honesty about limitations might be used by reviewers as grounds for rejection, a worse outcome might be that reviewers discover limitations that aren't acknowledged in the paper. The authors should use their best judgment and recognize that individual actions in favor of transparency play an important role in developing norms that preserve the integrity of the community. Reviewers will be specifically instructed to not penalize honesty concerning limitations.

3. **Theory assumptions and proofs**

Question: For each theoretical result, does the paper provide the full set of assumptions and a complete (and correct) proof?

Answer: [N/A]

Justification: [N/A]

Guidelines:

- The answer [N/A] means that the paper does not include theoretical results.
- All the theorems, formulas, and proofs in the paper should be numbered and cross-referenced.
- All assumptions should be clearly stated or referenced in the statement of any theorems.
- The proofs can either appear in the main paper or the supplemental material, but if they appear in the supplemental material, the authors are encouraged to provide a short proof sketch to provide intuition.
- Inversely, any informal proof provided in the core of the paper should be complemented by formal proofs provided in appendix or supplemental material.
- Theorems and Lemmas that the proof relies upon should be properly referenced.

4. **Experimental result reproducibility**

Question: Does the paper fully disclose all the information needed to reproduce the main experimental results of the paper to the extent that it affects the main claims and/or conclusions of the paper (regardless of whether the code and data are provided or not)?

Answer: [Yes]

Justification: Throughout Section 3, we present our benchmark suites. Throughout the appendix, and particularly in Appendix F and Appendix I, we present detailed descriptions of how each of our environments works as well as implementation details. We add that our code is also available with detailed documentation. Furthermore, for the dataset collection, we fully describe the steps taken to generate the MNIST 3D meshes (in Section 3.4 and Appendix D) and the real tactile data collection (in Section 3.5 and Appendix E).

Guidelines:

- The answer [N/A] means that the paper does not include experiments.
- If the paper includes experiments, a [No] answer to this question will not be perceived well by the reviewers: Making the paper reproducible is important, regardless of whether the code and data are provided or not.
- If the contribution is a dataset and/or model, the authors should describe the steps taken to make their results reproducible or verifiable.
- Depending on the contribution, reproducibility can be accomplished in various ways. For example, if the contribution is a novel architecture, describing the architecture fully might suffice, or if the contribution is a specific model and empirical evaluation, it may be necessary to either make it possible for others to replicate the model with the same dataset, or provide access to the model. In general. releasing code and data is often one good way to accomplish this, but reproducibility can also be provided via detailed instructions for how to replicate the results, access to a hosted model (e.g., in the case of a large language model), releasing of a model checkpoint, or other means that are appropriate to the research performed.
- While NeurIPS does not require releasing code, the conference does require all submissions to provide some reasonable avenue for reproducibility, which may depend on the nature of the contribution. For example
  (a) If the contribution is primarily a new algorithm, the paper should make it clear how to reproduce that algorithm.
  (b) If the contribution is primarily a new model architecture, the paper should describe the architecture clearly and fully.
  (c) If the contribution is a new model (e.g., a large language model), then there should either be a way to access this model for reproducing the results or a way to reproduce the model (e.g., with an open-source dataset or instructions for how to construct the dataset).

(d) We recognize that reproducibility may be tricky in some cases, in which case authors are welcome to describe the particular way they provide for reproducibility. In the case of closed-source models, it may be that access to the model is limited in some way (e.g., to registered users), but it should be possible for other researchers to have some path to reproducing or verifying the results.

5. **Open access to data and code**

   Question: Does the paper provide open access to the data and code, with sufficient instructions to faithfully reproduce the main experimental results, as described in supplemental material?

   Answer: [Yes]

   Justification: We provide all the code for our benchmark environments with extensive documentation.

   Guidelines:

   - The answer [N/A] means that paper does not include experiments requiring code.
   - Please see the NeurIPS code and data submission guidelines (`https://neurips.cc/public/guides/CodeSubmissionPolicy`) for more details.
   - While we encourage the release of code and data, we understand that this might not be possible, so [No] is an acceptable answer. Papers cannot be rejected simply for not including code, unless this is central to the contribution (e.g., for a new open-source benchmark).
   - The instructions should contain the exact command and environment needed to run to reproduce the results. See the NeurIPS code and data submission guidelines (`https://neurips.cc/public/guides/CodeSubmissionPolicy`) for more details.
   - The authors should provide instructions on data access and preparation, including how to access the raw data, preprocessed data, intermediate data, and generated data, etc.
   - The authors should provide scripts to reproduce all experimental results for the new proposed method and baselines. If only a subset of experiments are reproducible, they should state which ones are omitted from the script and why.
   - At submission time, to preserve anonymity, the authors should release anonymized versions (if applicable).
   - Providing as much information as possible in supplemental material (appended to the paper) is recommended, but including URLs to data and code is permitted.

6. **Experimental setting/details**

   Question: Does the paper specify all the training and test details (e.g., data splits, hyperparameters, how they were chosen, type of optimizer) necessary to understand the results?

   Answer: [Yes]

   Justification: We provide details in Section 4, Appendix G, and Appendix I.

   Guidelines:

   - The answer [N/A] means that the paper does not include experiments.
   - The experimental setting should be presented in the core of the paper to a level of detail that is necessary to appreciate the results and make sense of them.
   - The full details can be provided either with the code, in appendix, or as supplemental material.

7. **Experiment statistical significance**

   Question: Does the paper report error bars suitably and correctly defined or other appropriate information about the statistical significance of the experiments?

   Answer: [Yes]

   Justification: Our results in Section 4 and Appendix G show one standard deviation for each learning curve.

   Guidelines:

- The answer [N/A] means that the paper does not include experiments.
- The authors should answer [Yes] if the results are accompanied by error bars, confidence intervals, or statistical significance tests, at least for the experiments that support the main claims of the paper.
- The factors of variability that the error bars are capturing should be clearly stated (for example, train/test split, initialization, random drawing of some parameter, or overall run with given experimental conditions).
- The method for calculating the error bars should be explained (closed form formula, call to a library function, bootstrap, etc.)
- The assumptions made should be given (e.g., Normally distributed errors).
- It should be clear whether the error bar is the standard deviation or the standard error of the mean.
- It is OK to report 1-sigma error bars, but one should state it. The authors should preferably report a 2-sigma error bar than state that they have a 96% CI, if the hypothesis of Normality of errors is not verified.
- For asymmetric distributions, the authors should be careful not to show in tables or figures symmetric error bars that would yield results that are out of range (e.g., negative error rates).
- If error bars are reported in tables or plots, the authors should explain in the text how they were calculated and reference the corresponding figures or tables in the text.

8. **Experiments compute resources**

   Question: For each experiment, does the paper provide sufficient information on the computer resources (type of compute workers, memory, time of execution) needed to reproduce the experiments?

   Answer: [Yes]

   Justification: All implementation and performance details for the experiments are present in Section 4, Appendix G, and Appendix I.

   Guidelines:

   - The answer [N/A] means that the paper does not include experiments.
   - The paper should indicate the type of compute workers CPU or GPU, internal cluster, or cloud provider, including relevant memory and storage.
   - The paper should provide the amount of compute required for each of the individual experimental runs as well as estimate the total compute.
   - The paper should disclose whether the full research project required more compute than the experiments reported in the paper (e.g., preliminary or failed experiments that didn't make it into the paper).

9. **Code of ethics**

   Question: Does the research conducted in the paper conform, in every respect, with the NeurIPS Code of Ethics `https://neurips.cc/public/EthicsGuidelines`?

   Answer: [Yes]

   Justification: Our research fully conforms to the NeurIPS Code of Ethics. No human subjects or participants were involved in the creation of the datasets used, so there are no related privacy or consent concerns. All data and code are released publicly on GitHub and Hugging Face under permissive licenses (MIT for code, CC-BY for data), ensuring long-term accessibility, transparency, and reproducibility. We have verified that none of the datasets used are deprecated and that they comply with licensing terms. Our work poses no foreseeable risks related to safety, security, discrimination, or environmental harm.

   Guidelines:

   - The answer [N/A] means that the authors have not reviewed the NeurIPS Code of Ethics.
   - If the authors answer [No], they should explain the special circumstances that require a deviation from the Code of Ethics.

- The authors should make sure to preserve anonymity (e.g., if there is a special consideration due to laws or regulations in their jurisdiction).

10. **Broader impacts**

    Question: Does the paper discuss both potential positive societal impacts and negative societal impacts of the work performed?

    Answer: [N/A]

    Justification: This work introduces a benchmark for active tactile perception in robotic systems. It is foundational in nature and not tied to any direct applications or deployments. While it may contribute to progress in robotic manipulation over the long term, we do not foresee any immediate societal impact from this work.

    Guidelines:

    - The answer [N/A] means that there is no societal impact of the work performed.
    - If the authors answer [N/A] or [No], they should explain why their work has no societal impact or why the paper does not address societal impact.
    - Examples of negative societal impacts include potential malicious or unintended uses (e.g., disinformation, generating fake profiles, surveillance), fairness considerations (e.g., deployment of technologies that could make decisions that unfairly impact specific groups), privacy considerations, and security considerations.
    - The conference expects that many papers will be foundational research and not tied to particular applications, let alone deployments. However, if there is a direct path to any negative applications, the authors should point it out. For example, it is legitimate to point out that an improvement in the quality of generative models could be used to generate Deepfakes for disinformation. On the other hand, it is not needed to point out that a generic algorithm for optimizing neural networks could enable people to train models that generate Deepfakes faster.
    - The authors should consider possible harms that could arise when the technology is being used as intended and functioning correctly, harms that could arise when the technology is being used as intended but gives incorrect results, and harms following from (intentional or unintentional) misuse of the technology.
    - If there are negative societal impacts, the authors could also discuss possible mitigation strategies (e.g., gated release of models, providing defenses in addition to attacks, mechanisms for monitoring misuse, mechanisms to monitor how a system learns from feedback over time, improving the efficiency and accessibility of ML).

11. **Safeguards**

    Question: Does the paper describe safeguards that have been put in place for responsible release of data or models that have a high risk for misuse (e.g., pre-trained language models, image generators, or scraped datasets)?

    Answer: [N/A]

    Justification: None of our data or benchmark suites has any significant risk of misuse.

    Guidelines:

    - The answer [N/A] means that the paper poses no such risks.
    - Released models that have a high risk for misuse or dual-use should be released with necessary safeguards to allow for controlled use of the model, for example by requiring that users adhere to usage guidelines or restrictions to access the model or implementing safety filters.
    - Datasets that have been scraped from the Internet could pose safety risks. The authors should describe how they avoided releasing unsafe images.
    - We recognize that providing effective safeguards is challenging, and many papers do not require this, but we encourage authors to take this into account and make a best faith effort.

12. **Licenses for existing assets**

Question: Are the creators or original owners of assets (e.g., code, data, models), used in the paper, properly credited and are the license and terms of use explicitly mentioned and properly respected?

Answer: [Yes]

Justification: All external assets used in this work are properly cited, and their respective licenses are respected. We list the primary datasets used for our benchmarks in Table 1, including MNIST [6], CIFAR-10 [67], and Tiny ImageNet [68]. For our tactile benchmarks, we use the ABC Dataset [69]. To ensure reproducibility and ease of access, we host a version of this dataset on Hugging Face (available at `https://huggingface.co/datasets/TimSchneider42/tactile-mnist-abc-dataset`). We credit the original authors and specify that the individual 3D models are provided under the Onshape Terms of Use (Section 1.g.ii), which allows for the reproduction and distribution of public content. Our custom 3D models (MNIST 3D and Starstruck) and the Real Tactile MNIST Dataset are released under a CC-BY 4.0 license.

Guidelines:

- The answer [N/A] means that the paper does not use existing assets.
- The authors should cite the original paper that produced the code package or dataset.
- The authors should state which version of the asset is used and, if possible, include a URL.
- The name of the license (e.g., CC-BY 4.0) should be included for each asset.
- For scraped data from a particular source (e.g., website), the copyright and terms of service of that source should be provided.
- If assets are released, the license, copyright information, and terms of use in the package should be provided. For popular datasets, `paperswithcode.com/datasets` has curated licenses for some datasets. Their licensing guide can help determine the license of a dataset.
- For existing datasets that are re-packaged, both the original license and the license of the derived asset (if it has changed) should be provided.
- If this information is not available online, the authors are encouraged to reach out to the asset's creators.

13. **New assets**

Question: Are new assets introduced in the paper well documented and is the documentation provided alongside the assets?

Answer: [Yes]

Justification: All the code, models, and datasets we have released come with detailed documentation. Additionally, Section 3, Appendix F, Appendix D, Appendix E, and Appendix I provide details of the released assets.

Guidelines:

- The answer [N/A] means that the paper does not release new assets.
- Researchers should communicate the details of the dataset/code/model as part of their submissions via structured templates. This includes details about training, license, limitations, etc.
- The paper should discuss whether and how consent was obtained from people whose asset is used.
- At submission time, remember to anonymize your assets (if applicable). You can either create an anonymized URL or include an anonymized zip file.

14. **Crowdsourcing and research with human subjects**

Question: For crowdsourcing experiments and research with human subjects, does the paper include the full text of instructions given to participants and screenshots, if applicable, as well as details about compensation (if any)?

Answer: [N/A]

Justification: The paper does not involve crowdsourcing or research with human subjects.

Guidelines:

- The answer [N/A] means that the paper does not involve crowdsourcing nor research with human subjects.
- Including this information in the supplemental material is fine, but if the main contribution of the paper involves human subjects, then as much detail as possible should be included in the main paper.
- According to the NeurIPS Code of Ethics, workers involved in data collection, curation, or other labor should be paid at least the minimum wage in the country of the data collector.

15. **Institutional review board (IRB) approvals or equivalent for research with human subjects**

    Question: Does the paper describe potential risks incurred by study participants, whether such risks were disclosed to the subjects, and whether Institutional Review Board (IRB) approvals (or an equivalent approval/review based on the requirements of your country or institution) were obtained?

    Answer: [N/A]

    Justification: The paper does not involve crowdsourcing or research with human subjects.

    Guidelines:

    - The answer [N/A] means that the paper does not involve crowdsourcing nor research with human subjects.
    - Depending on the country in which research is conducted, IRB approval (or equivalent) may be required for any human subjects research. If you obtained IRB approval, you should clearly state this in the paper.
    - We recognize that the procedures for this may vary significantly between institutions and locations, and we expect authors to adhere to the NeurIPS Code of Ethics and the guidelines for their institution.
    - For initial submissions, do not include any information that would break anonymity (if applicable), such as the institution conducting the review.

16. **Declaration of LLM usage**

    Question: Does the paper describe the usage of LLMs if it is an important, original, or non-standard component of the core methods in this research? Note that if the LLM is used only for writing, editing, or formatting purposes and does *not* impact the core methodology, scientific rigor, or originality of the research, declaration is not required.

    Answer: [N/A]

    Justification: This research does not involve LLMs as any important, original, or non-standard components.

    Guidelines:

    - The answer [N/A] means that the core method development in this research does not involve LLMs as any important, original, or non-standard components.
    - Please refer to our LLM policy in the NeurIPS handbook for what should or should not be described.

# Appendix

## Table of Contents

Table 2: Hyperparameters determined by the HEBO [82] Bayesian optimizer for `APPLE-SAC` and `APPLE-CrossQ`. The *no vision-encoder* configuration was trained on the CircleSquare environment, while the *vision-encoder* configuration was trained on the TactileMNIST environment. Hyperparameters prepended with *Rel.* are relative to the total number of steps throughout the training.

| **Hyperparameter** | `APPLE-SAC` | | `APPLE-CrossQ` | |
|---|---|---|---|---|
| | **no vis.-enc.** | **vis.-enc.** | **no vis.-enc.** | **vis.-enc.** |
| Optimizer type | ADAMW | ADAMW | ADAMW | ADAMW |
| Learning rate (actor) | $5 \cdot 10^{-5}$ | $5 \cdot 10^{-4}$ | $1 \cdot 10^{-5}$ | $3 \cdot 10^{-4}$ |
| Learning rate (critic) | $5 \cdot 10^{-4}$ | $5 \cdot 10^{-5}$ | $1 \cdot 10^{-4}$ | $6 \cdot 10^{-5}$ |
| LR schedule (both) | none | cosine decay | none | none |
| Rel. LR cosine warm-up (both) | N/A | 0.15 | N/A | N/A |
| Initial UTD | 0.75 | 0.25 | 5.0 | 0.25 |
| Final UTD | 4.0 | 1.5 | 0.5 | 3.5 |
| Rel. UTD warm-up | 0.9 | 0.4 | 0.3 | 0.45 |

# A Hyperparameter Optimization

To enable a fair comparison of the baseline approaches, we conduct extensive hyperparameter searches with the HEBO [82] Bayesian optimizer. We found that the optimal hyperparameters for each method may vary from environment to environment, in particular, if one environment requires a vision encoder and the other one does not. To address this issue, we optimize two different sets of hyperparameters, one for environments requiring a vision encoder and one for environments that do not, whenever applicable. For an overview of the input modalities of all environments, refer to Table 16.

From each of the two domains (with and without a vision encoder), we pick one environment to tune the hyperparameters on. In particular, we choose CircleSquare as an environment without a vision encoder and TactileMNIST as an environment with a vision encoder. To evaluate the performance of a set of hyperparameters, we run training for a single seed on the respective environment (250K steps on CircleSquare and 2.5M steps on TactileMNIST, except for `HAM` and `APPLE-PPO`, which we train for 1M steps on CircleSquare) and measure the episode returns averaged over the course of the training. Averaging instead of just taking the final returns makes sure that sample efficiency is taken into consideration when evaluating hyperparameters, as two configurations with the same final returns might have taken a significantly different number of environment interactions to reach that performance.

As Bayesian optimization suffers from the curse of dimensionality, we select a subset of hyperparameters for the hyperparameter search, which we found to be impactful for performance. Specifically, we tune the learning rate of all approaches, and, if applicable, the type of learning rate schedule (none, cosine decay, linear), the parameters of the learning rate schedule (number of warm-up steps), and the optimizer used (ADAM, ADAMW, and SGD). For `APPLE`'s off-policy methods, we additionally tune the update-to-data (UTD) schedule, that is, the initial and final UTD ratio, and the number of warm-up steps.

While we were unable to find parameters for `HAM` that result in meaningful performance, we gained some insights into the influence of these hyperparameters on `APPLE-SAC` and `APPLE-CrossQ`. While both `APPLE-SAC` and `APPLE-CrossQ` reached similar performance on the tasks they were tuned on, neither was consistently better on unseen environments, as shown in Section 4 and Appendix G. Since there was no significant difference in performance between `APPLE-CrossQ`'s vision-encoder hyperparameters and non-vision-encoder hyperparameters when applied to the CircleSquare task, we chose to proceed using the vision-encoder hyperparameters for all environments to keep the configuration simple.

The results of the hyperparameter search are shown in Table 2.

# B Illustration of Active Perception in the Toolbox Environment

A formal definition of the active perception process, framed as a special case of a POMDP, is presented in Section 3.1. An example of this process in our benchmark suite is shown in Fig. 6, using the Toolbox environment for illustration. At step 0 of each episode, the environment is initialized and the robot begins from a uniformly sampled random position. At every subsequent step, the agent receives an observation from the previous interaction. In this case, that is a tactile image from the GelSight sensor along with the corresponding sensor pose. The agent processes this input to both predict the desired property (e.g., the pose of a wrench) and select a new interaction point.

The robot is then moved to the selected location, where a new observation is acquired. If the episode has not yet terminated, the process repeats: the agent integrates information from the new data point, updates its internal belief, and chooses the next action. At the end of each step, relevant data (observations, predictions, actions) are stored for evaluation.

Over time, the agent learns an effective exploration strategy. In the case of the wrench, for instance, it may discover that interacting with the object's extremities helps to disambiguate its orientation. Once the episode concludes, we can analyze the stored data to evaluate estimation accuracy and the efficiency of the exploration strategy. For experimental results and benchmarks, refer to Section 4 and additional evaluations in Appendix G.

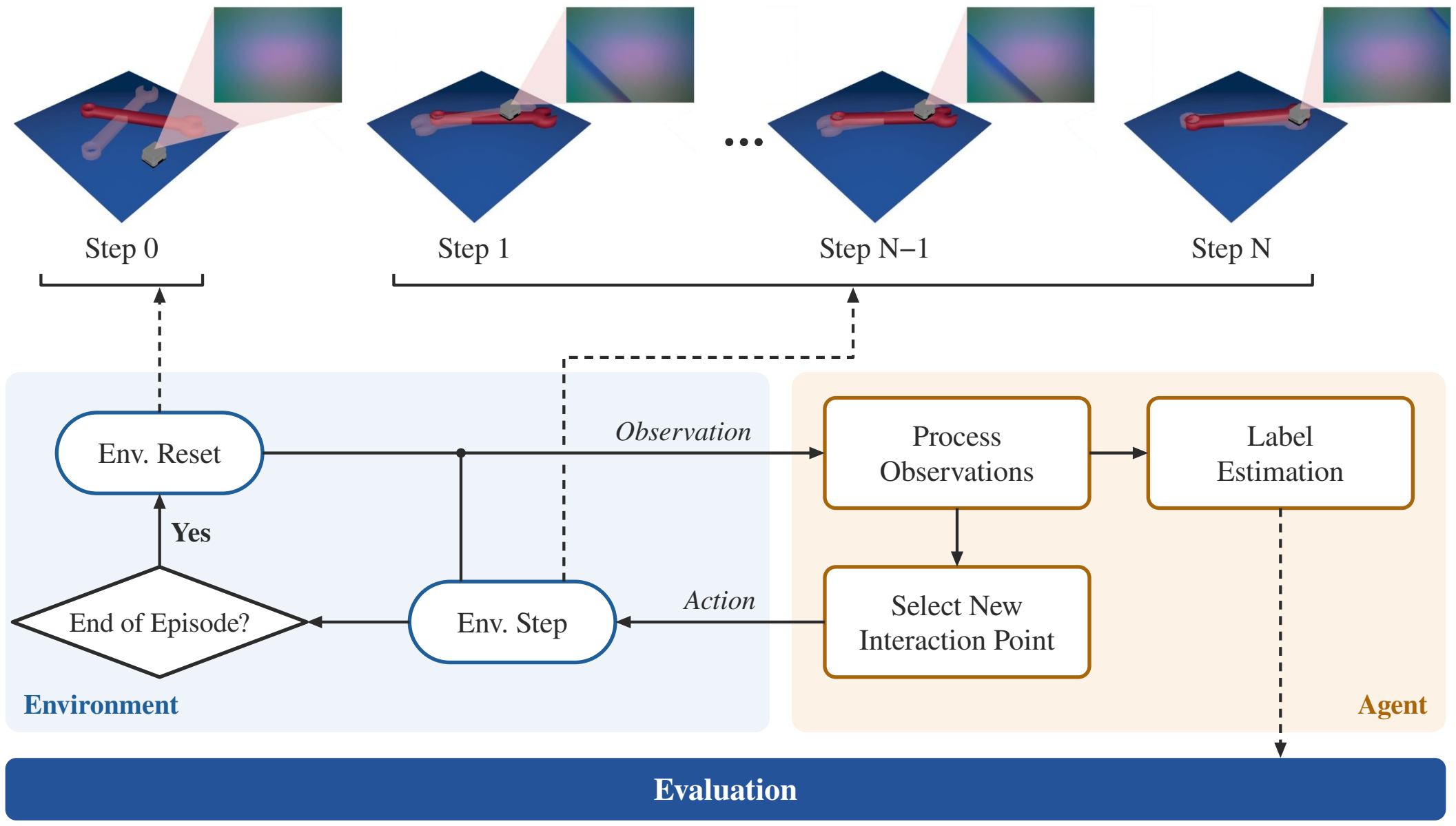

Figure 6: Active perception process in the Toolbox environment.

# C CycleGAN-based Tactile Rendering

The sim-to-real gap between synthetic and real-world images is a critical challenge for transferring trained algorithms to real-world applications. One approach to generating synthetic tactile images is Taxim [65], an example-based simulator that renders synthetic data from depth maps. However, disparities remain between Taxim's outputs and real tactile images. While Taxim produces smooth renderings, real tactile images exhibit finer surface textures (e.g., from 3D printing).

To enhance synthetic rendering, we leverage our dataset to train a data-driven image-to-image model. Previous work has demonstrated the effectiveness of such models in bridging the sim-to-real gap in visual tactile sensors. For instance, [83, 84] employed Pix2Pix [85] to convert synthetic depth maps into realistic images, but this method relies on paired datasets (synthetic-to-real image pairs), which can be expensive to acquire. In contrast, we employ CycleGAN, which has proven effective for tactile renderings [72] and offers two key advantages: first, it supports unpaired training, removing the need for precisely aligned synthetic and real images; second, it enables bidirectional translation, allowing both depth-to-real and real-to-depth synthesis, thereby increasing the model's flexibility and applicability.

We assembled the CycleGAN training dataset from 20,000 unaligned synthetic depth images and 20,000 real tactile images from our real-world dataset (see Section 3.5 and Appendix E for dataset details), reducing the share of non-contact images so that most images show a contact. To address the inherent instability of CycleGAN training, we applied several preprocessing steps. First, we normalized the depth maps: since synthetic depth images encode non-contact images with a default value of 0 (indicating no interaction), we adjusted them to 1 to ensure depth consistency across both contact and non-contact images. Additionally, we converted the 1-channel depth maps into 3-channel heightmaps for better alignment with GelSight RGB images, which are illuminated by side-mounted red, green, and blue LEDs that create a strong correlation between pixel color and spatial position. In this conversion, the first two channels represent the 2D pixel coordinates (x, y), while the third channel retains the original depth value (z). This enhanced representation preserves spatial context during training and enables random cropping—an operation that would otherwise disrupt positional alignment in standard 1-channel depth maps. Qualitative examples of the processed depth maps and CycleGAN results are shown in Fig. 7.

Beyond rendering, the CycleGAN helps transfer policies trained in simulation to real tactile data: in Appendix H, we use its depth-to-RGB direction to render realistic tactile images during training and close the remaining domain gap with a full round trip through both directions on the real images, which, combined with domain randomization, transfers policies trained in simulation to real tactile data.

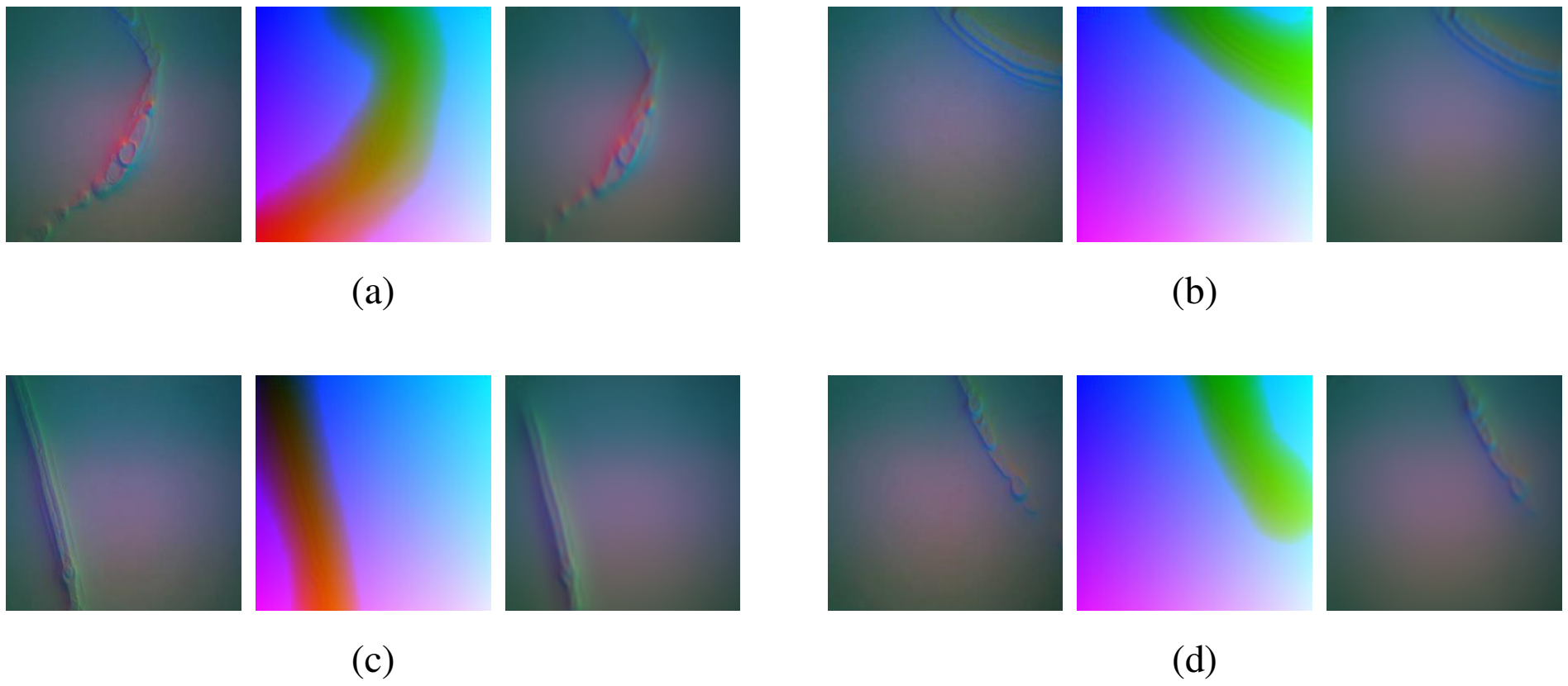

Figure 7: Qualitative results over four cycles illustrating the translation between depth images and RGB images. Each triplet in the figure shows, in order: real tactile (RGB) images, depth images generated using CycleGAN, and the RGB images reconstructed from the generated depth maps. These triplets demonstrate the consistency and reversibility of the RGB-to-depth and depth-to-RGB transformations.

## D 3D Mesh Dataset Generation Details

In this work, we introduce two 3D mesh datasets: MNIST 3D, which contains 3D models of handwritten MNIST digits, and the Starstruck dataset, which contains procedurally generated arrangements of geometric shapes used in the Starstruck environment. We detail their generation process in the following.

### D.1 MNIST 3D

**Algorithm 1** Generating 3D Meshes from High-Resolution MNIST Digits

**Require:** High-resolution MNIST digit image $I$
Initialize an empty list $S$
**for all** erosion size $k$ in increasing order **do**
    Apply binary erosion with kernel size $k$ to $I$, yielding $I_k$
    Append $I_k$ to list $S$
**end for**
Stack all $I_k$ in $S$ along the depth axis to form 3D tensor $T$
Mirror $T$ along the depth axis to obtain symmetric occupancy grid $T'$
Apply Marching Cubes to $T'$ to extract a surface mesh $M$
Smooth mesh $M$
Scale $M$ such that 5 pixels correspond to 1 mm
**return** Mesh $M$

As described in Section 3.4, the MNIST 3D dataset consists of 13,580 auto-generated, 3D-printable meshes derived from the high-resolution MNIST dataset [73]. The mesh generation process is detailed in Algorithm 1. In short, each digit image undergoes iterative binary erosion with increasing kernel sizes to create a series of thinner versions. These eroded images are then stacked into a 3D tensor and mirrored along the third axis to yield a symmetric 3D occupancy grid. Finally, we employ the marching cubes algorithm to convert this grid into a mesh, followed by surface smoothing to enhance mesh quality.

Meshes are scaled such that five pixels correspond to 1 mm, fitting each digit within a bounding box of approximately $10 \times 10$ cm. In practice, most digit models occupy a smaller volume, as only a few MNIST characters span the full image area. An illustrative subset of the MNIST 3D dataset is shown in Fig. 8.

We split MNIST 3D into five splits, detailed in Table 3.

### D.2 Starstruck

In addition to the MNIST digits, we also introduce a dataset of geometric shapes designed for the Starstruck environment. As shown in Fig. 19, each scene may contain squares, circles, and up to three stars. Here, each data point corresponds to one arrangement of these geometric shapes.

To generate data points, we first choose the number of distractors (circles and squares) randomly between zero and five. We then determine the type of each distractor by randomly choosing between a circle and a square. Afterward, the distractors and stars are placed on the platform randomly. Beginning with the stars, we place the objects sequentially. For every object, we first sample a uniformly random location on the plate. Whenever a newly placed object would intersect a previously placed object, we sample a new position for the newly placed object. Should this process fail to find a suitable location after 100 attempts, we discard all prior placements and start from scratch.

Instead of storing a fixed set of arrangements, we generate the Starstruck dataset procedurally on the fly. Arrangement $i$ contains $1 + (i \bmod 3)$ stars, so every split is balanced across the three classes (one, two, and three stars), and its distractors and layout are drawn from a pseudo-random number generator seeded with the split and $i$. Hence, every arrangement can be reproduced from its index alone, and the dataset can be made arbitrarily large at no storage cost. In practice, both the training and the test split contain 300,000 arrangements (100,000 per class), which prevents the agent from memorizing them.

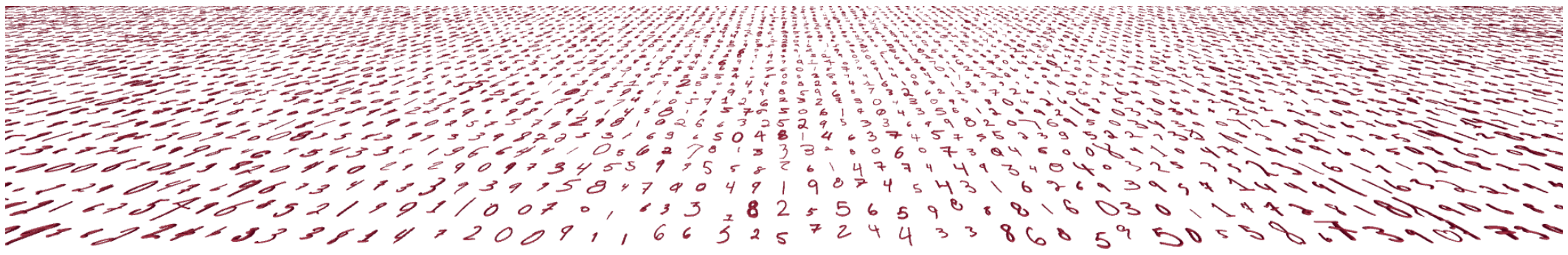

Figure 8: The *MNIST 3D* dataset comprises 13,580 3D models of handwritten digits. Each model is auto-generated from a high-resolution MNIST variant [73].

Table 4: Data point structure for the Real Tactile MNIST Dataset.

| Field | Description | Type / Shape | Seq | Single |
|---|---|---|---|---|
| **Identification** | | | | |
| `id` | Unique identifier | `int` | ✓ | ✓ |
| `label` | Digit label (0–9) | `int` | ✓ | ✓ |
| `object_id` | Original MNIST image ID | `int` | ✓ | ✓ |
| `info` | Metadata (e.g., gel type, notes) | `dict` | ✓ | ✓ |
| **Spatial (Cell Frame)** | | | | |
| `pos_in_cell` | Intended 2D contact positions | $T \times 2$ `np.ndarray` | ✓ | ✓ |
| `gel_pose_cell_frame_seq` | Full 3D pose per video frame | list[$T$] of list[$N_i$] of `Transformation` | ✓ | — |
| `gel_pose_cell_frame` | 3D pose at peak contact | list[$T$] of `Transformation` | — | ✓ |
| **Temporal** | | | | |
| `video_length_frames` | Frames per clip | list[$T$] of `int` | ✓ | — |
| `time_stamp_rel_seq` | Frame timestamps (relative) | list[$T$] of list[$N_i$] of `timedelta` | ✓ | — |
| `touch_start_time_rel` | Start of contact | list[$T$] of `timedelta` | ✓ | — |
| `touch_end_time_rel` | End of contact | list[$T$] of `timedelta` | ✓ | — |
| **Sensor Data** | | | | |
| `sensor_video` | Raw video sequences | list[$T$] of `torchvision.io.VideoReader` | ✓ | — |
| `sensor_image` | Extracted tactile snapshot | list[$T$] of `np.ndarray` | — | ✓ |

**Notes:** $T$ is the number of touches in a data point; $N_i$ is the number of frames in the $i$-th video.

Table 3: Overview of the splits of the MNIST 3D dataset.

| Split | Objects per class | Objects total | Share | Description |
|---|---|---|---|---|
| train | 1,148 | 11,480 | 84.5% | Training split. Used by the TactileMNIST environment. |
| test | 100 | 1,000 | 7.4% | Test split. Used by the TactileMNIST-test environment. |
| holdout | 50 | 500 | 3.7% | Holdout split. |
| printed_train | 50 | 500 | 3.7% | Corresponds to the physical 3D-printed digits used in the *training* split of the Real Tactile MNIST dataset collection. |
| printed_test | 10 | 100 | 0.7% | Corresponds to the physical 3D-printed digits used in the *test* split of the Real Tactile MNIST dataset collection. |
| Total | 1,358 | 13,580 | 100% | |

# E Real Tactile MNIST Dataset

As described in Section 3.5, to collect high-quality, real-world tactile data on 600 3D-printed MNIST digits, we used a Franka Research 3 robot arm equipped with a GelSight Mini tactile sensor. To that end, we replaced the end-effector of the robot with a custom-made 3D-printed mount for the GelSight Mini sensor, as shown in Fig. 4. In the following, we further detail the data collection protocol as well as the data point structure for our dataset.

**Data collection protocol** Each of the 600 3D-printed MNIST digits was placed in a $12 \times 12$ cm cell on a rubber mat. The robot moved systematically across the dataset, visiting one digit at a time and performing 256 distinct tactile interactions per object. For each interaction, a target location was sampled randomly within the cell, and the robot was commanded to lower the sensor perpendicularly onto the digit's surface. The coordinate frame of each cell was defined in the center with the x-axis pointing to the right of the robot, the y-axis pointing away from the robot, and the z-axis pointing up, orthogonal to the cell.

The tactile interaction was terminated automatically when the robot detected a normal force exceeding 5 N. At this point, the timestamp of contact was registered, and the sequence was recorded. This force threshold was chosen to ensure consistent contact without damaging the elastomer or over-deforming the object geometry. Note, however, that the Franka robot does not have force sensors in its wrist and instead relies on torque sensors in its joints and a dynamics model to compute the acting forces. Depending on the pose of the robot, this force estimate varies in accuracy, which is why the force being applied to the object is not guaranteed to be consistent.

To maintain data quality and preserve sensor integrity, the elastomer gel pad of the GelSight sensor was replaced halfway through the data collection process (after 76,800 touches). This replacement helped mitigate the impact of gel degradation on the quality of the tactile readings.

**Dataset structure** The dataset contains a total of 153,600 tactile interactions (600 digits $\times$ 256 touches). These are stored as video sequences capturing the contact event from the moment the sensor begins moving toward the digit until full contact is achieved and the interaction is complete.

We provide two formats of this dataset:

- **Sequence data points** (`seq`): Each touch is stored as a short video, capturing the dynamic contact event.
- **Single-image data points** (`single`): Each touch is represented by a single frame extracted at the moment of contact.

Each data point in both formats represents a group of 256 touches per digit and includes metadata for downstream use in learning-based or analytical tasks. Each "`seq`" data point contains fine-grained temporal and spatial information about the tactile interaction, useful for modeling contact dynamics or time-series prediction. The "`single`" variant simplifies data loading and reduces computational overhead. A summary of the data fields is shown in Table 4.

# F Environment Details

In this section, we provide a more detailed description of each environment included in our framework, along with the metrics used to evaluate them within the context of our benchmark protocol. Notably, both `ap_gym` and TMBS encompass tasks that can be categorized as either regression or classification (see Table 1 for an overview). Accordingly, we organize the evaluation metrics into two groups: classification metrics and regression metrics.

## F.1 Environment-Specific Evaluation Metrics

In Section 3.6, we explain both the aggregated and environment-specific metrics for our `ap_gym` and TMBS. Particularly, our environment-specific metrics offer several insights into the exploration and efficiency capabilities of our methods. Depending on the type of task (classification or regression), different metrics are available. We briefly explain all tracked metrics in Table 5.

Table 5: Overview of all evaluation metrics computed by `ap_gym` and TMBS.

| **Environment type** | **Metric name** | **Step-based** | **Description** |
|---|---|---|---|
| Classification | Accuracy | ✓ | Can either be zero or one in each step. One if the correct label has the highest predicted probability, else zero. |
| | Correct label prob. | ✓ | Probability of the correct label in the prediction of the agent. |
| | First correct class. | ✗ | Step relative to the start of the episode at which the agent predicted the correct class for the first time. If there is no correct prediction in an episode, this metric is not computed. |
| | Last incorrect class. | ✗ | Step relative to the start of the episode at which the agent predicted an incorrect class for the last time. |
| Regression | Mean squared error | ✓ | Mean squared error between the agent's prediction and the ground-truth value. |
| | Euclidean distance | ✓ | Square root of the mean squared error. |
| Pose Estimation | Linear error | ✓ | The distance between the predicted object position and the actual object position. |
| | Angular error | ✓ | The angle between the predicted object orientation and the actual object orientation. |
| Center of Mass Estimation | Linear error | ✓ | The distance between the predicted object center of mass and the actual object position. |
| Volume Estimation | Relative error | ✓ | Predicted volume divided by the actual volume of the object (1 for a perfect prediction). |

### F.2 `ap_gym` Environments

The `ap_gym` environments are mainly divided into image classification tasks, image search tasks, and 2D localization tasks. In the following, we detail each of these.

#### F.2.1 Image Classification Task

In the image classification environments, an agent must classify an image by navigating through it using a small glimpse window. This glimpse is never large enough to capture the full image in a single view, necessitating sequential exploration to accumulate sufficient information. As can be seen in Fig. 11–12, the agent's current glimpse is marked in blue. Additionally, the agent's previous glimpses are visualized with a color gradient from red to green, where red indicates a predicted probability of 0 for the correct class, and green indicates a predicted probability of 1. All image classification environments in `ap_gym` share the properties described in Table 6.

Table 6: Environment specifications shared by all image classification tasks.

| Property | Specification |
|---|---|
| **Action Space** | The action $\mathbf{A} \in [-1, 1]^2$ describes the relative movement of the glimpse sensor. The value is first projected onto the unit circle and then scaled by 0.2 (10 % of the image), before being added to the current sensor position. |
| **Prediction Space** | The agent outputs logits $\mathbf{Y} \in \mathbb{R}^K$ corresponding to predicted probabilities for each class label. |
| **Prediction Target Space** | True label $Y^* \in \{0, ..., K-1\}$. |
| **Observation Space** | The observation space is a dictionary with the following keys:<br>• `glimpse`: a tensor $\in [0,1]^{G \times G \times C}$ that represents a glimpse of the image.<br>• `glimpse_pos`: a vector $\in [-1, 1]^2$ that contains the normalized position of the glimpse within the image.<br>• `time_step`: a scalar $\in [-1, 1]$ that represents the normalized current time step between 0 and the step limit. |
| **Loss Function** | Cross-entropy loss |
| **Reward Function** | At each step, the sum of:<br>1. $-10^{-3} \cdot \|a_t\|$ (action regularization).<br>2. Negative cross-entropy loss between prediction and target. |
| **Initialization** | The glimpse starts at a uniformly random position within the image. |
| **Termination** | The episode ends with the terminate flag set if the step limit (16) is reached. |

**Notation:** $K \in \mathbb{N}$ is the number of classes; $G \in \mathbb{N}$ is the glimpse size; $C \in \mathbb{N}$ is the number of image channels (1 = grayscale, 3 = RGB).

## Implemented Environments

The implemented image classification environments from `ap_gym` are detailed in Table 7.

Table 7: Available image classification environments in `ap_gym`.

| Environment ID | Type | Samples | Size | Glimpse | Step Limit | Classes | Description |
|---|---|---|---|---|---|---|---|
| `CircleSquare` | Grayscale | 1,568 | 28x28 | 5x5 | 16 | 2 | Images of either a circle or square |
| `MNIST` | Grayscale | 60,000 | 28x28 | 5x5 | 16 | 10 | Handwritten MNIST [6] digits |
| `CIFAR10` | RGB | 50,000 | 32x32 | 5x5 | 16 | 10 | CIFAR10 [67] natural images |
| `TinyImageNet` | RGB | 100,000 | 64x64 | 10x10 | 16 | 200 | TinyImageNet [68] natural images |

**CircleSquare:** In this environment, the agent must distinguish between a circle and a square present anywhere in the environment. As shown in Fig. 9, a color gradient guides the agent toward the object. The main challenge in `CircleSquare` is its sparse-reward nature, in that the agent must learn to take many steps before gaining any information.

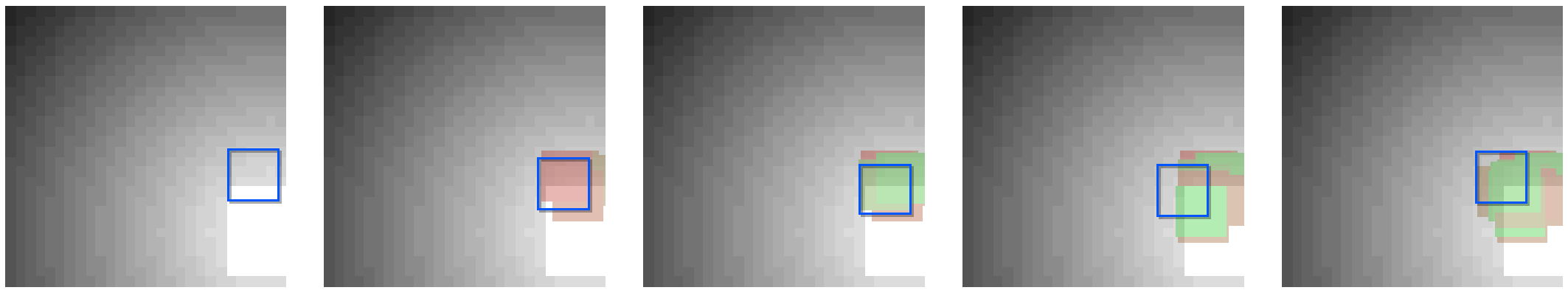

Figure 9: Sequence of glimpses taken by the agent in the CircleSquare environment; the color gradient guides the agent toward the object.

**MNIST:** In this environment, the agent must classify MNIST [6] digits with partial visibility. In contrast to CircleSquare, the agent generally receives information more quickly, but always sees an incomplete picture and, thus, must learn contour-following strategies to gather enough information for a successful classification. This environment contains the train and test variants. In Fig. 10, we show one episode of the MNIST environment.

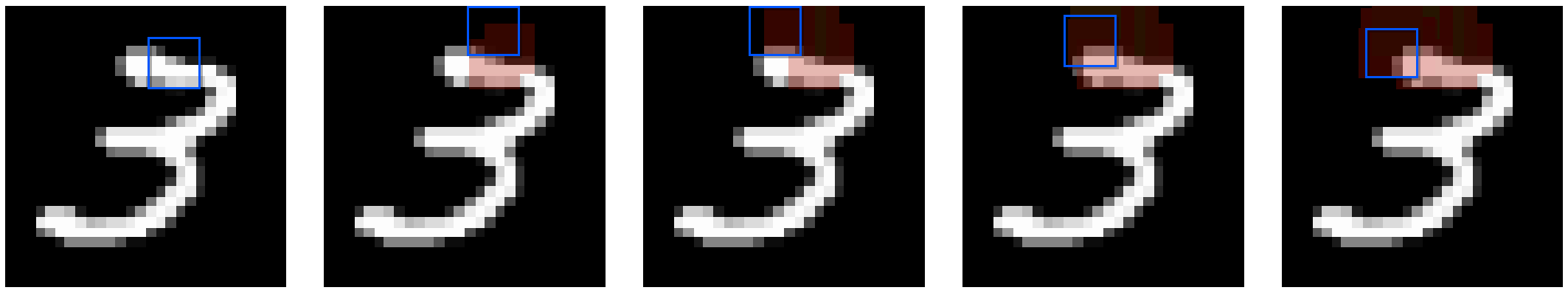

Figure 10: Sequence of glimpses taken by the agent in the MNIST environment on the test set. We show a failed episode here, as can be seen by the red window representing the agent's glance.

**CIFAR10:** This environment (see Fig. 11) challenges the agent to classify CIFAR10 images by sequentially observing glimpses. In contrast to the `MNIST` environment, the data of CIFAR10 is much more diverse and complex. This environment has the train and test variants.

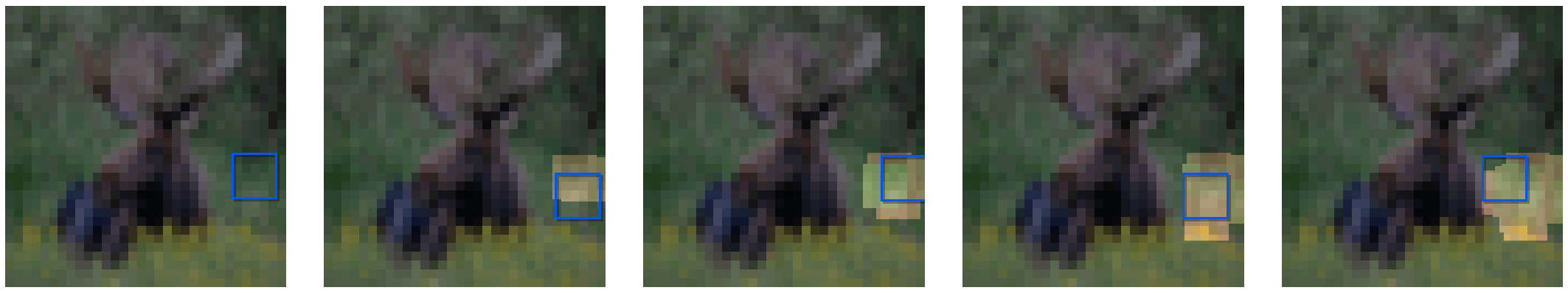

Figure 11: Sequence of glimpses taken by the agent in the CIFAR10 environment.

**TinyImageNet:** This environment increases classification difficulty due to larger images and a greater number of classes. The higher-resolution data requires more sophisticated exploration strategies. Variants of this environment include the train and test splits. In Fig. 12, we show snapshots of an episode in the `TinyImageNet` environment.

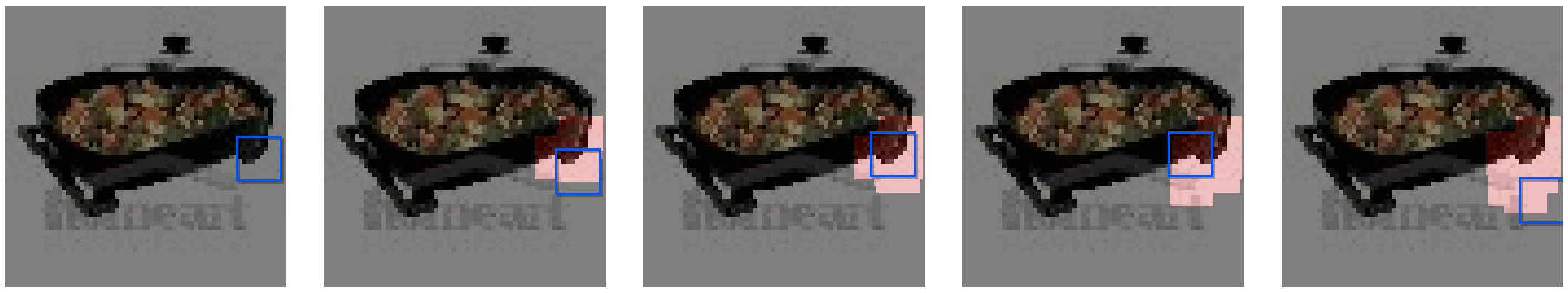

Figure 12: Sequence of glimpses taken by the agent in the TinyImageNet environment.

### F.2.2 Image Search Environments

In the image search environments, an agent must localize a target glimpse within a larger image by moving a small observation window. Since the glimpse cannot cover the full image at once, the agent must navigate sequentially to gather sufficient context and then regress the coordinates of the target region. As shown in Fig. 14, the agent's current glimpse is outlined in blue, the target glimpse in transparent purple, and the agent's prediction as an opaque purple box. Past glimpses are colored from red (large error) to green (small error) according to how close each prediction was to the true target. All image search environments in `ap_gym` share the properties in Table 8.

Table 8: Shared properties of all image search tasks.

| Property | Specification |
|---|---|
| **Action Space** | The action $\mathbf{A} \in [-1, 1]^2$, projected onto the unit circle and scaled by a maximum step length (default 0.20 of image size), describes the relative movement of the glimpse sensor. |
| **Prediction Space** | $\mathbf{Y} \in \mathbb{R}^2$ contains the agent's prediction of the target glimpse position. |
| **Prediction Target** | $Y^* \in \mathbb{R}^2$ contains the true position of the target glimpse. |
| **Observation Space** | The observation is a dictionary with the following keys:<br>• `glimpse`: a vector $\in [0, 1]^{G \times G \times C}$ that represents a glimpse of the image.<br>• `glimpse_pos`: a vector $\in [-1, 1]^2$ that contains the normalized position of the glimpse within the image.<br>• `target_glimpse`: a vector $\in [0, 1]^{G \times G \times C}$ that represents the target glimpse.<br>• `time_step`: a scalar $\in [-1, 1]$ that represents the normalized current time step between 0 and the step limit. |
| **Loss Function** | Mean squared error |
| **Reward Function** | At each step, the sum of:<br>1. $-10^{-3} \cdot \|a_t\|$ (action regularization).<br>2. The negative mean squared error between the agent's prediction and the true coordinates of the target glimpse. |
| **Initialization** | The agent starts at a uniformly random position within the image. |
| **Termination** | The episode ends if the step limit is reached. |

**Notation:** $G \in \mathbb{N}$ is the glimpse size; $C \in \mathbb{N}$ is the number of image channels (1 = grayscale, 3 = RGB).

## Implemented Environments

The available image search environments are listed in Table 9.

Table 9: Image search environments in `ap_gym`.

| Environment ID | Type | Samples | Image Size | Glimpse | Steps | Channels | Description |
|---|---|---|---|---|---|---|---|
| `CIFAR10Search` | RGB | 50 000 | 32×32 | 5×5 | 16 | 3 | CIFAR10 [67] natural images. |
| `TinyImageNetSearch` | RGB | 100 000 | 64×64 | 10×10 | 16 | 3 | TinyImageNet [68] natural images. |

**CIFAR10Search:** In `CIFAR10Search` (see Fig. 13), the agent must localize a given 5×5 patch within a 32×32 RGB image. Limited visibility forces a strategic search for the target glimpse to reduce localization error.

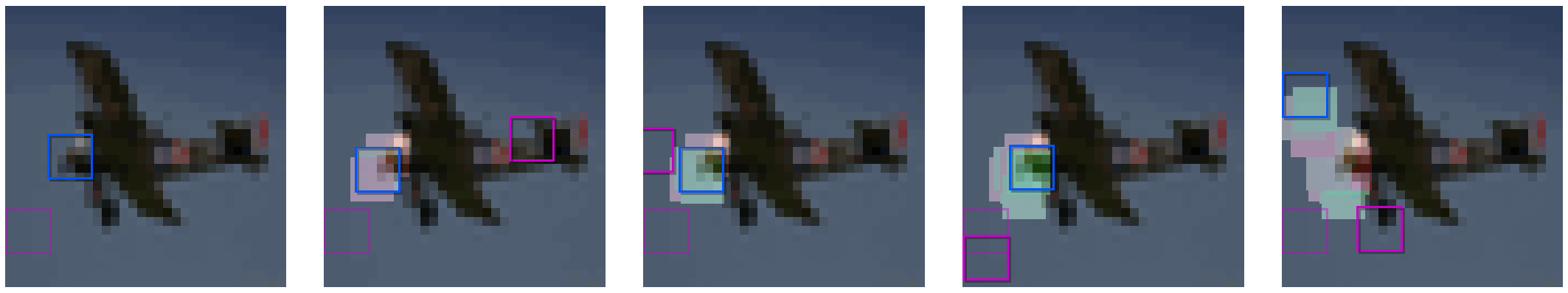

Figure 13: Agent trajectory in `CIFAR10Search`. The blue window represents the current glimpse, the transparent purple box the target, and the opaque purple box the prediction. Past glimpses are colored with a gradient from red to green, where red indicates a high predicted error and green indicates a lower error.

**TinyImageNetSearch:** Similar to `CIFAR10Search`, in `TinyImageNetSearch` (Fig. 14), the agent's task is to localize a given glimpse in a natural image. However, the greater resolution and diversity make localization more challenging than `CIFAR10Search`.

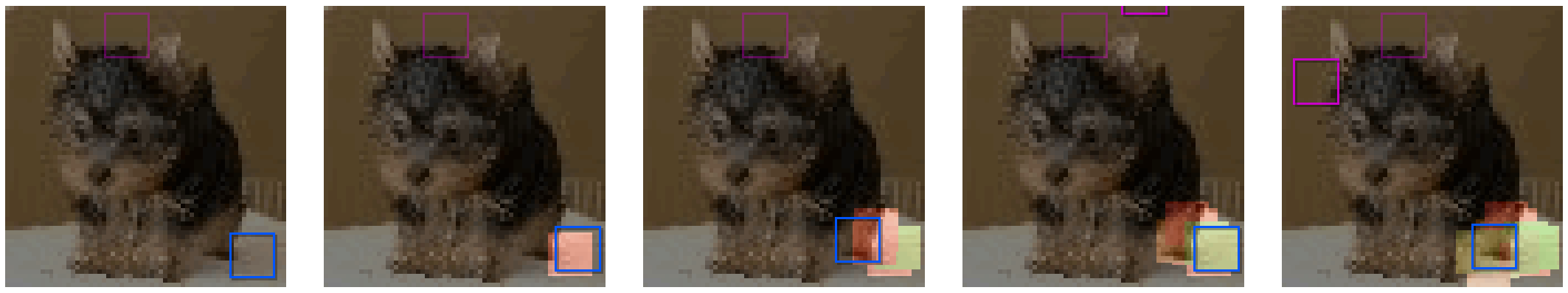

Figure 14: Agent trajectory in `TinyImageNetSearch`. The blue window represents the current glimpse, the transparent purple box the target, and the opaque purple box the prediction. Past glimpses are colored with a gradient from red to green, where red indicates a high predicted error and green indicates a lower error.

### F.2.3 2D Localization Environments

In the 2D localization task, an agent must predict its own position by actively exploring its surroundings. Generally, our 2D localization environments can be divided into the following types:

- *LightDark*: The agent must estimate its position based on noisy observations, where the noise level depends on the brightness of the surrounding area.
- *Static LIDAR environments*: The agent is placed at a random position in a fixed 2D bitmap map, with white pixels denoting free space and black pixels indicating obstacles. At each step, it receives 8-beam LIDAR distance readings and exact odometry. Since the map remains constant across episodes, the agent can gradually learn its layout to improve localization.
- *Dynamic LIDAR environments*: Each episode begins with a procedurally generated random map, ensuring minimal repetition. To enable localization in these unseen maps, the full map is provided as part of the observation. As in the static case, the agent also receives 8-beam LIDAR data and odometry at each step.

An episode of the LightDark task is shown in Fig. 15. Examples of agent trajectories in the static variants, LIDARLocRoomsStatic and LIDARLocMazeStatic, are shown in Fig. 16 and Fig. 17. The dynamic variants, LIDARLocRooms and LIDARLocMaze, look the same, except that they use a different map in every episode. Additionally, all 2D localization tasks share the common properties described in Table 10.

Table 10: Properties of 2D localization environments in `ap_gym`.

| **Property** | **LIDAR Environments** | **LightDark Environment** |
|---|---|---|
| **Action Space** | Describes the agent's relative movement. The action, $\mathbf{a} \in [-1,1]^2$, is projected onto the unit circle and added to the position (in pixels). | Describes the agent's relative movement. The action, $\mathbf{a} \in [-1,1]^2$, is projected onto the unit circle and scaled by 0.15. |
| **Prediction Space** | $\mathbf{Y} \in \mathbb{R}^2$ (normalized predicted position of the agent). | $\mathbf{Y} \in \mathbb{R}^2$ (predicted position of the agent). |
| **Prediction Target** | $Y^* \in \mathbb{R}^2$ (normalized true agent position). | $Y^* \in \mathbb{R}^2$ (true agent position). |
| **Observation Space** | The observation space is a dictionary with the following keys:<br>• `lidar`: a vector $\in [0,1]^8$ that contains distances measured by the LIDAR sensor.<br>• `odometry`: a vector $\in [-1,1]^2$ with the agent's normalized relative displacement.<br>• `time_step`: a scalar $\in [-1,1]$ that represents the normalized current time step between 0 and the step limit.<br>• `map`: a vector $\in [0,1]^{M\times M\times 1}$ with a grayscale image representation of the environment (dynamic maps only). $M \in \mathbb{N}$ is the map size. | The observation space is a dictionary with the following keys:<br>• `noisy_position`: a vector $\in [-2,2]^2$ that contains a noisy estimate of the agent's position depending on the brightness of the area the agent is in.<br>• `time_step`: a scalar $\in [-1,1]$ that represents the normalized current time step between 0 and the step limit. |
| **Loss Function** | Mean squared error | Mean squared error |
| **Reward Function** | At each step, the sum of:<br>1. $-10^{-3} \cdot \lVert a_t \rVert^2$ (action regularization);<br>2. A constant reward of 0.1;<br>3. The negative mean squared error between the agent's prediction and its true position. | At each step, the sum of:<br>1. $-10^{-3} \cdot \lVert a_t \rVert^2$ (action regularization);<br>2. A constant reward of 1.0 to ensure that the reward stays positive and the agent does not learn to terminate the episode on purpose.<br>3. The negative mean squared error between the agent's prediction and its true position. |
| **Initialization** | The agent begins at a uniformly random, valid location within the environment. | The agent's initial position is uniformly randomly sampled from the range $[-1,1]^2$. |
| **Termination** | The episode ends if the step limit is reached. | The episode ends if either the step limit is reached or the agent moves out of bounds. |

### Implemented Environments

The implemented 2D localization environments from `ap_gym` are detailed in Table 11.

Table 11: Available 2D localization environments in `ap_gym`.

| Environment ID | Map Type | Static? | Size | Description |
|---|---|---|---|---|
| `LightDark` | LightDark | No | continuous | Localization with brightness-varying noise. |
| `LIDARLocRoomsStatic` | Rooms | Yes | 32×32 | Static rooms. |
| `LIDARLocMazeStatic` | Maze | Yes | 21×21 | Static maze. |
| `LIDARLocRooms` | Rooms | No | 32×32 | Random rooms each episode. |
| `LIDARLocMaze` | Maze | No | 21×21 | Random maze each episode. |

**LightDark:** In the `LightDark-v0` environment (Fig. 15), the agent must estimate its position using noisy observations, where observation noise depends on the brightness of the region it is in. Bright areas yield low-variance signals while dark regions introduce greater uncertainty. The agent can move toward light to improve accuracy. What makes this environment particularly challenging is that the agent must learn to move toward the light without knowing its own precise position. Hence, it faces uncertainty not only with regard to its prediction, but also with regard to the outcomes of its action.

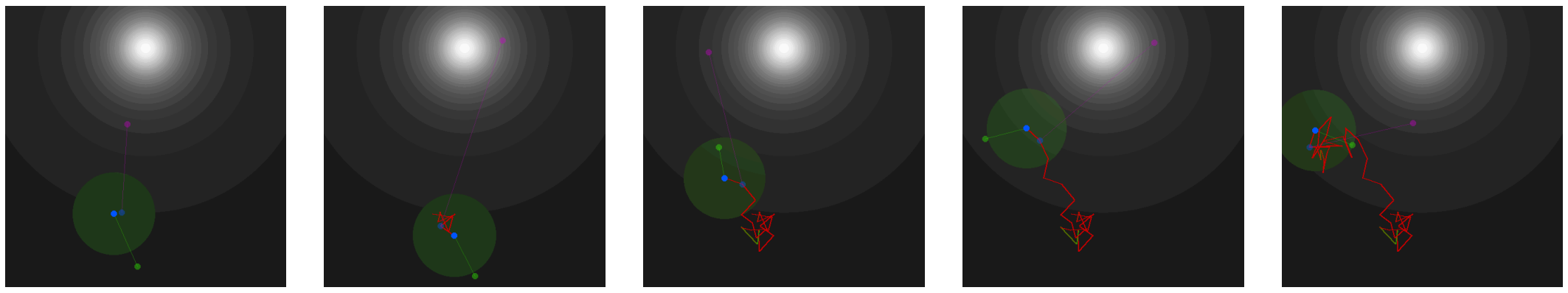

Figure 15: Agent trajectory in the `LightDark` environment. The blue dot indicates the agent's current position, while the light blue dot marks its previous position (i.e., the target for prediction). The purple dot shows the agent's last prediction, and the green dot is the noisy observation the agent receives. The green transparent circle around the agent reflects observation uncertainty, which is higher in dark regions and lower in bright ones. The background encodes environmental visibility: white areas correspond to low-noise (bright) regions, and dark areas to high-noise (dark) regions. The line representing the path the agent has taken is encoded with a gradient from red to green, where red indicates a high predicted error and green indicates lower error.

**LIDARLocRoomsStatic:** In the `LIDARLocRoomsStatic` environment (see Fig. 16), the agent must localize itself with LIDAR sensors in a map with wide open areas. As the open areas are large, it often does not receive any information from its LIDAR sensors (e.g., when it is in the middle of the room). The agent, therefore, must navigate around the map to gather information and localize itself. In this version of the environment, the map stays constant, meaning that the agent can memorize the layout of the rooms over the course of the training.

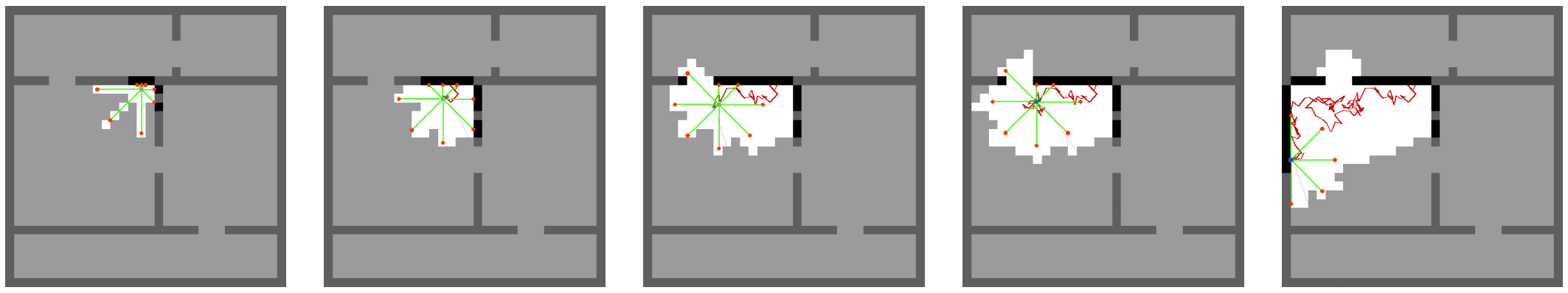

Figure 16: Agent trajectory in the `LIDARLocRoomsStatic` environment. Here, we see the 8 LIDAR beams extending from the agent, which often return no reading when the agent is far away from the walls. The purple dot is the agent's predicted pose, and the line representing the path the agent has taken is encoded with a gradient from red to green, where red indicates a high predicted error and green indicates lower error.

**LIDARLocMazeStatic:** In the `LIDARLocMazeStatic` environment, the agent has to localize itself with LIDAR sensors in a map with narrow corridors. As the corridors are narrow, it will always receive information from its LIDAR sensors, but many regions of the maze look alike. The agent must therefore navigate around the map to gather information. In this variant, the map stays constant, meaning that the agent can memorize the layout of the maze over the course of the training. In Fig. 17, we show a visualization of this environment.

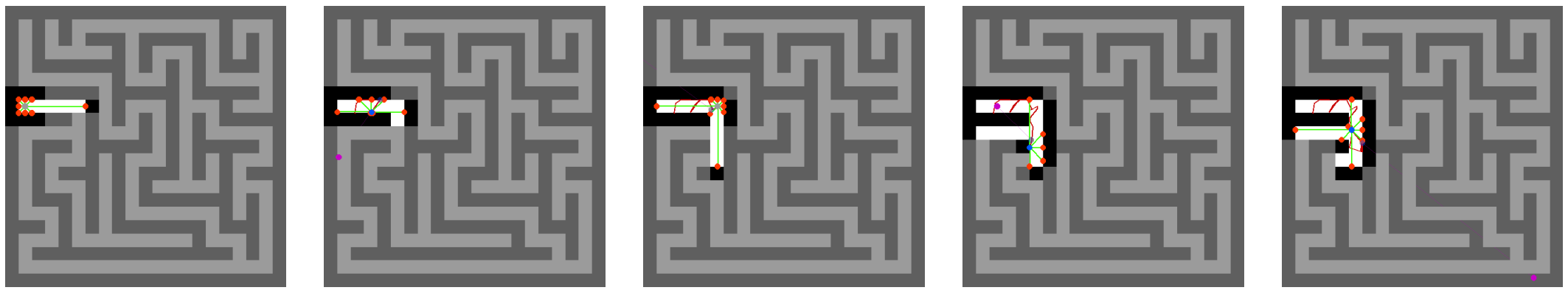

Figure 17: Agent trajectory in the `LIDARLocMazeStatic` environment, where the corridors are narrow and many regions look alike. The purple dot is the agent's predicted pose, and the line representing the path the agent has taken is encoded with a gradient from red to green, where red indicates a high predicted error and green indicates lower error.

**LIDARLocRooms:** The `LIDARLocRooms` localization task is the same as `LIDARLocRoomsStatic` (Fig. 16), but with a different map in every episode. The agent explores larger open areas which have sparse LIDAR readings—especially near the center of rooms. Hence, the agent has to learn to process the map it is provided as additional input.

**LIDARLocMaze:** The `LIDARLocMaze` task is the same as `LIDARLocMazeStatic` (Fig. 17), but the agent navigates procedurally generated mazes with narrow corridors. These layouts change every episode, so the agent receives the full map as part of its observation. Despite always receiving LIDAR readings (due to proximity to walls), the similarity of many areas makes localization challenging.

### F.3 TMBS Environments

The Tactile MNIST Benchmark Suite (TMBS) is designed to evaluate the efficiency of active tactile perception strategies, testing an agent's ability to strategically explore, classify, and localize objects while minimizing the number of interactions. As described in Section 3.3, TMBS provides a unified framework based on a simulated GelSight Mini tactile sensor [15], supporting three rendering modes: Taxim, CycleGAN, and Depth. For visualization, the sensor, object, and platform where the object is placed are rendered in 3D. Although the physical interaction between sensor and object is not simulated, object positions are randomly shifted to emulate tactile interaction. Our suite includes eight distinct environments covering both classification and regression tasks, namely (i) *TactileMNIST* for digit classification, (ii) *Starstruck* for shape counting, (iii) *PrimitiveGeometries* for the classification of geometric primitives, (iv) *Toolbox* for pose estimation, (v) *ABCCenterOfMass* for center-of-mass estimation of CAD models, (vi) *TactileMNISTVolume* for digit volume estimation, (vii) *PrimitiveGeometriesVolume* for volume estimation of geometric primitives, and (viii) *ABCVolume* for volume estimation of CAD models. These are detailed in the next sections.

#### F.3.1 Classification Environments

In tactile classification environments, the agent has to classify a 3D object by exploring it with a GelSight Mini tactile sensor. The agent does not have access to the object's location or orientation, and also receives no visual input. Instead, it must actively control the sensor to find and classify the object. In Table 12 we detail the main environment specifications.

Table 12: Shared properties of all tactile classification environments.

| Property | Specification |
|---|---|
| **Action Space** | The action space $\mathbf{A}$ is a dictionary with the following keys:<br>• `sensor_target_pos_rel`: a vector $\in [-1, 1]^3$ containing the normalized relative linear target movement.<br>• `sensor_target_rot_rel` (only with sensor rotation): a vector $\in [-1, 1]^3$ containing the relative angular target movement as a rotation vector. |
| **Prediction Space** | The agent outputs logits $\mathbf{Y} \in \mathbb{R}^K$ corresponding to predicted probabilities for each class label. |
| **Prediction Target Space** | True label $Y^* \in \{0, ..., K-1\}$. |
| **Observation Space** | The observation space is a dictionary with the following keys:<br>• `sensor_img`: a tensor $\in [-1, 1]^{64 \times 64 \times 3}$ representing the tactile image from the GelSight sensor.<br>• `sensor_pos`: a vector $\in [-1, 1]^3$ indicating the normalized position of the sensor in the workspace.<br>• `time_step`: a scalar $\in [-1, 1]$ representing the normalized current time step. |
| **Loss Function** | Cross-entropy loss |
| **Reward Function** | At each step, the sum of:<br>1. $-10^{-3} \cdot \\|a_t\\|^2$ (action regularization).<br>2. Negative cross-entropy loss between prediction and target. |
| **Initialization** | The tactile sensor starts at a randomly sampled pose in the workspace. |
| **Termination** | The episode ends with the terminate flag set if the step limit is reached. |

**Notation:** $K \in \mathbb{N}$ is the number of classes.

## Implemented Environments

The implemented tactile classification environments from TMBS are detailed in Table 13.

Table 13: Available tactile classification environments in TMBS.

| Environment ID | Dataset | Classes | Steps | Perturbation | Sensor Rot. | Description |
|---|---|---|---|---|---|---|
| `TactileMNIST` | MNIST 3D | 10 | 16 | Yes | No | Classify objects from the *MNIST 3D* dataset using tactile feedback. |
| `Starstruck` | 3D shapes | 3 | 32 | No | No | Count the number of stars in a tactile scene. |
| `PrimitiveGeometries` | Primitives | 4 | 32 | Yes | Yes | Tell apart geometric primitives of random size by their outline. |

**TactileMNIST:** In this environment (see Fig. 18 for an episode of the TactileMNIST environment), the agent must classify 3D models of handwritten digits by touch alone. See Section 3.4 for a description of the MNIST 3D dataset. Starting from a random pose, it first locates the digit and then follows its contour to accumulate shape information. With the object perturbation enabled, the digit shifts slightly under the sensor, requiring robust strategies that are invariant to these small shifts in the object's pose. Variations of this environment include the test and train splits, as well as the three rendering modes — CycleGAN, Taxim, and Depth.

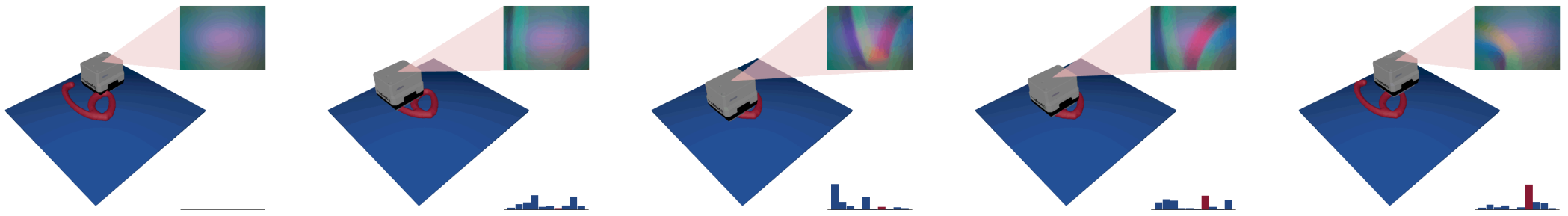

Figure 18: Sequence of tactile glimpses taken by the agent in the TactileMNIST classification environment. The agent observes only local tactile glimpses and its own position. As it actively explores the 3D digit surface, its prediction confidence increases, driven by accumulating shape information. The probabilities for each label are displayed on the bar plot on the side.

**Starstruck:** In the Starstruck environment (see Fig. 19), the agent must count the number of stars in a scene cluttered with other geometric objects, including circles and squares. Since all stars look the same, distinguishing stars from other objects is straightforward. Instead, the main challenge posed in this environment is to learn an effective search strategy to systematically cover as much space as possible. Here, the scenes are prearranged and provided in a small dataset. Variations include the test and train variants, as well as depth and Taxim renderings.

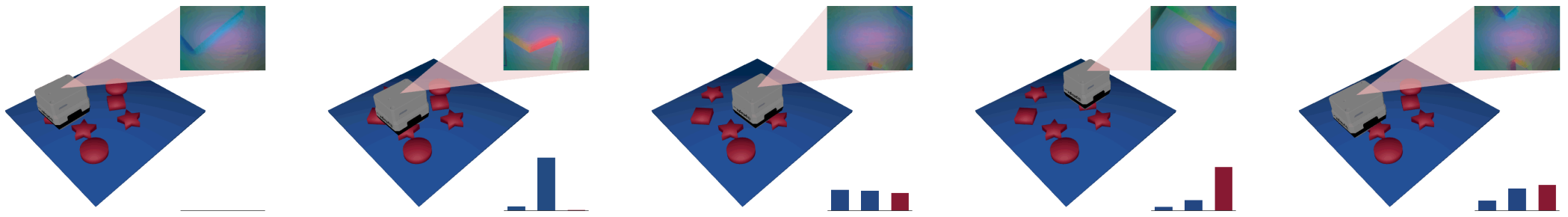

Figure 19: Sequence of tactile glimpses taken by the agent in the Starstruck counting task. The agent perceives only local tactile information and its own position as it explores the scene. To accurately estimate the number of stars, it must develop an efficient search strategy. The presence of distractor shapes (e.g., circles and squares) and scene clutter increases the difficulty of the task. On the side of each episode, the probabilities of each label are shown.

**PrimitiveGeometries:** In the PrimitiveGeometries environment, the agent must tell apart four classes of geometric primitives: cylinders, cuboids, triangular prisms, and half cylinders. Every free dimension of an object (radius or side lengths of the footprint, and a height between 20 and 40 mm) is drawn independently and uniformly from a class-specific range, so the classes overlap in size and differ only in their outline. The objects are large relative to the platform, so finding them is easy. Instead, the challenge lies in telling the shapes apart: a touch on a flat top face looks the same for every class, so the agent has to seek out the edges and sides of the object, which it can touch by tilting the sensor, as sensor rotation is enabled. The objects are generated procedurally: each of the 400,000 objects in the train split and the 400,000 objects in the test split is derived deterministically from its index, and since the dimensions are drawn from a continuous distribution, the agent cannot memorize them. Fig. 20 shows an example episode.

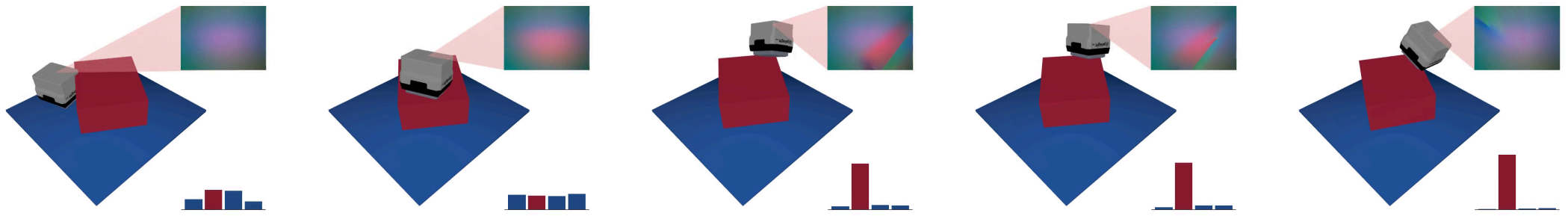

Figure 20: Sequence of tactile glimpses taken by the agent in the PrimitiveGeometries classification environment. The agent observes only local tactile glimpses and its own pose. It first presses on the flat top face of the cuboid, which looks the same for every class, then moves to the top edges and tilts the sensor over a corner; the straight edge imprints let it quickly settle on the correct class (cuboid). The probabilities for each label (cylinder, cuboid, triangular prism, half cylinder) are displayed on the bar plot on the side.

### F.3.2 Regression Environments

In tactile regression environments, the agent must infer a continuous property of a 3D object by exploring it with a GelSight Mini tactile sensor. The agent receives no visual input or ground-truth pose; instead, it must actively control the sensor to locate and explore the object.

All tactile regression environments share the properties in Table 14.

Table 14: Shared properties of all tactile regression environments.

<table>
<tr><th>Property</th><th>Toolbox</th><th>TactileMNISTVolume</th><th>PrimitiveGeometries-Volume, ABCVolume</th><th>ABCCenterOfMass</th></tr>
<tr><td>Action Space</td><td colspan="2">The action space $\mathbf{A}$ is a dictionary with the following keys:<br>• <code>sensor_target_pos_rel</code>: a vector $\in [-1,1]^3$ containing the normalized relative linear target movement.</td><td colspan="2">The action space $\mathbf{A}$ is a dictionary with the following keys:<br>• <code>sensor_target_pos_rel</code>: a vector $\in [-1,1]^3$ containing the normalized relative linear target movement.<br>• <code>sensor_target_rot_rel</code>: a vector $\in [-1,1]^3$ containing the relative angular target movement as a rotation vector.</td></tr>
<tr><td>Prediction Space</td><td>$\mathbf{Y} \in \mathbb{R}^4$, representing the pose of the object, which includes the normalized 2D position on the platform and the sine and cosine of the wrench rotation around the up-facing Z-axis.</td><td>$Y^* \in \mathbb{R}$, representing the normalized volume of the object.</td><td>$Y^* \in \mathbb{R}$, representing the normalized volume of the object.</td><td>$Y^* \in \mathbb{R}^2$, representing the center of mass of the object projected onto the platform.</td></tr>
<tr><td>Prediction Target Space</td><td colspan="4">See prediction space above</td></tr>
<tr><td>Observation Space</td><td colspan="4">The observation space is a dictionary with the following keys:<br>• <code>sensor_img</code>: a tensor $\in [-1,1]^{64\times 64\times 3}$ representing the tactile image from the GelSight sensor.<br>• <code>sensor_pos</code>: a vector $\in [-1,1]^3$ indicating the normalized position of the sensor in the workspace.<br>• <code>time_step</code>: a scalar $\in [-1,1]$ representing the normalized current time step.</td></tr>
<tr><td>Loss Function</td><td>Mean squared error.</td><td colspan="2">Normalized Huber loss (see below).</td><td>Mean squared error.</td></tr>
<tr><td>Reward Function</td><td colspan="4">At each step, the sum of:<br>1. $-10^{-3} \cdot \|a_t\|^2$ (action regularization).<br>2. The negative loss of the current prediction of the agent.</td></tr>
<tr><td>Initialization</td><td colspan="4">The glimpse starts at a uniformly random position within the workspace.</td></tr>
<tr><td>Termination</td><td colspan="4">The episode ends with the terminate flag set if the step limit is reached.</td></tr>
</table>

In all volume estimation environments, the prediction target is the object volume, standardized to zero mean and unit variance with the statistics of the training split (the test split uses the same statistics). As the volume distributions are heavy-tailed, the loss is the Huber loss of the standardized volume with $\delta = 1$, i.e., quadratic for errors up to one standard deviation and linear beyond, so that the few very large objects do not dominate the loss. As for all tasks (see Section 3.6), we normalize the loss such that the best blind guess yields a loss of 1. Under the Huber loss, this guess, i.e., the constant prediction with the lowest expected loss on the training objects, is not the mean volume but its Huber location estimate.

### Implemented Environments

The implemented tactile regression environments from TMBS are detailed in Table 15.

Table 15: Available tactile regression environments in TMBS.

| Environment ID | Dataset | Dims | Steps | Perturbation | Sensor Rot. | Description |
|---|---|---|---|---|---|---|
| `Toolbox` | 3D Tools | 4 | 64 | Yes | No | Estimate the 2D position and orientation of a tool (e.g., wrench) placed in the environment. |
| `ABCCenterOfMass` | ABC Dataset | 2 | 32 | Yes | Yes | Estimate the center of mass of a realistic industrial CAD model. |
| `TactileMNISTVolume` | MNIST 3D | 1 | 32 | Yes | No | Estimate the normalized volume of an MNIST 3D digit model. |
| `PrimitiveGeometriesVolume` | Primitives | 1 | 32 | Yes | Yes | Estimate the normalized volume of a geometric primitive of random size. |
| `ABCVolume` | ABC Dataset | 1 | 32 | Yes | Yes | Estimate the normalized volume of a realistic industrial CAD model. |

**Toolbox:** In the `Toolbox` environment (see Fig. 21), the agent must locate a tool on a platform and regress its canonical 2D coordinates and orientation. Touching only the handle or head yields partial information, so the agent must explore, strategically seeking out the wrench's ends to resolve ambiguities and accurately estimate its pose. Here, the pose is described as a 4D vector, with the 2D $(x, y)$ coordinates on the platform and the Z-axis rotation represented as a 2D sine and cosine.

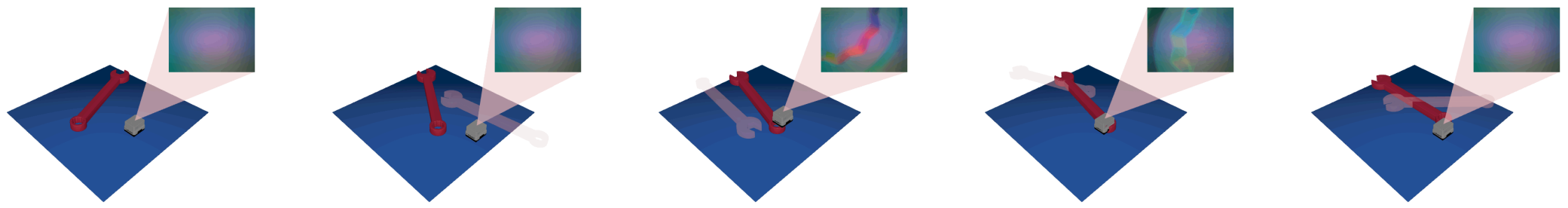

Figure 21: Agent trajectory in the `Toolbox` environment. The agent collects local tactile readings of the wrench, sequentially integrating observations to regress its $(x, y)$ position and orientation (represented as sine and cosine). The predicted pose is shown in transparent red in the simulation platform.

**ABCCenterOfMass:** In the `ABCCenterOfMass` environment (see Fig. 22), the agent must estimate the center of mass of an industrial CAD model from the ABC Dataset [69] using only tactile exploration. Throughout the episode, the object moves randomly, so the agent must robustly track its contour to accumulate sufficient surface samples for accurate positional regression.

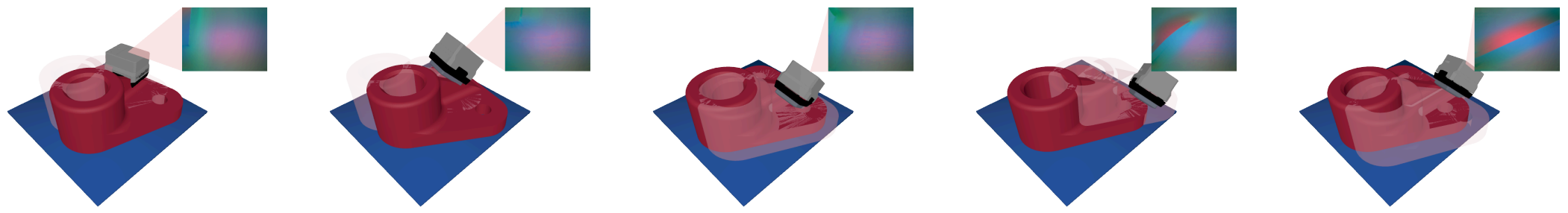

Figure 22: Agent trajectory in the `ABCCenterOfMass` environment. The sensor explores the object's surface under pose perturbations to regress the object's normalized center of mass. We visualize here the predicted position (transparent red) overlaid with the object mesh throughout the episode.

**TactileMNISTVolume:** In the `TactileMNISTVolume` environment (see Fig. 23), the agent must estimate the volume of a 3D handwritten-digit model using only tactile exploration. Throughout the episode, the object moves randomly, so the agent must robustly follow the digit's contour to accumulate sufficient surface samples for an accurate volume regression.

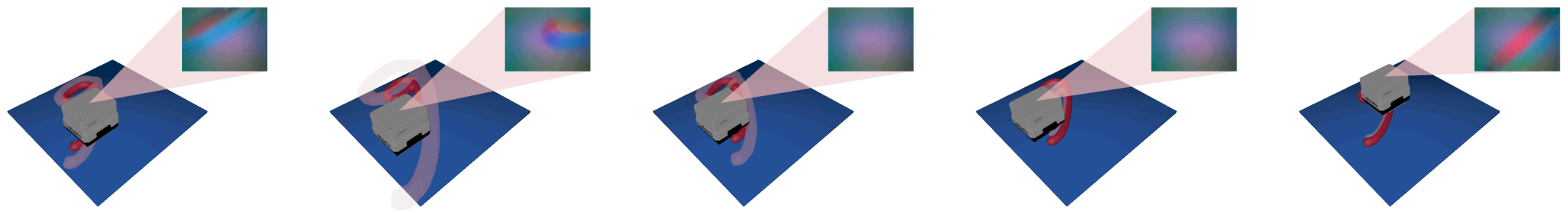

Figure 23: Agent trajectory in the `TactileMNISTVolume` environment. The sensor explores the digit surface under pose perturbations to regress the object's normalized volume. We visualize here the predicted volume (transparent red) overlaid with the object mesh throughout the episode.

**PrimitiveGeometriesVolume:** In the `PrimitiveGeometriesVolume` environment, the agent must estimate the volume of a geometric primitive from the `PrimitiveGeometries` environment:

a cylinder, a cuboid, a triangular prism, or a half cylinder with randomly drawn dimensions. As the objects are simple, a few well-placed touches suffice to measure them: the agent has to find out the shape of the object, measure its footprint, and determine its height, for which it can tilt the sensor to touch the sides of the object. Since the volume of every primitive is known in closed form, the prediction targets are exact. Fig. 24 shows an example episode.

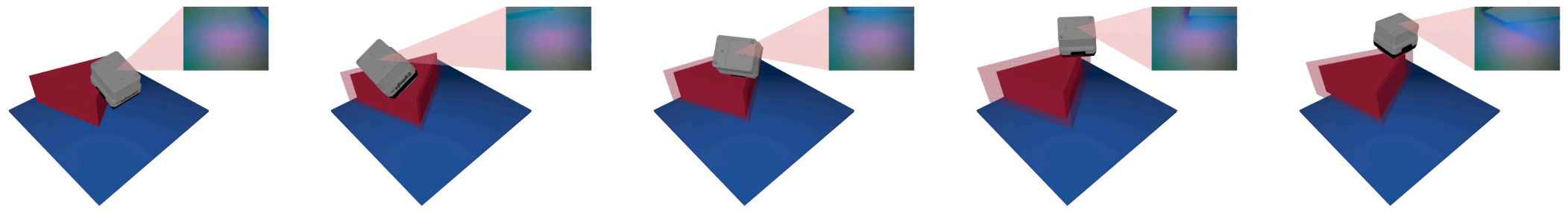

Figure 24: Agent trajectory in the `PrimitiveGeometriesVolume` environment. The agent tracks the top ridge of a triangular prism, tilting the sensor to press against its edges and corners under pose perturbations to regress the object's normalized volume. We visualize here the predicted volume (transparent red) overlaid with the object mesh throughout the episode.

**ABCVolume:** The `ABCVolume` environment (see Fig. 25) is a harder version of `PrimitiveGeometriesVolume`: the agent must estimate the volume of an industrial CAD model from the ABC Dataset [69] using only tactile exploration. As these models are complex and diverse, a few touches no longer suffice to determine their volume, and the agent must adapt its exploration to each object. Throughout the episode, the object moves randomly, so the agent must robustly track its contour to accumulate sufficient surface samples for accurate volume regression.

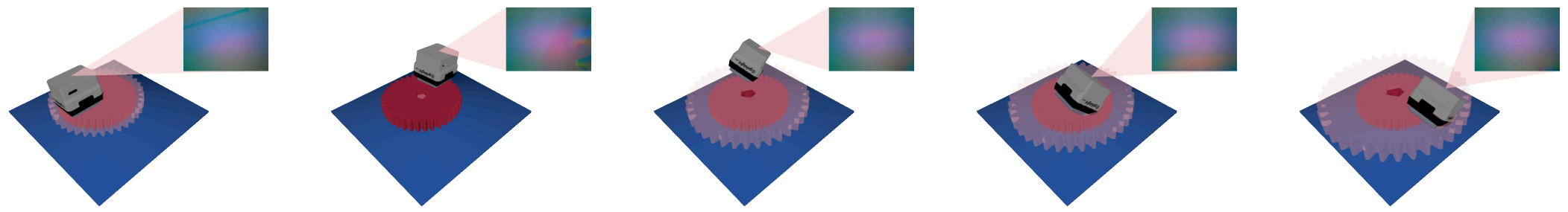

Figure 25: Agent trajectory in the `ABCVolume` environment. The sensor explores the object surface under pose perturbations to regress the object's normalized volume. We visualize here the predicted volume (transparent red) overlaid with the object mesh throughout the episode.

### F.3.3 Sim-to-Real Environments

For the sim-to-real experiments (Appendix H), TactileMNIST and TactileMNISTVolume come in three additional variants. The *RealSnap* variants (TactileMNISTRealSnap and TactileMNISTVolumeRealSnap) replay the Real Tactile MNIST Dataset interactively: in each step, the commanded sensor position is snapped to the closest of the next 16 recorded touches of the current digit, and the corresponding real tactile image is returned as the observation. Each episode starts at the first recorded touch of a randomly drawn digit, and touches are visited in the order in which they were recorded. The *Snap* variants give simulated agents the same action semantics by sampling 16 touch positions uniformly over the cell in each step and moving the sensor to the one closest to the commanded position. The *DR* variants of the Snap environments additionally apply the domain randomization of the sim-to-real recipe (background depth overlay) to the simulated depth maps before they are rendered with the CycleGAN. Correspondingly, the RealSnap environments can pass each real tactile image through the round trip of the CycleGAN before returning it. All variants have a step limit of 16; the remaining details, including the rewards and the prediction targets, are the same as for the regular environments.

## G Detailed Experimental Results

In this section, we present experimental results across all environments from both of our benchmark suites. We group the experiments based on similar evaluation metrics and apply our benchmark to active perception methods introduced in [45] and [44]. Specifically, `APPLE-SAC`, `APPLE-CrossQ`, `APPLE-RND`, and `APPLE-PPO` refer to the *Active Perception Policy Learning* (`APPLE`) framework proposed by [45]. `APPLE` is a reinforcement learning (RL) framework that employs a transformer-based agent trained in a supervised setting. The variants `SAC` [75], `PPO` [77], and `CrossQ` [76] denote the specific RL algorithms used within `APPLE`. In contrast, `RND` represents a random exploration agent that uses `APPLE`'s perception module but does not train the agent to propose actions. As another baseline, we evaluate `HAM`, the Haptic Attention Module proposed in [44]. `HAM` is a REINFORCE-based agent that does not incorporate a vision encoder. Therefore, we evaluate `HAM` only in environments that do not necessitate a vision encoder. Finally, we have also found that `APPLE-PPO` becomes unstable in combination with a vision encoder, so we only evaluate it on non-vision/non-tactile environments as well.

**Classification environments** First, in Fig. 26 and Fig. 27, we present the results of the benchmarked methods on the classification environments from both `ap_gym` and TMBS, which are summarized in Fig. 5. All classification results are evaluated using both average and final accuracy metrics. Overall, `APPLE-CrossQ` and `APPLE-SAC` demonstrate the strongest performance across the classification tasks. In contrast, `HAM` does not outperform the random baseline `APPLE-RND` in any of the classification environments, and `APPLE-PPO` does so only on CircleSquare and MNIST. A possible explanation for this weak performance could be the on-policy nature of these approaches, which does not allow for the reuse of samples for learning later on in the training. As expected, `APPLE-RND` does not achieve high performance in any of the environments, underlining the need for active perception.

To gain more insights into the behavior of policies during training, all environments log per-step metrics, which allows for an analysis of the predictions of the agent throughout individual episodes. Hence, the environments provide detailed protocols of the evaluation metrics over individual steps of an episode. As an example, see Fig. 28, which shows how the accuracy and correct label probability of the agents develop over the course of an episode of the TactileMNIST task.

Despite the comparatively strong performance of `APPLE-CrossQ` and `APPLE-SAC` on some tasks, the challenges presented by the classification environments of this benchmark are far from being solved. Especially in the more complex environments, such as TinyImageNet with 200 classes and Starstruck with a challenging counting task, the performance exhibited by all approaches falls short of optimal. Both sample efficiency and final performance leave room for improvement, and new approaches are needed to close these optimality gaps and solve these tasks.

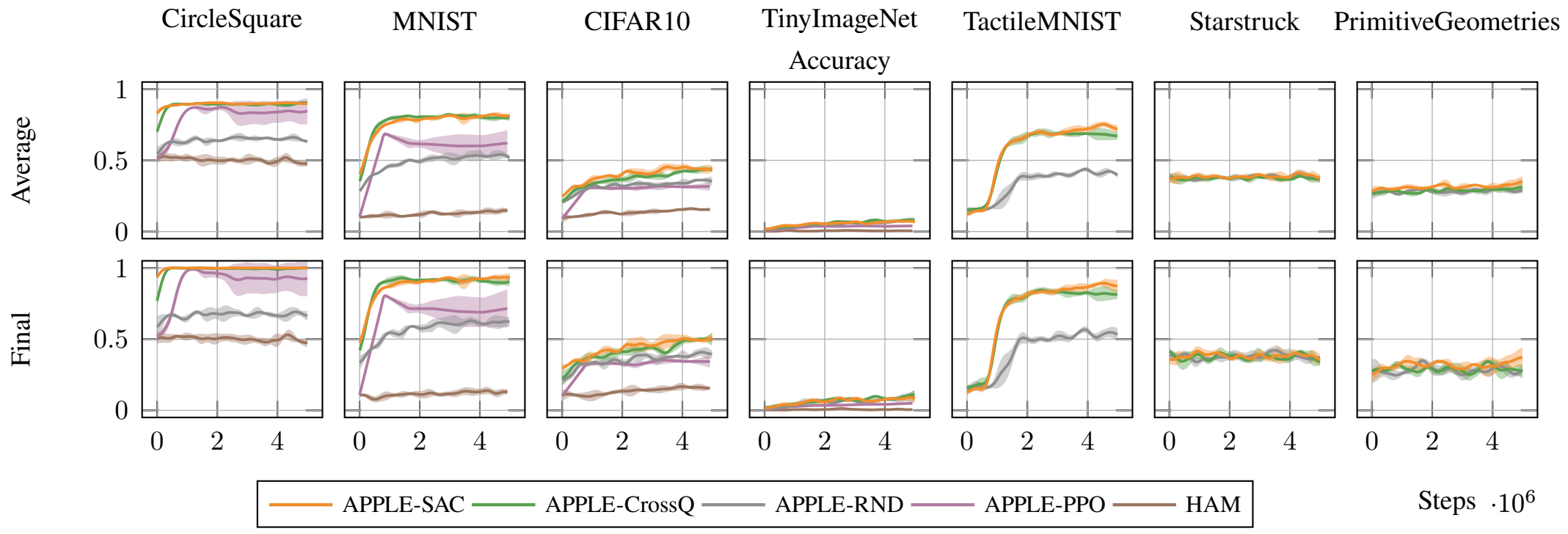

Figure 26: Average and final prediction accuracies for the baseline methods `APPLE-SAC`, `APPLE-CrossQ`, `APPLE-PPO`, `HAM` [44], and a random baseline `APPLE-RND` for the classification environments. All methods were trained on 5 seeds for 5M environment steps. Shaded areas represent one standard deviation. Metrics are computed on the evaluation variants of the tasks, except for CircleSquare, which has only two objects. For Starstruck, a correct classification requires predicting the exact number of stars. As `HAM` does not have a vision encoder, we evaluate it on non-tactile environments only. For `APPLE-PPO`, we found it to be unstable in combination with a vision encoder, so we also only evaluate it on non-tactile environments.

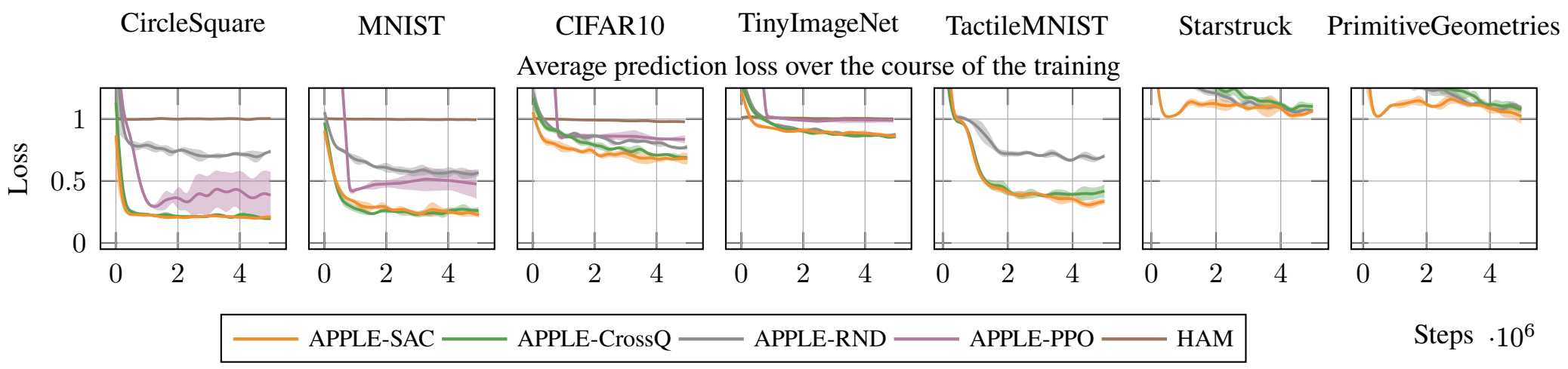

Figure 27: Average prediction loss for the baseline methods `APPLE-SAC`, `APPLE-CrossQ`, `APPLE-PPO`, `HAM` [44], and a random baseline `APPLE-RND` for the classification environments. All methods were trained on 5 seeds for 5M environment steps. Shaded areas represent one standard deviation. The loss is computed on the evaluation variants of the tasks, except for CircleSquare, which has only two objects.

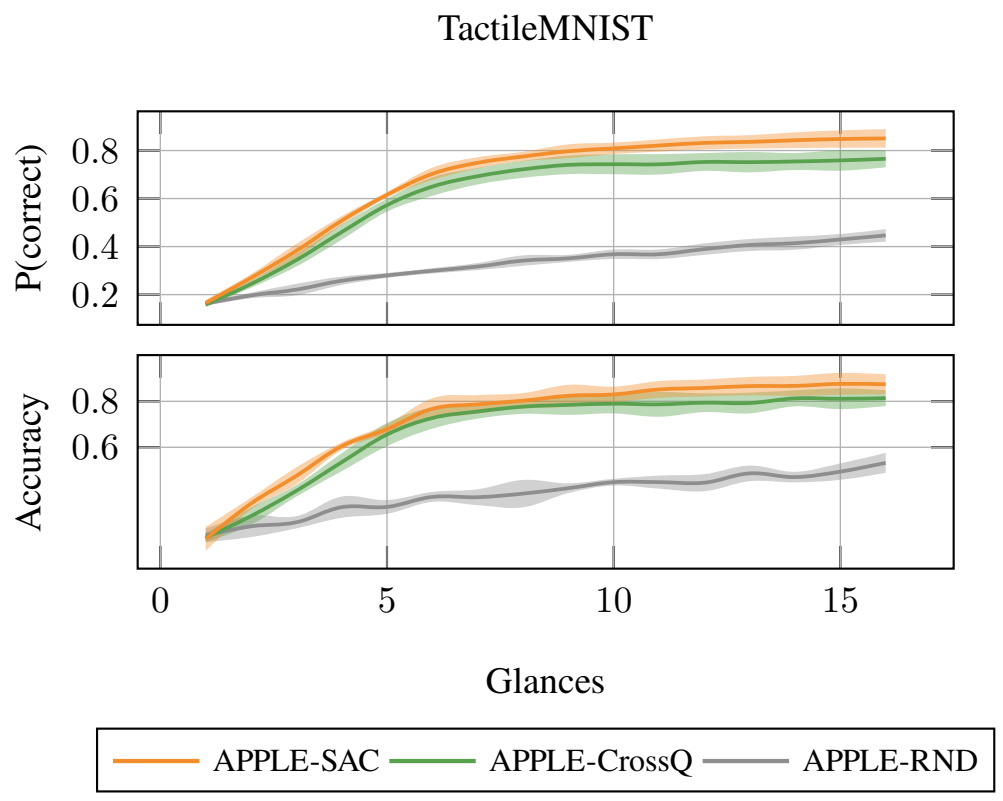

Figure 28: Exploration efficiency comparison of the final policies at the end of the training in the TactileMNIST environment. Shown are the predicted probability of the correct label (top) and accuracy (bottom) after $N$ tactile glances over the 16 steps until the episode terminates. The data is averaged over 5 seeds and smoothed over time steps within each seed.

**Image search environments** Next, in Fig. 29, we show results from the image search environments. These environments are particularly challenging and remain an open problem. In this setting, the agent receives only a small image glimpse and must actively explore a larger image to localize the target. By the end of the training, all methods achieve a loss below 1, i.e., they slightly outperform blind guessing, with `APPLE-CrossQ` performing best on both tasks, while `APPLE-SAC` is not clearly better than `APPLE-RND`. We expect that learning can be improved through targeted hyperparameter optimization tailored to these environments.

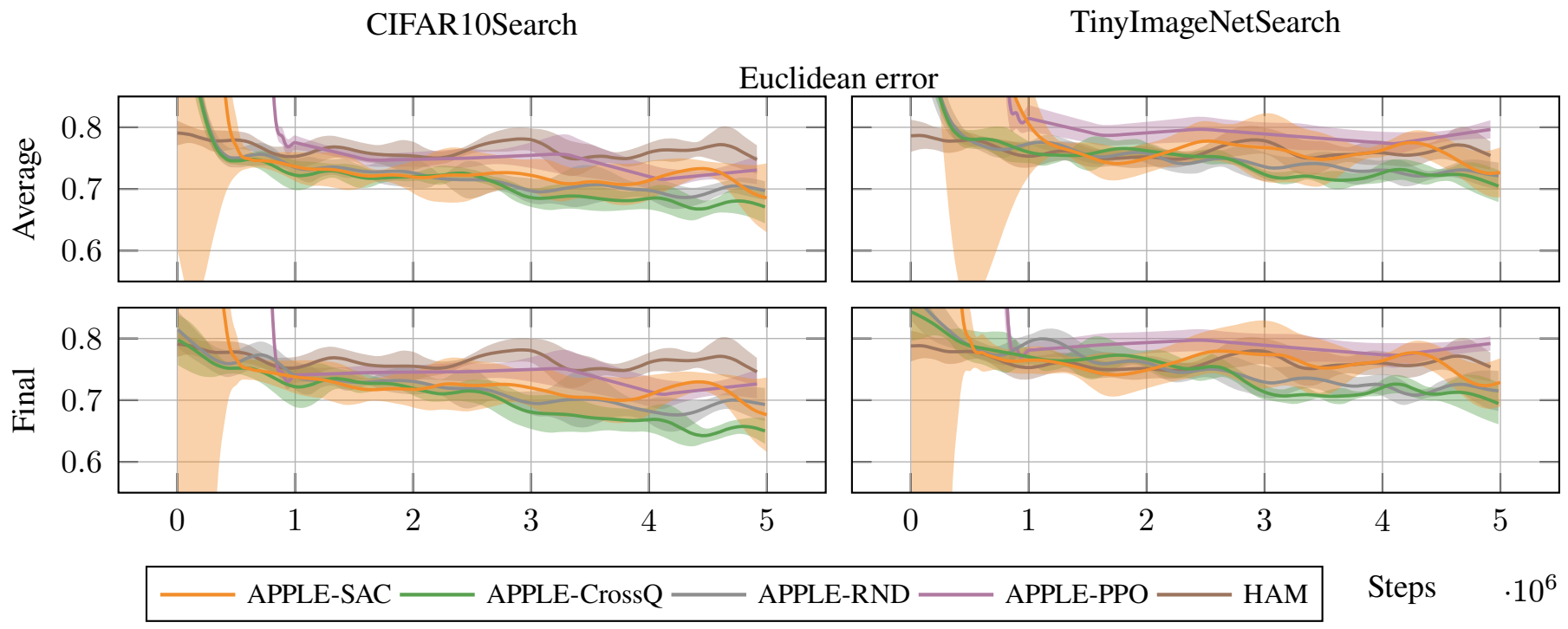

Figure 29: Average and final prediction errors for the baseline methods `APPLE-SAC`, `APPLE-CrossQ`, `APPLE-PPO`, `HAM` [44], and a random baseline `APPLE-RND` for the image search environments. All methods were trained on 5 seeds for 5M environment steps. Shaded areas represent one standard deviation.

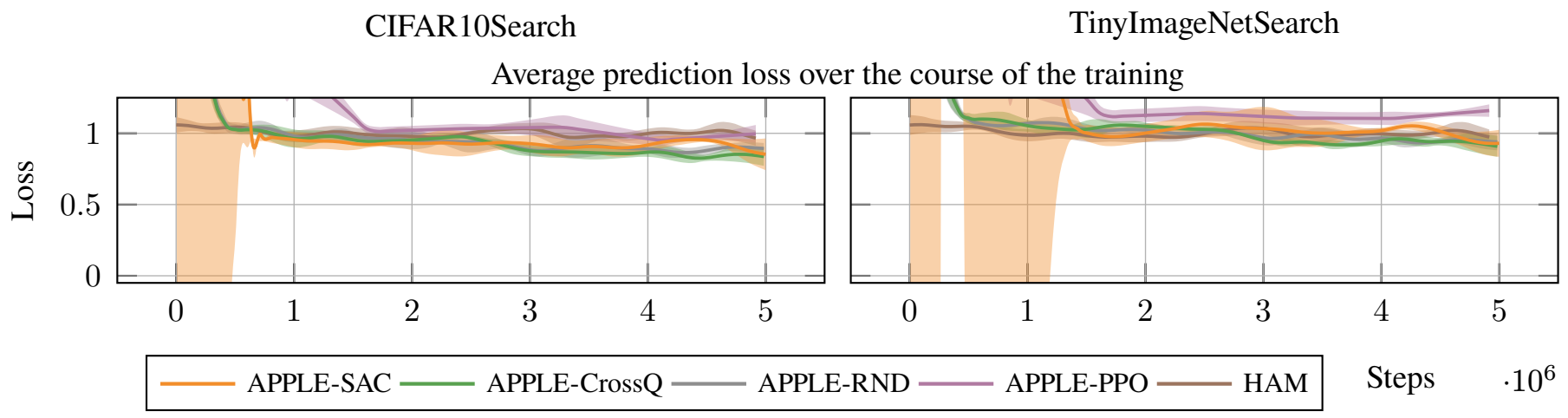

Figure 30: Average prediction loss for the baseline methods `APPLE-SAC`, `APPLE-CrossQ`, `APPLE-PPO`, `HAM` [44], and a random baseline `APPLE-RND` for the image search environments. All methods were trained on 5 seeds for 5M environment steps. Shaded areas represent one standard deviation. The loss is computed on the evaluation variants of the tasks. None of the methods performed substantially better than always predicting the mean on these challenging tasks.

**2D Localization environments** Next, we show the results for the 2D localization environments, in which the agent has to estimate its 2D pose. These are depicted in Fig. 31 and Fig. 32. In general, the baseline methods seem to struggle more with this set of tasks than with the classification environments: on LIDARLocMazeStatic, `APPLE-SAC` does not even outperform the random baseline `APPLE-RND`. These issues could be partly explained by the fact that none of the baseline methods was tuned on this type of task.

On the static LIDAR environments, only `APPLE-CrossQ` clearly outperforms `APPLE-RND` on both maps, `APPLE-SAC` and `APPLE-PPO` do so on LIDARLocRoomsStatic only, and `HAM` barely improves over blind guessing. Furthermore, the non-static LIDAR environments remain largely unsolved: on LIDARLocRooms, `APPLE-CrossQ` and `APPLE-SAC` clearly outperform `APPLE-RND`, but on LIDARLocMaze, both make only modest progress. The consistently low performance across these tasks is likely due to the substantial complexity and difficulty inherent in their design. In contrast to the static LIDAR environments, where agents can effectively memorize a fixed map, the non-static LIDAR environments present a novel map in each episode. As a result, solving these tasks requires the agent to generalize across maps by learning to associate its LIDAR observations with the corresponding bitmap representation provided as an additional input. This demands a level of perceptual and spatial reasoning that is inherently challenging — even for humans — which helps explain the limited success of current methods.

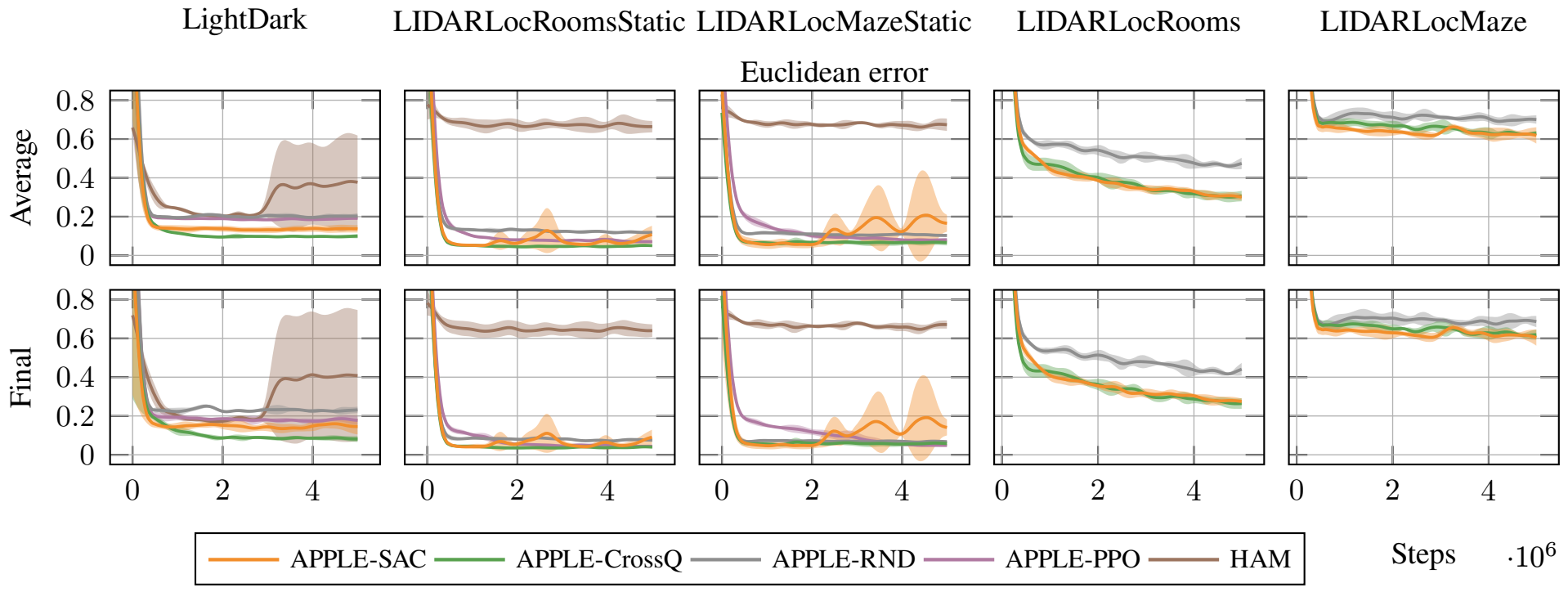

Figure 31: Average and final prediction errors for the baseline methods `APPLE-SAC`, `APPLE-CrossQ`, `APPLE-PPO`, `HAM` [44], and a random baseline `APPLE-RND` for the localization environments. All methods were trained on 5 seeds for 5M environment steps. Shaded areas represent one standard deviation. Note that LIDARLocMaze and LIDARLocRooms require a vision encoder, and thus `HAM` and `APPLE-PPO` cannot be evaluated on them.

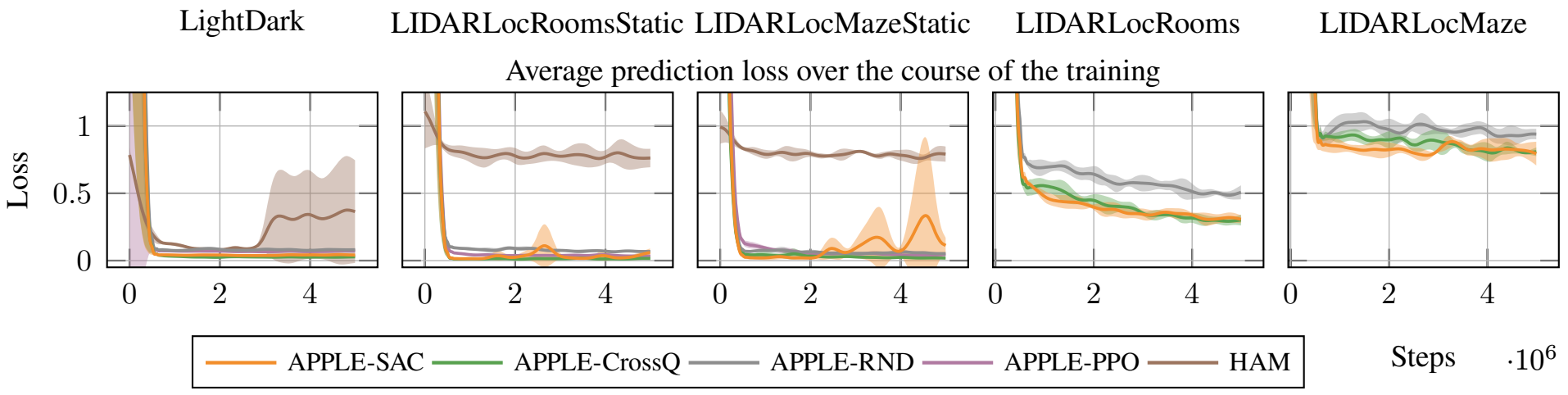

Figure 32: Average prediction loss for the baseline methods `APPLE-SAC`, `APPLE-CrossQ`, `APPLE-PPO`, `HAM` [44], and a random baseline `APPLE-RND` for the localization environments. All methods were trained on 5 seeds for 5M environment steps. Shaded areas represent one standard deviation. Note that LIDARLocMaze and LIDARLocRooms require a vision encoder, and thus `HAM` and `APPLE-PPO` cannot be evaluated on them.

**Pose estimation** Figure 33 and Fig. 35 show a comparison of `APPLE-SAC`, `APPLE-CrossQ`, and `APPLE-RND` on the pose estimation task within the Toolbox environment — summarized in Fig. 5. In this evaluation, `APPLE-CrossQ` achieves the lowest error across both the linear and angular components of the pose (final error of 3.6 cm and 74°, vs. 3.8 cm and 85° for `APPLE-SAC` and 4.4 cm and 89° for `APPLE-RND`), although it varies strongly across seeds. Qualitatively, in some seeds, `APPLE-CrossQ` learned an exploration strategy of first moving the sensor in a circle to find the wrench and then moving toward one of the ends to determine the exact position and orientation. `APPLE-SAC` did not learn an effective strategy within 5M steps and, thus, is close to `APPLE-RND`.

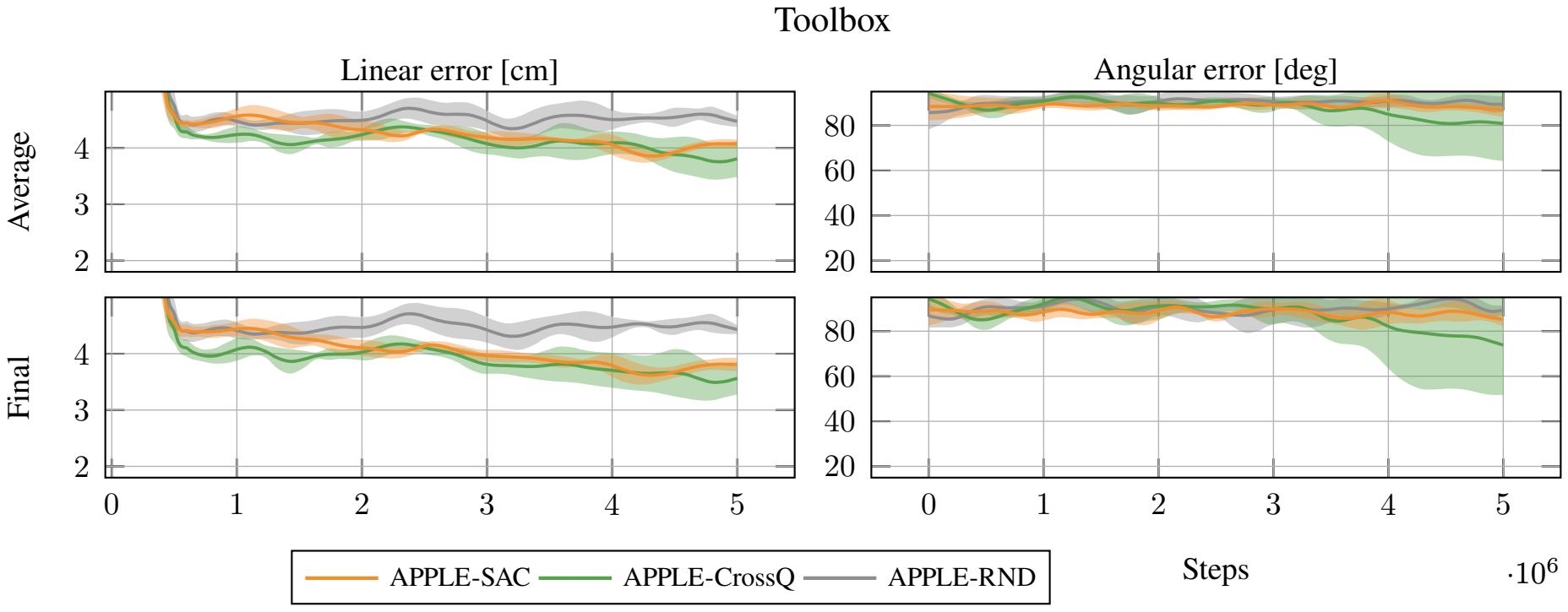

Figure 33: Average and final prediction errors for the baseline methods `APPLE-SAC`, `APPLE-CrossQ`, and a random baseline `APPLE-RND` for the Toolbox environment. All methods were trained on 5 seeds for 5M environment steps. Shaded areas represent one standard deviation. We compute the linear and angular displacement between the prediction and the actual object pose as a metric.

**Longer training on Toolbox** Fig. 34 and Fig. 36 show the Toolbox results of runs trained for 10M instead of 5M environment steps. Since the learning rate schedule and the ramp-up of the update-to-data ratio are defined relative to the total number of steps, these runs train differently from the start, not only for longer. In contrast to the 5M runs, `APPLE-SAC` learns an effective strategy in all seeds: after 5M steps, its final error is 2.4 cm and 26° (vs. 3.8 cm and 85° for the 5M runs), and it reaches 1.6 cm and 20° after 10M steps. Qualitatively, `APPLE-SAC` learns the same strategy that some `APPLE-CrossQ` seeds found: it first moves the sensor in a circle to find the wrench and then moves toward one of its ends. `APPLE-RND` does not improve over its 5M result, and `APPLE-CrossQ` improves only moderately (final error of 3.1 cm and 59° after 10M steps), again varying strongly across seeds. This difference suggests that `APPLE-SAC` is sensitive to these schedules on Toolbox and that Toolbox is not yet saturated after 5M steps.

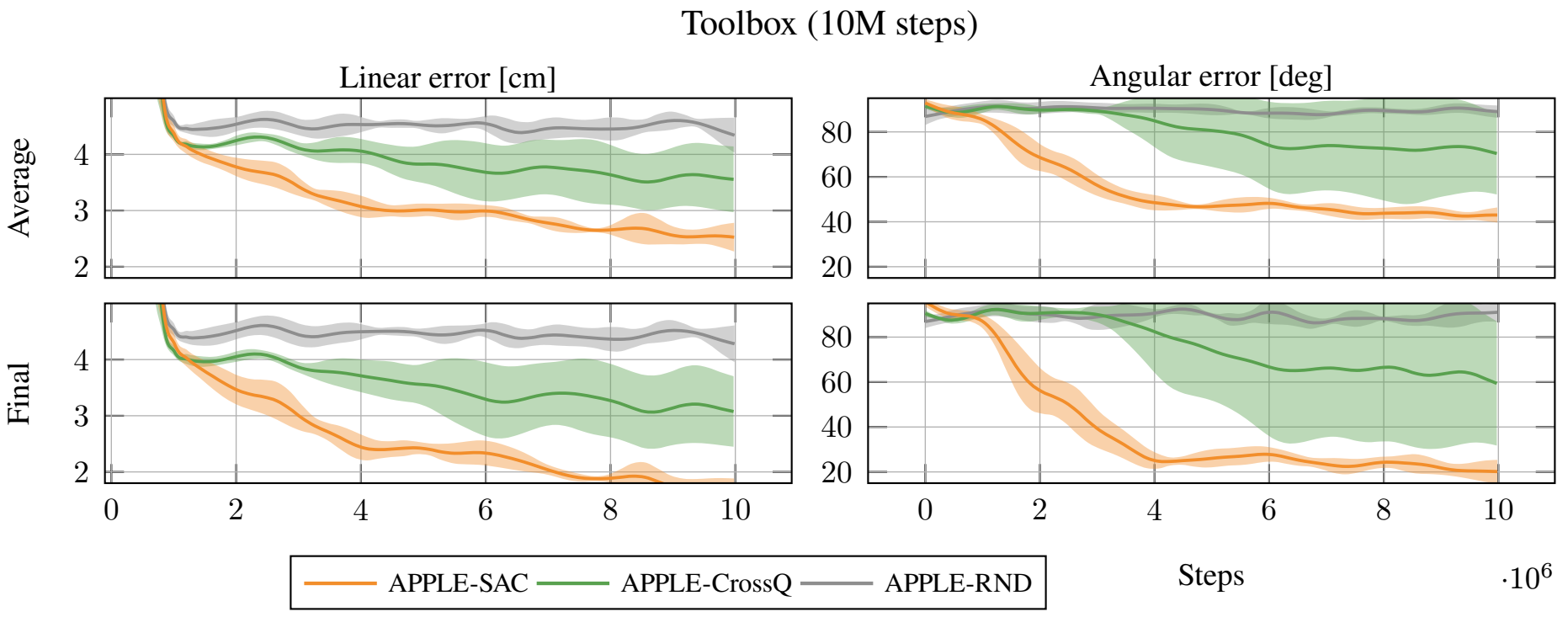

Figure 34: Average and final prediction errors for the baseline methods `APPLE-SAC`, `APPLE-CrossQ`, and a random baseline `APPLE-RND` for the Toolbox environment, trained for 10M environment steps on 5 seeds. Shaded areas represent one standard deviation.

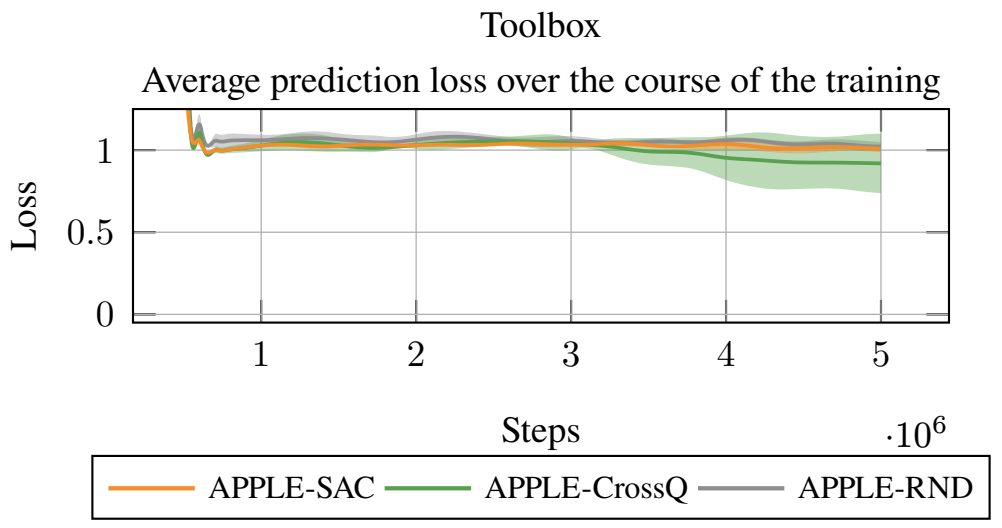

Figure 35: Average prediction loss for the baseline methods `APPLE-SAC`, `APPLE-CrossQ`, and a random baseline `APPLE-RND` for the Toolbox environment. All methods were trained on 5 seeds for 5M environment steps. Shaded areas represent one standard deviation.

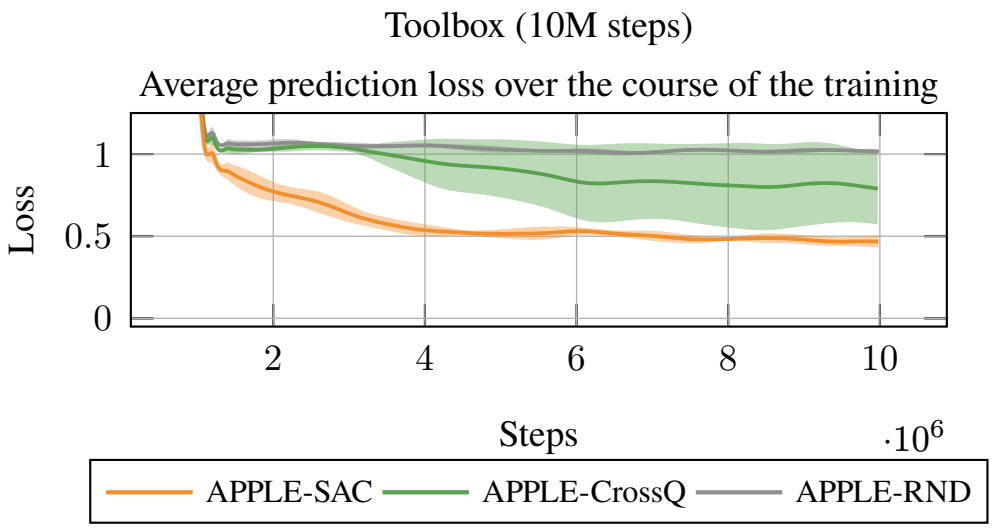

Figure 36: Average prediction loss for the baseline methods `APPLE-SAC`, `APPLE-CrossQ`, and a random baseline `APPLE-RND` for the Toolbox environment, trained for 10M environment steps on 5 seeds. Shaded areas represent one standard deviation.

**Center of mass estimation** In ABCCenterOfMass, the agent must develop a strategy to estimate the center of mass of an object from the ABC Dataset. This task proved to be one of the most challenging among the tasks of the TMBS. As shown in Fig. 37 and Fig. 38, only `APPLE-SAC` outperforms the random baseline `APPLE-RND` in this setting (average prediction loss of 0.54 vs. 0.66; final linear error of 1.1 cm vs. 1.4 cm), while `APPLE-CrossQ` is not better than `APPLE-RND` at the end of training.

**Volume estimation environments** Finally, in TactileMNISTVolume, PrimitiveGeometriesVolume, and ABCVolume, the agent must develop a strategy to estimate the volume of an object. On TactileMNISTVolume, `APPLE-CrossQ` and `APPLE-SAC` demonstrated similar performance, both outperforming `APPLE-RND` (average error of 1.0 cm$^3$ for `APPLE-CrossQ` and 1.1 cm$^3$ for `APPLE-SAC` vs. 1.3 cm$^3$). ABCVolume, in contrast, is challenging for both `APPLE` baselines (Fig. 39 and Fig. 40): neither ends with a clearly lower error than `APPLE-RND` (final error of 9.0 cm$^3$ for `APPLE-SAC` and 9.2 cm$^3$ for `APPLE-CrossQ` vs. 9.1 cm$^3$).

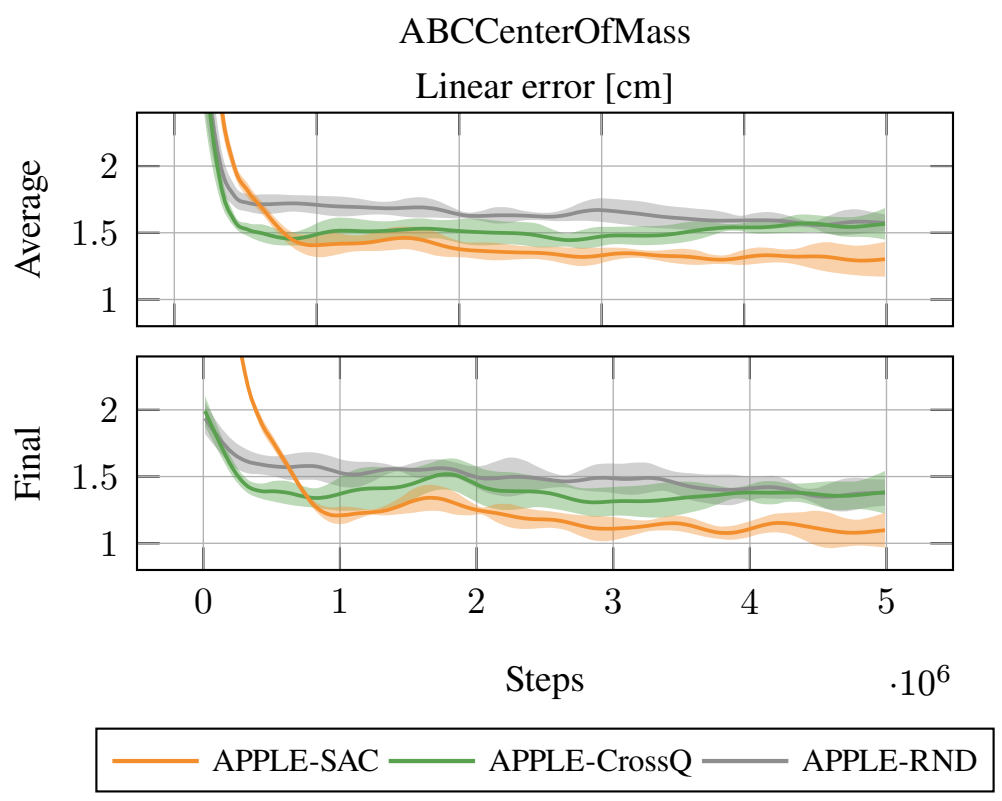

Figure 37: Average and final prediction errors for the baseline methods `APPLE-SAC`, `APPLE-CrossQ`, and a random baseline `APPLE-RND` for the ABCCenterOfMass environment. All methods were trained on 5 seeds for 5M environment steps. Shaded areas represent one standard deviation. Metrics are computed on the evaluation variant of the task (ABCCenterOfMass-test), which contains objects unseen during training.

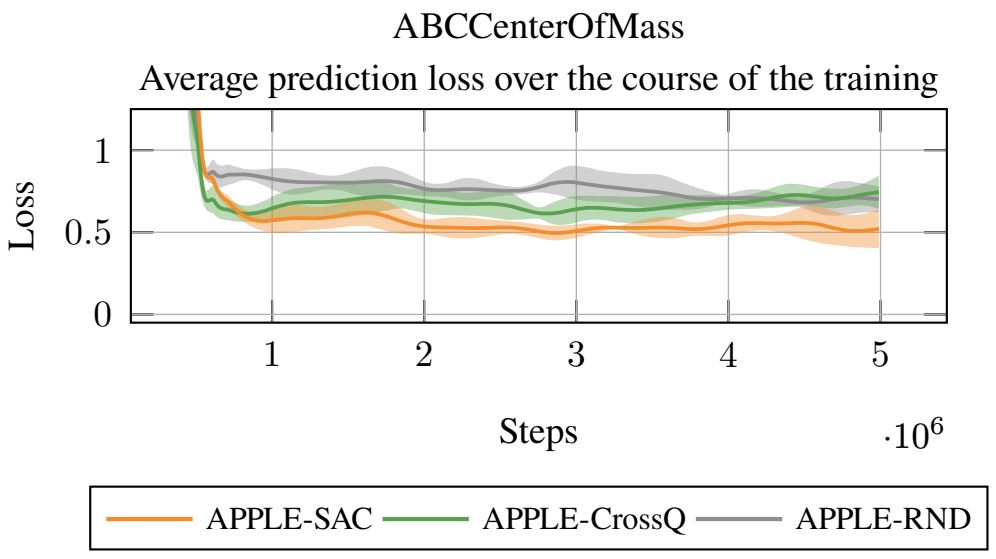

Figure 38: Average prediction loss for the baseline methods `APPLE-SAC`, `APPLE-CrossQ`, and a random baseline `APPLE-RND` for the ABCCenterOfMass environment. All methods were trained on 5 seeds for 5M environment steps. Shaded areas represent one standard deviation. Metrics are computed on the evaluation variant of the task.

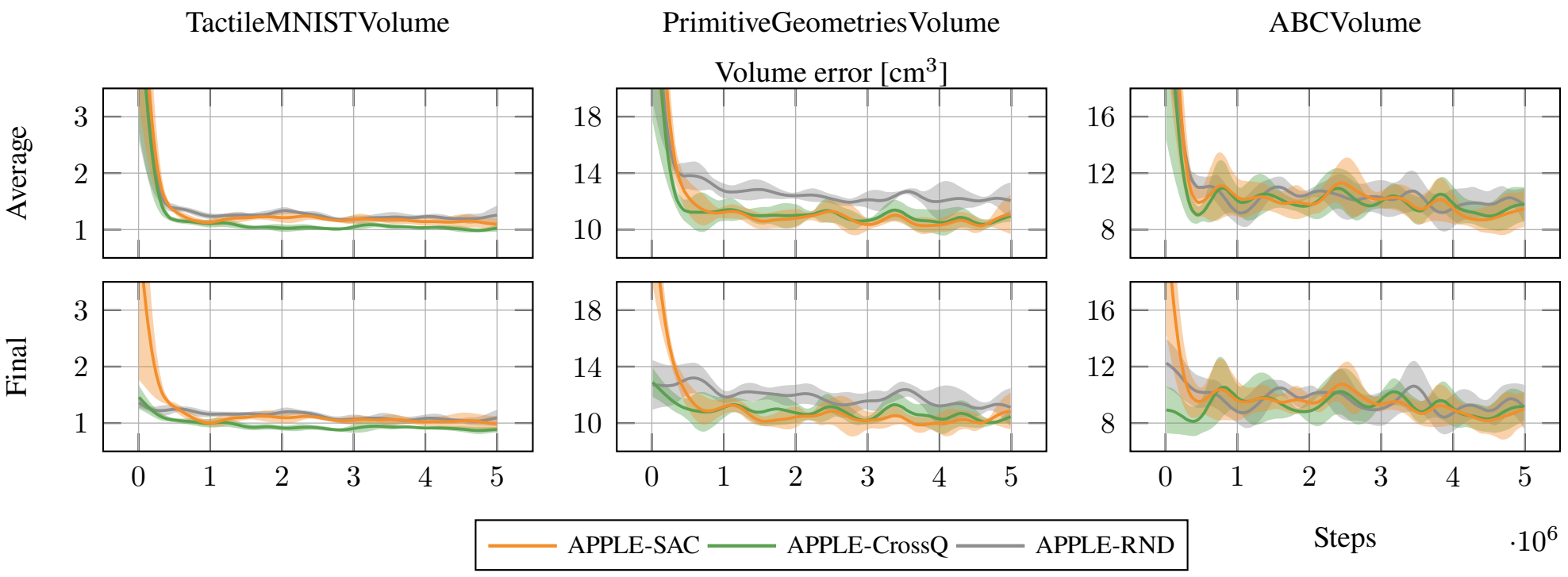

Figure 39: Average and final prediction errors for the baseline methods `APPLE-SAC`, `APPLE-CrossQ`, and a random baseline `APPLE-RND` for the TactileMNISTVolume, PrimitiveGeometriesVolume, and ABCVolume environments. All methods were trained on 5 seeds for 5M environment steps. Shaded areas represent one standard deviation. Metrics are computed on the evaluation variants of the tasks (e.g., TactileMNISTVolume-test), which contain objects unseen during training.

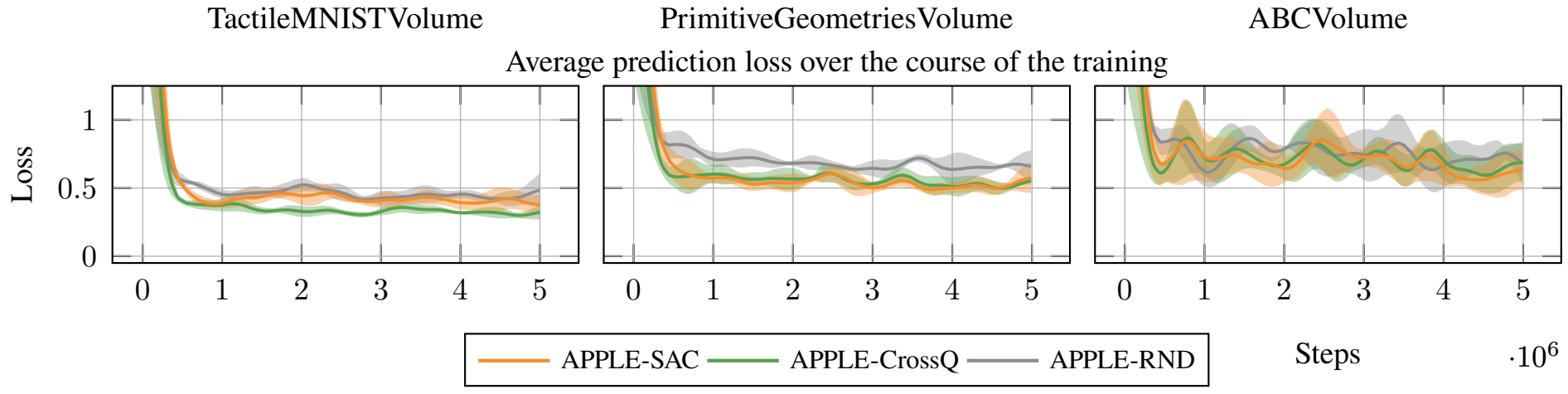

Figure 40: Average prediction loss for the baseline methods `APPLE-SAC`, `APPLE-CrossQ`, and a random baseline `APPLE-RND` for the volume estimation environments. All methods were trained on 5 seeds for 5M environment steps. Shaded areas represent one standard deviation. Metrics are computed on the evaluation variants of the tasks.

## H Sim-to-Real Transfer Experiments

TMBS is designed as a simulated benchmark, and its main purpose is the reproducible comparison of active perception methods in simulation. Nevertheless, the Real Tactile MNIST dataset (Section 3.5) allows us to test whether policies and classifiers trained in our simulation transfer to real tactile data. In this section, we describe a recipe that achieves such transfer on the TactileMNIST classification task and analyze which of its components matter. We consider these experiments a proof of concept rather than a solution to the sim-to-real problem: the real-world evaluation is restricted to the touches recorded in our dataset, and a substantial gap between simulated and real accuracy remains.

**Interactive evaluation on real data** Since the Real Tactile MNIST dataset is static, a real-world episode cannot query arbitrary sensor poses. Instead, the *TactileMNISTRealSnap* environment replays the recorded touches interactively: for each step, the agent's commanded sensor position is snapped to the closest touch (Euclidean distance in the plane of the cell) among the next 16 prerecorded touches of the current digit (out of the 256 touches recorded per digit), and the corresponding real tactile image is returned as the observation. The selected touch then becomes the current one, so an episode always moves forward through the recording: the touches of a digit are visited in the order in which they were recorded, and skipped touches cannot be revisited. This ordering retains temporal coherence: as the digits could shift slightly during data collection (Section 3.5), touches that lie far apart in the recording may observe the object in different poses, and replaying them out of order would mix these poses within one episode. Each episode starts at the first recorded touch of a randomly drawn test digit. To give the agent the same action semantics during training, we train in the *TactileMNISTSnap* simulation environment, which mimics this selection by sampling 16 touch positions uniformly over the cell in each step and moving the sensor to the one closest to the commanded position. Episodes have a step limit of 16 in both environments. We evaluate on the test split of the real dataset.

**Recipe** We found the following three components to be needed for sim-to-real transfer:

1. **Realistic rendering.** During training, the simulated depth map of each touch is rendered into a realistic tactile image with the depth-to-RGB generator of our CycleGAN (Appendix C). In an ablation, we replace the CycleGAN entirely: we render with Taxim during training and evaluate on the raw real images.
2. **Real-image round trip.** A small domain gap between the CycleGAN renderings and real tactile images remains, one major source being the degradation of the gel over the course of the data collection (Appendix E). To close this gap, each real tactile image is passed through a full round trip at test time: the inverse (RGB-to-depth) generator of the same CycleGAN translates it into a depth map, which the depth-to-RGB generator then renders again. The policy thus sees CycleGAN renderings in both domains. The round trip is possible because a CycleGAN by design learns generators for both directions.
3. **Background depth overlay.** The round trip does not produce clean depth maps: even for real touches without contact, the inverse generator estimates small, structured depth artifacts. To expose the agent to these artifacts during training, we add the estimated depth map of a randomly drawn real touch *without* object contact (sensor more than 1 mm above the contact height) to each simulated depth map before rendering. As the real artifacts are 6–25% weaker on touches with contact, the overlay is scaled by a factor drawn from $\mathcal{U}(0.65, 0.95)$ if the simulated touch indents the gel by more than 0.1 mm.

The final recipe applies no image-space randomization (noise, gain, offset, or lighting changes): since the policy sees CycleGAN renderings in both domains, the remaining differences stem from the depth maps estimated by the round trip, which the overlay targets. The overlay is calibrated to the artifacts of the round trip and only makes sense in combination with it.

**Setup** We train `APPLE-SAC` with a vision encoder for 2.5M environment steps on 5 seeds per configuration. For the full recipe, we additionally train `APPLE-CrossQ`. We report the classification accuracy after the last touch of an episode on the real test split and on the simulated training environment. For each seed, we average the accuracy over the evaluations between 2.0M and 2.5M steps, and we report the mean and standard deviation over seeds. Chance level is 10%. All environments of the recipe are available in TMBS: we train on `TactileMNISTSnap-CycleGAN-DR-train-v0` and evaluate on `TactileMNISTRealSnap-CycleGAN-test-v0`, which applies the real-image round trip; for volume estimation, the corresponding environments are `TactileMNISTVolumeSnap-CycleGAN-DR-train-v0` and `TactileMNISTVolumeRealSnap-CycleGAN-test-v0`.

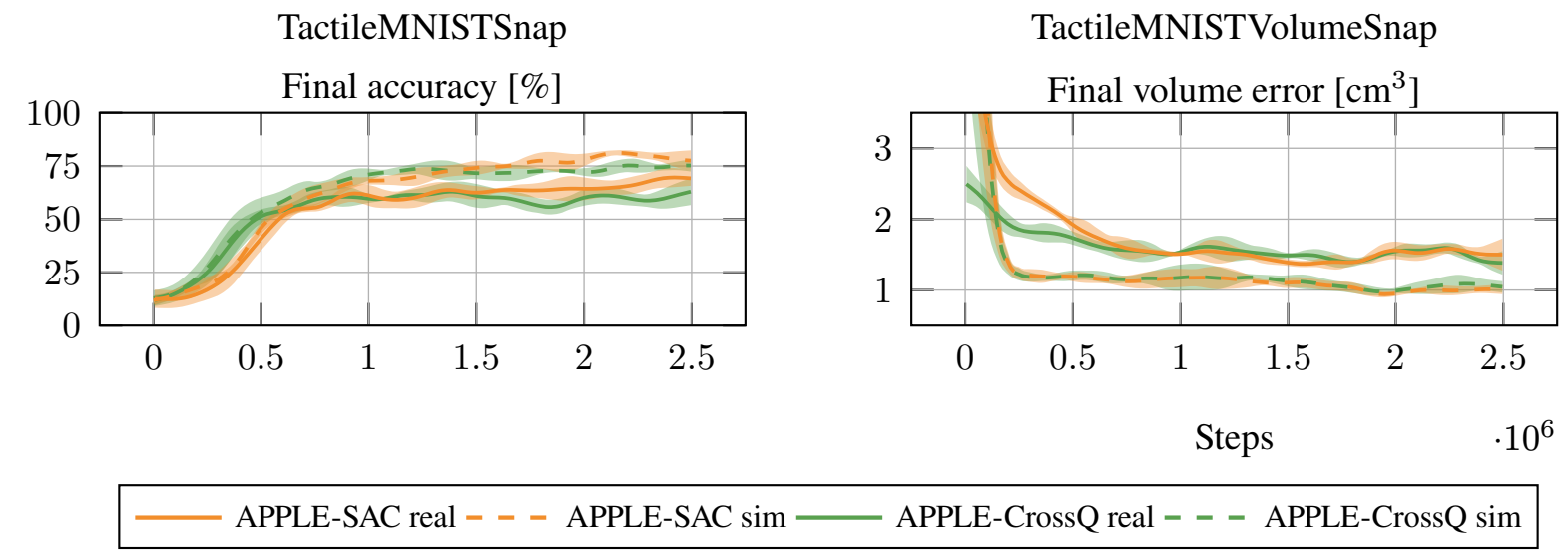

Figure 41: Final classification accuracy on TactileMNIST (left) and final volume error on TactileMNISTVolume (right) of `APPLE-SAC` and `APPLE-CrossQ` trained with the full sim-to-real recipe, evaluated on real data (*TactileMNISTRealSnap* and *TactileMNISTVolumeRealSnap*, test split; solid) and in simulation (*TactileMNISTSnap* and *TactileMNISTVolumeSnap*; dashed). All methods were trained on 5 seeds for 2.5M environment steps. Shaded areas represent one standard deviation.

**Results** Fig. 42 summarizes the results, and Fig. 41 shows the learning curves on real and simulated data. With the full recipe, `APPLE-SAC` reaches a final accuracy of $67.0 \pm 3.6\%$ on the real test digits ($54.9 \pm 2.1\%$ averaged over all steps of an episode), compared to $79.6 \pm 2.1\%$ in simulation, i.e., it retains 84% of its simulated accuracy. Without the CycleGAN, i.e., rendering with Taxim during training and evaluating on the raw real images, `APPLE-SAC` reaches an even higher simulated accuracy of $84.4 \pm 1.6\%$ but only $11.9 \pm 2.8\%$ on the real test digits, close to chance.

The ablations show that the background depth overlay is an essential component: without it, real accuracy drops to $40.3 \pm 1.7\%$, while the simulated accuracy is even slightly higher ($81.1 \pm 1.8\%$). Hence, a policy that performs well in simulation is no indication of transfer unless the agent was exposed to the artifacts of the real-image pipeline during training.

The real-image round trip is equally essential: evaluated on the raw real images, `APPLE-SAC` reaches only $18.1 \pm 3.4\%$, close to chance. Evaluating the final `APPLE-CrossQ` policies of the full recipe on the raw real images yields a similar accuracy of $15.4 \pm 4.2\%$.

`APPLE-CrossQ` reaches a lower real accuracy than `APPLE-SAC` with the same recipe ($60.5 \pm 3.2\%$). The main reason is not the transfer itself: `APPLE-CrossQ` already reaches a lower accuracy in the randomized simulation ($74.4 \pm 2.4\%$ after 2.5M steps, Fig. 41) and retains a similar fraction of it on real data (81%).

**Volume estimation** We applied the same recipe to the volume estimation task on the real digits (*TactileMNISTVolumeRealSnap*), standardizing the volume targets with the statistics of the simulated training objects in both domains. After 2.5M steps, the final absolute volume error on the real test digits is $1.54 \pm 0.05$ cm³ for `APPLE-SAC` and $1.51 \pm 0.04$ cm³ for `APPLE-CrossQ` (simulation: 1.00 cm³ and 1.06 cm³, see Fig. 41, right). Both agents systematically overestimate the volumes of the real digits: the mean ratio of estimated to true volume is 1.08 for `APPLE-SAC` (1.10 in simulation) and 1.19 for `APPLE-CrossQ` (1.09 in simulation). Evaluated on the raw real images without the round trip, the final `APPLE-SAC` policies reach an error of $5.20 \pm 0.94$ cm³, and the volumes are overestimated by a factor of 2.27.

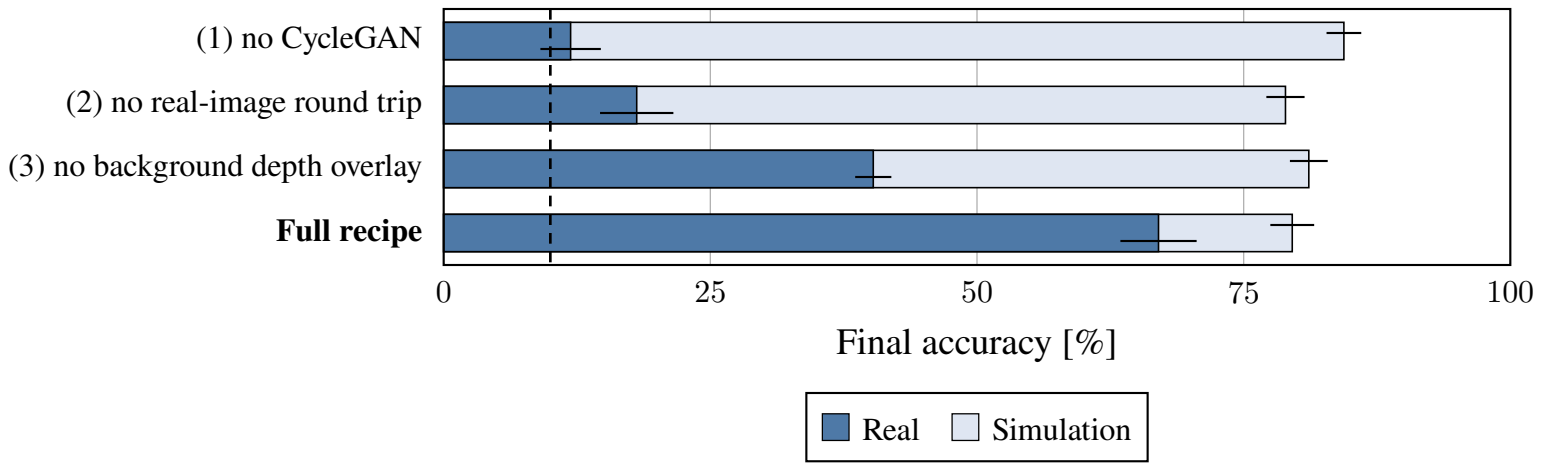

Figure 42: Final classification accuracy of `APPLE-SAC` on real (*TactileMNISTRealSnap*, test split; solid) and simulated tactile data (*TactileMNISTSnap*; light) for the full sim-to-real recipe and its ablations. Each ablation removes one component from the full recipe; without the CycleGAN, we render with Taxim during training and, since the real-image round trip requires the CycleGAN, evaluate without it on the raw real images. Bars show the mean over 5 seeds of the accuracy averaged between 2.0M and 2.5M environment steps; lines show one standard deviation. The dashed line marks chance level (10%).

Table 16: Overview of the input modalities provided by the different environments. By *scalar input*, we mean real-valued vectors of arbitrary length. *Visual input* refers to image data represented by real-valued three-dimensional tensors of dimensions $(W \times H \times C)$, where $W$ and $H$ are width and height, respectively, and $C$ is the number of channels. Some environments, such as CircleSquare or MNIST, provide the agent with only small glimpses of an image. Although these small glimpses are technically images, they are so small that it does not make sense to use a vision encoder to encode them. Hence, instead, we flatten them and process them as scalar data.

| Suite | Environment | Task type | Scalar input | Vision input | $(W \times H \times C)$ | Vision encoder |
|---|---|---|---|---|---|---|
| `ap_gym` | CircleSquare | Classification | ✓ | (✓) | $(5 \times 5 \times 1)$ | ✗ |
| | MNIST | Classification | ✓ | (✓) | $(5 \times 5 \times 1)$ | ✗ |
| | CIFAR10 | Classification | ✓ | (✓) | $(5 \times 5 \times 3)$ | ✗ |
| | TinyImageNet | Classification | ✓ | (✓) | $(10 \times 10 \times 3)$ | ✗ |
| | CIFAR10Search | Regression | ✓ | (✓) | $(5 \times 5 \times 3)$ | ✗ |
| | TinyImageNetSearch | Regression | ✓ | (✓) | $(10 \times 10 \times 3)$ | ✗ |
| | LightDark | Regression | ✓ | ✗ | | ✗ |
| | LIDARLocRoomsStatic | Regression | ✓ | ✗ | | ✗ |
| | LIDARLocMazeStatic | Regression | ✓ | ✗ | | ✗ |
| | LIDARLocRooms | Regression | ✓ | ✓ | $(32 \times 32 \times 1)$ | ✓ |
| | LIDARLocMaze | Regression | ✓ | ✓ | $(21 \times 21 \times 1)$ | ✓ |
| TMBS | TactileMNIST | Classification | ✓ | ✓ | $(64 \times 64 \times 3)$ | ✓ |
| | Starstruck | Classification | ✓ | ✓ | $(64 \times 64 \times 3)$ | ✓ |
| | PrimitiveGeometries | Classification | ✓ | ✓ | $(64 \times 64 \times 3)$ | ✓ |
| | Toolbox | Regression | ✓ | ✓ | $(64 \times 64 \times 3)$ | ✓ |
| | ABCCenterOfMass | Regression | ✓ | ✓ | $(64 \times 64 \times 3)$ | ✓ |
| | TactileMNISTVolume | Regression | ✓ | ✓ | $(64 \times 64 \times 3)$ | ✓ |
| | PrimitiveGeometriesVolume | Regression | ✓ | ✓ | $(64 \times 64 \times 3)$ | ✓ |
| | ABCVolume | Regression | ✓ | ✓ | $(64 \times 64 \times 3)$ | ✓ |

# I Implementation Details

The implementation of `APPLE`, `APPLE-PPO`, and `HAM` used in the experiments of this work is based on JAX [70] and the Flax deep-learning framework [78], and uses the Hugging Face transformer implementations [79]. For performance reasons, the entire training loop is JIT-compiled, and all interaction with the environment happens through host callbacks. This scheme ensures maximal performance at the cost of limiting flexibility in the implementation.

A bottleneck in many deep-learning applications is the transfer of data between VRAM (GPU) and RAM (CPU). To mitigate this bottleneck as much as possible, we store the replay buffer in the VRAM, which limits the interactions between the GPU and the CPU to stepping the environment and logging data. However, typically, there is less VRAM than RAM present, which limits the size of the replay buffer in the case of environments with vision input. We found that our Nvidia RTX A5000 GPUs with 24GB of VRAM can hold around 3M transitions in the replay buffer before running out of space if we scale visual inputs to $32 \times 32$px. Hence, for the vision-encoder configurations, we use a replay buffer size of 3M, while for the non-vision-encoder configurations, we use a replay buffer size equal to the total number of environment steps. An overview of which environments require vision encoders is given in Table 16.

All our experiments were conducted on Nvidia RTX A5000 or RTX 3090 GPUs with the hyperparameters defined in Table 2. For vision-encoder configurations, we fully allocate one GPU for each run. Depending on the algorithm, a run of $5M$ steps takes around 40–50 hours.

Since non-vision-encoder configurations do not require much VRAM, we run multiple runs on the same GPU to use resources more efficiently. Depending on the environment and algorithm, the data of up to 28 parallel runs fits into the VRAM of a single RTX A5000/3090 GPU. Hence, the runtime of these runs varies greatly with the number of runs per GPU and is, thus, not meaningful to report. However, we found that `HAM` and `APPLE-PPO` runs are generally about four times as fast as `APPLE` runs.

Table 17: Comparison of Active Tactile Perception Benchmarks. Tactile modalities are **bolded**.

| **Method** | **Dataset Available** | **Active Benchmark** | **Sensor Modality** |
|---|---|---|---|
| Active Vision Grasp [51] | ✗ | ✓ | Vis |
| R3ED [52] | ✓ | ✓ | Vis |
| Robotic Vision Challenge [53] | ✗ | ✓ | Vis |
| NBV_Bench [50] | ✓ | ✓ | Vis |
| AVD [48, 49] | ✓ | ✓ | Vis |
| Active Object Search [54] | ✓ | ✓ | Vis |
| ActiView [55] | ✓ | ✓ | Vis+Text |
| TIP Bench. [19] | ✗ | ✗ | **Tac** |
| FoTa [61] | ✓ | ✗ | **Tac** |
| YCB-Slide [80] | ✓ | ✗ | **Tac** |
| TacBench [11] | ✓ | ✗ | **Tac** |
| ActiveCloth [46] | ✓ | ✗ | **Tac** |
| Touch and Go [59] | ✓ | ✗ | **Tac**+Vis |
| FeelSight [12] | ✓ | ✗ | **Tac**+Vis |
| ViTac [56] | ✗ | ✗ | **Tac**+Vis |
| SSVTP [47] | ✓ | ✗ | **Tac**+Vis |
| GelFabric [57] | ✓ | ✗ | **Tac**+Vis |
| VisGel [58] | ✓ | ✗ | **Tac**+Vis |
| PHYSICLEAR [21] | ✓ | ✗ | **Tac**+Text |
| TVL [20] | ✓ | ✗ | **Tac**+Vis+Text |
| Touch100k [60] | ✓ | ✗ | **Tac**+Vis+Text |
| ObjectFolder [62, 63, 81] | ✓ | ✗ | **Tac**+Vis+Audio |
| **Ours** | ✓ | ✓ | **Tac** |

# J Related Datasets

A comparison of existing tactile datasets and benchmarks is provided in Table 17. Prior work has primarily focused on passive tactile recognition, such as object classification, material understanding, and multimodal perception settings, where tactile observations are collected from predefined or fixed interactions, or vision-based active perception tasks. In contrast, our proposed TMBS explicitly targets *active tactile perception*, where tactile exploration itself is an integral component of the perceptual process. To the best of our knowledge, TMBS is the first benchmark suite designed specifically to support reproducible evaluation of active tactile perception strategies under standardized conditions.

## Appendix References